%% file: main.tex
\pdfoutput=1

\documentclass[10pt,twocolumn,letterpaper]{article}

\usepackage[pagenumbers]{cvpr}      %

\input{preamble}

\usepackage{stfloats}
\usepackage{placeins}

\newsavebox{\supptabbox}

\definecolor{cvprblue}{rgb}{0.21,0.49,0.74}
\usepackage[pagebackref,breaklinks,colorlinks,allcolors=cvprblue]{hyperref}

\def\confName{}
\def\confYear{}

\title{\ours{}: Who Is Talking to Whom? Grounding Acoustic Direction\\ in Vision for Online Multi-Party Conversation}

\author{Nader Lobandi$^{*}$ \quad Lombe Chileshe$^{*}$ \quad Mohammad Mehdi Hosseini \quad Mohammad H. Mahoor$^{\dag}$\\
University of Denver\\
2199 S. University Blvd., Denver, CO 80208\\
{\tt\small \{nader.lobandi, Lombe.Chileshe, MohammadMehdi.Hosseini, Mohammad.Mahoor\}@du.edu}
}

\begin{document}
\input{fig1_data}
\input{fig2_data}
\twocolumn[{%
\renewcommand\twocolumn[1][]{#1}%
\maketitle
\vspace{-1.2em}
\begin{center}
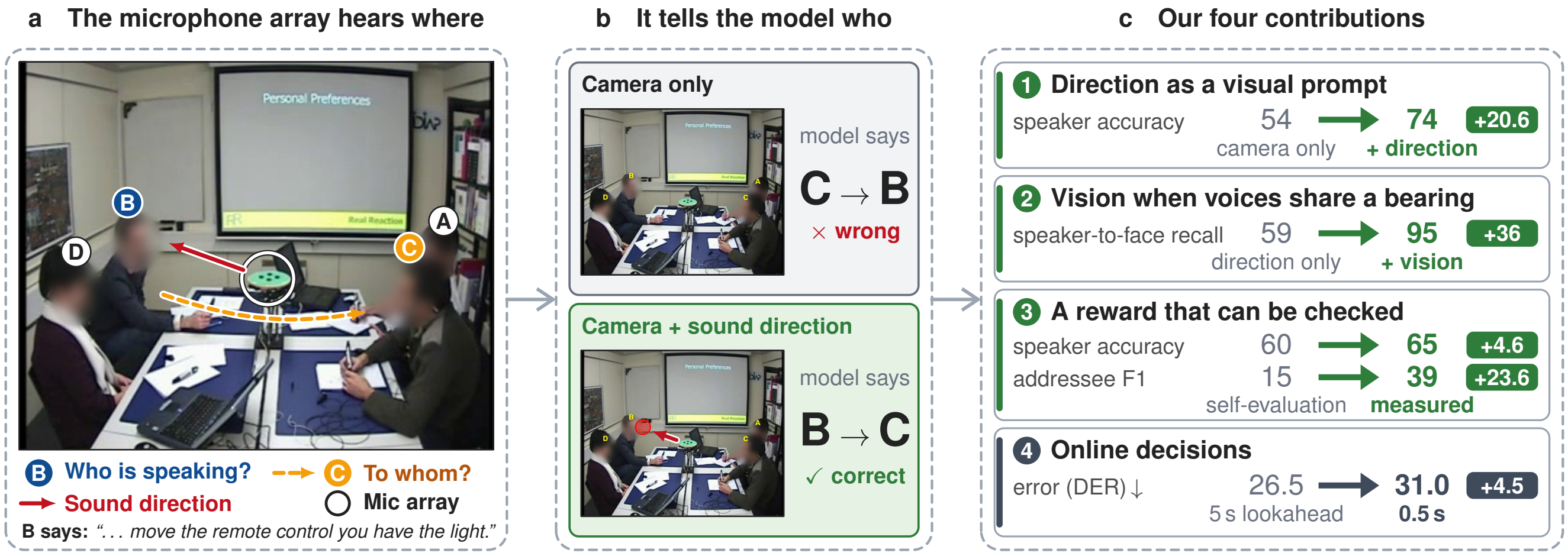

\input{fig_teaser}
\captionof{figure}{\textbf{Sound tells us \emph{where}; the camera tells us \emph{who}.}
\textbf{(a)}~A table-top microphone array measures the direction a voice comes from
($\hat\theta=\FigAz^\circ$), and a registration learned without labels places it in the camera image.
\textbf{(b)}~With the camera alone, a vision--language model confuses the speaker with the listener; shown the
direction as a red disc, it answers correctly. \textbf{(c)}~The four contributions and what each buys: the direction as a prompt, vision when overlapping voices
share a bearing, a reward checked against the measurement in place of self-evaluation, and online decisions at a
measured cost (green: ours, gains in points; dark grey: a cost). Faces are blurred here only.}
\label{fig:teaser}
\end{center}
\vspace{0.2em}
}]
\iftoggle{cvprfinal}{{\let\thefootnote\relax\footnotetext{\raggedright $^{*}$These authors contributed equally to this work.\quad
$^{\dag}$Corresponding author: {\tt Mohammad.Mahoor@du.edu}}}}{}

\begin{abstract}
Understanding a group conversation means deciding, from video and audio, \emph{who} is speaking and
\emph{to whom}. Leading systems treat these tasks offline on single-channel clips and align models with a
self-evaluation reward. We propose \ours{}, a geometry-first
alternative built on one measured signal: the \emph{direction} from which a microphone array beside the
camera hears a sound. A label-free calibration learned from the stream registers this direction into the camera
image, and an online registry of faces, voices and directions needs no enrollment. The registered direction
serves as (i)~a visual prompt for a vision--language model, a marker on the speaker or an arrow toward an
off-screen source; (ii)~one input of a reliability-weighted fusion with visual active-speaker detection,
for speakers in the same direction from the array; and (iii)~a \emph{verifiable} reward for alignment: the named speaker
must sit where the sound comes from, the confidence must match that check, and always answering ``group'' earns
nothing. Every decision is online, with the lookahead as an explicit parameter. On $9{,}498$
addressee-labelled utterances from three AMI sites and egocentric EasyCom, the direction prompt raises
speaker accuracy from $54\%$ to $74\%$; addressee accuracy moves by about two points, because telling one
addressee from the group depends on the words more than on where people are. For co-located speakers on AVDIAR, visual active-speaker detection holds
$95\%$ where direction alone falls to $59\%$. The verifiable reward beats self-evaluation under GRPO,
reward-ranked fine-tuning and DPO; shrinking the lookahead from $5$\,s to $0.5$\,s costs $4.5$ DER points
on AMI, and learning the registration online costs a dip in the first two minutes before the system
overtakes vision alone. Code and evaluation protocols will be released.
\end{abstract}

\section{Introduction}
Social and assistive systems increasingly operate in rooms with several people talking at once.
Knowing who is speaking, when, and to whom is a prerequisite for natural interaction, yet it remains
hard precisely in the cases that matter: overlapping speech, occlusion, and speakers who step out of
view. Recent audio-visual benchmarks such as AMUSE~\cite{amuse} frame multi-party understanding as a
set of recognition tasks and align a large multimodal model with a reflective, self-evaluation reward.
They make two simplifying assumptions: they operate \emph{offline} on
pre-recorded clips, with access to future context, and they treat audio as a single channel, discarding
the spatial information that a physical microphone array provides.

We take a geometry-first view (Fig.~\ref{fig:teaser}). An array beside the camera measures the \emph{direction of arrival} (DOA) of
sound. Registered into the camera image through a calibration $\Pi$, a DOA becomes an image location
$u_t=\Pi(\hat\theta_t)$, or an explicit \emph{off-screen} flag when it falls outside the field of view.
This single, cheap signal is \emph{verifiable}, i.e.\ it can be checked against a measurement, and it does
two things. First, it improves recognition: it
tells a model where the current sound comes from, which is exactly the information vision lacks under
overlap, occlusion, and off-camera speech. Second, because direction is measured rather than asserted,
it defines a reward for model alignment whose terms can be \emph{checked} against the geometry, in
contrast to a model grading its own answers. We call the system \ours{} (Binding Embodied Audio to Regions
for INteraction Grounding): a bearing is a direction, and the array is the agent's own ear.

We make four contributions as follows.
\begin{itemize}
  \item \textbf{Direction as a visual prompt.} Marking the registered direction on the frames of a zero-shot
  vision--language model lifts speaker accuracy from $54\%$ to $74\%$ on $9{,}498$ utterances, and a
  shuffled-direction control confirms that the gain comes from where the direction points. Addressee
  accuracy moves by about two points; we trace the residual to the group-versus-individual decision, which
  is a question of language more than of~\mbox{sensing}.
  \item \textbf{Vision under co-location.} We analyze \emph{when} vision (lip-motion active-speaker detection)
  recovers the speaker where direction fails, i.e.\ when two concurrent speakers share a bearing: on AVDIAR, the
  visual-only rule and a reliability-weighted fusion hold $95\%$ where direction alone falls to $59\%$.
  \item \textbf{A verifiable reward.} We introduce a reward built on the measured direction (direction consistency for the
  speaker, a calibration penalty on the stated confidence, and an addressee term that no constant answer
  can exploit) and show that it beats a self-evaluation reward under GRPO, reward-ranked fine-tuning and DPO.
  \item \textbf{Online inference.} We characterize the online cost: how much accuracy is lost when the system
  must decide causally, as a function of lookahead, and when it must learn the array--camera registration
  from the stream.
\end{itemize}

\section{Related Work}

Benchmarks that reason jointly over audio and video in multi-speaker conversation define tasks such as
speaker association, temporal grounding and next-speaker prediction; AMUSE~\cite{amuse} is representative
and aligns a multimodal model with Reflective Reward Optimization (RRO), whose reward combines the model's
own reflective feedback, perceptual-agent consistency scores and task correctness, none of them a physical
measurement. Reinforcement learning with
verifiable rewards has since been applied to the base model we use,
Qwen2.5-Omni~\cite{echoink,avatarrl,omnir1,avreasoner}, but in every case the reward is agreement with a
text label. These systems process pre-recorded, single-channel clips. We instead register an
array's direction into the image, for recognition and as a verifiable reward~\mbox{(Table~\ref{tab:capability})}.

\input{tab_capability}

For \emph{who spoke when} we build on neural diarization (pyannote~\cite{pyannote}, ECAPA-TDNN
embeddings~\cite{ecapa}) and on diart~\cite{diart}, whose lookahead is an explicit parameter that our
online study sweeps. Active-speaker detection associates a visible face with speech from lip and face
motion (TalkNet~\cite{talknet}, LoCoNet~\cite{loconet}; the AVA-ActiveSpeaker protocol~\cite{avaasd});
these cues weaken under overlap, occlusion and off-camera speech~\cite{unitalk}; a face-based detector
cannot represent an off-camera speaker at all, and an array direction can. Direction of arrival from a compact array is classically
estimated with generalized cross-correlation (GCC-PHAT)~\cite{gccphat} and steered response power
(SRP-PHAT)~\cite{srpphat}, both with the phase transform, and spatial audio has recently entered
language models~\cite{bat,sing,phasecoder,directivity,savvy}. None registers a measured direction
into the camera image, uses it with faces for speaker detection, or trains with it.

Who an utterance is addressed to is weakly cued by the previous speaker and more strongly by gaze and
head orientation~\cite{gazeaddressee,mmvap,gazeturn}. LLM addressee recognition has been benchmarked in
multi-party dialogue~\cite{addresseellm,meetingllm} and, for speech addressed to the camera wearer, online
on Ego4D~\cite{onlinettm}; we predict the addressee among all co-present parties.
Streaming video LLMs~\cite{videollmonline,dispider,vispeak} decide when to respond from a growing stream
and online benchmarks~\cite{ovobench} score answers against arrival time; none consumes a microphone array. CHiME-8 MMCSG~\cite{mmcsg} scores speaker-attributed
transcription on smart glasses jointly on error rate and latency. An extended discussion of related work is in the supplementary material.

\section{Method}
At time $t$ the system maintains a conversational state: participant positions and head poses, a partition
of the faces into identities, the current speakers and the addressee. The tasks are speaker detection (assign
the active voice to a visible face or \emph{off-screen}), addressee detection (one participant or
\emph{group}) and diarization (who speaks when) as an online read-out. Decisions are causal: a decision at
$t$ may use evidence up to $t+\Delta$, where the lookahead $\Delta$ is the experimental variable and
$\Delta=\infty$ is the offline reference. We report the online cost as the rise in diarization error rate
(DER) from the longest lookahead we test ($5$\,s) to a shorter one.

\begin{figure*}[t]
\centering
\input{fig_method}
\caption{\textbf{A reward that can be checked.} \textbf{(a)}~The stream yields label-free measurements for each
utterance: the voice's direction from the microphone array, and gaze and wording as evidence $e_t$.
\textbf{(b)}~The policy samples several answers (speaker, addressee, confidence). \textbf{(c)}~Each is scored
against the measurements: the named speaker must sit where the sound comes from, the confidence must match that
check, and the addressee is scored by a frozen critic minus the answer's base rate, so that always answering
``group'' does not pay. On this real training utterance the verifiable reward ranks \FtwoSpka$\,\to\,$\FtwoAdra{} first of
\FtwoN, whereas self-evaluation ranks it last. \textbf{(d)}~Either reward drives the same optimizers, which update
only the LoRA weights (dots: the sampled answers ordered by reward; (\dotreinf{}) reinforced, (\dotunused{})
unused, (\dotdiscour{}) discouraged). Faces~are~blurred~here~only.}
\label{fig:pipeline}
\end{figure*}
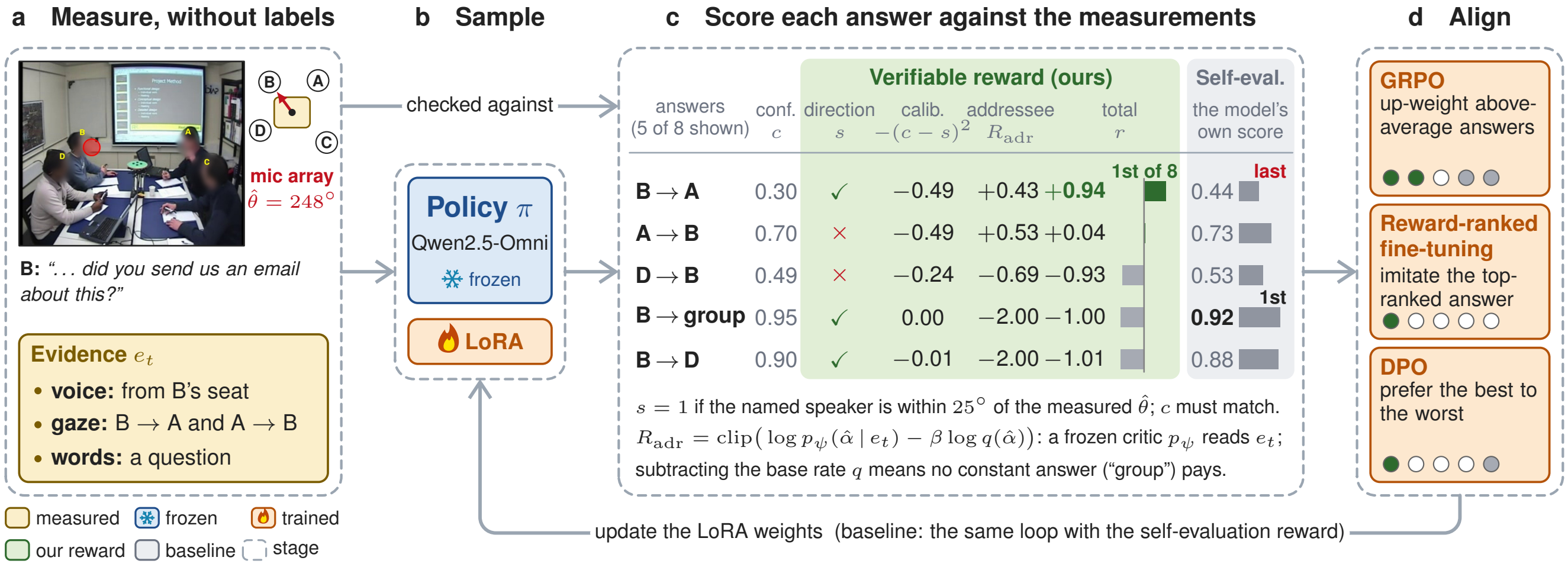

\subsection{Online Registry, Direction and Commitment}
\label{sec:registration}
Camera frames and array signals are processed into face tracks, voice embeddings, acoustic directions,
and speech segments, which update a persistent per-speaker registry; task read-outs and the alignment
reward consume this state (Fig.~\ref{fig:pipeline}).
For each anonymous identity we store a rolling voice embedding (ECAPA~\cite{ecapa}), a face
representation, a seat (the participant's usual face position and direction from the array, from consolidated
face tracks), and the acoustic direction. Identities are created, updated and merged online without prior enrollment.
Speaker detection and diarization are read out from the registry and the registered direction, and the
addressee models take their inputs from the same state, so a single state serves all three.
The online read-out commits labels at time $t$ using lookahead $\Delta$ and does not rewrite them.
Sweeping $\Delta$ yields the latency--accuracy curve of Sec.~\ref{sec:results}, and comparing the committed
transcripts across horizons quantifies what later evidence \emph{would} revise (supplementary material, Table~\STabRevision).

A microphone array yields a direction of arrival $\hat\theta_t$. With an array--camera calibration $\Pi$
it maps to an image coordinate $u_t=\Pi(\hat\theta_t)$, and a source whose $u_t$ lies outside the frame
is flagged \emph{off-screen}. We estimate DOA with SRP-PHAT; in heavy ambient noise (EasyCom's restaurant) plain single-frame SRP is
dominated by competing sources, and we instead use a close-mic-referenced DOA that localizes the target
speaker to a median $7^\circ$ on-camera and $17^\circ$ off-camera. The registration is label-free: each face is matched to the sound direction at which its active-speaker
score rises most, and a direction maps to a pixel by interpolating between these matched pairs, the
\emph{anchors} (supplementary material). On AMI this achieves a median error of $4.3^\circ$ ($5.5$\,px). AMI documents its room geometry; we do not use it, so that the registration is learned as it would be in an unknown scene.
When two concurrent speakers share a bearing, the registered direction cannot separate them. We assign
the active voice to a face by three rules: \emph{direction-only}, the face nearest $u_t$;
\emph{visual-only}, the face with the highest active-speaker score; and a \emph{reliability-weighted
fusion} of the two scores, in which the direction's weight grows with the sharpness of its acoustic peak,
so that vision decides when the peaks of two speakers merge into one broad peak. The study stratifies by the angular
separation $\Delta\theta$ between~concurrent~speakers.

\subsection{Direction Prompt and Evidence Card}
\label{sec:prompt}
The registered direction reaches the model inside the image. On AMI we tile the four close-up cameras
into one frame and draw a red border around the panel of the participant whose seat the sound comes
from; on EasyCom we draw a translucent red circle at $u_t$ in the egocentric frame, or an arrow at the
left or right edge when the source is off-screen. Every participant carries a letter tag, and the
model answers with a speaker, an addressee (a letter or ``group'') and a confidence. An optional
\emph{evidence card} appends, as text, measurements computed causally up to the end of the utterance
and without labels from the meeting at hand: the array's speaker and the previous one, whom the speaker looks at and who looks
back (head pose and gaze placed in one room frame), the final pitch movement, the likely
dialogue-act class of the words, and the one-person-versus-group estimate of the critic below.

\subsection{Verifiable Reward and Alignment}
\label{sec:reward}
A model prediction $\hat y=(\hat\phi,\hat\alpha,c)$ names a speaker, an addressee and a confidence
$c\in[0,1]$. We define the reward
\begin{equation}
\label{eq:reward}
r(\hat y)=s(\hat\phi,\hat\theta_t)-\big(c-s(\hat\phi,\hat\theta_t)\big)^2+R_{\mathrm{adr}}(\hat\alpha),
\end{equation}
in which every term is checkable against a measured quantity, unlike a self-evaluation reward, which is
free to be confidently wrong. It has three parts. (i)~\emph{Direction consistency}: $s=1$ if the
predicted speaker's seat lies within $25^\circ$ of an SRP-PHAT peak during the utterance, and $0$
otherwise. (ii)~\emph{Calibration}: the penalty $(c-s)^2$ makes hedging unprofitable, since a confidence
is rewarded only when it matches what the array measured. (iii)~\emph{Addressee}: a causal critic
$p_\psi(\alpha\mid e_t)$ estimates the addressee from the evidence $e_t$ available at the end of the
utterance (the card's cues). The critic is a gradient-boosted classifier trained on addressee labels
from meetings that are disjoint from the training utterances and from the site under test, so no label
of a rewarded utterance, and no label of the tested site, enters the reward. A critic's raw probability
rewards the majority answer, and a policy trained on it learns to say ``group''. We~therefore~use
\begin{equation}
\label{eq:radr}
R_{\mathrm{adr}}(\alpha)=\mathrm{clip}\big(\log p_\psi(\alpha\mid e_t)-\beta\log q(\alpha),\,-2,\,2\big),
\end{equation}
where $q$ is the critic's average answer distribution over the training utterances. At $\beta=1$ any
constant answer has non-positive expected reward by Jensen's inequality; we use $\beta=0.6$, chosen on
training meetings: there the correct answer earns $-0.25$ on average and always answering ``group'' earns $-0.42$.

We align the policy with three optimizers on the same $594$ AMI training utterances, which carry no
addressee labels: group relative policy optimization (GRPO)~\cite{grpo}, reward-ranked fine-tuning~\cite{raft} (sample eight answers per
utterance and fine-tune on the one the reward ranks first), and direct preference optimization (DPO)~\cite{dpo} on (best, worst) pairs.
The policy sees the frames with the direction marker; untrained, the small model reads the marker poorly
($51.5\%$ speaker accuracy), and the reward teaches it to use it. GRPO with the addressee term is unstable, so
the GRPO comparison uses terms (i)--(ii), the \emph{direction reward}, and keeps the addressee tokens out of
the loss; the two offline optimizers use the full \emph{verifiable reward} with the evidence card in the
prompt. The baseline swaps the reward for our implementation of a self-evaluation reward, after the
self-evaluation part of AMUSE's RRO: the mean of the model's own answer probability, its self-check score
and its stated confidence, on identical data.

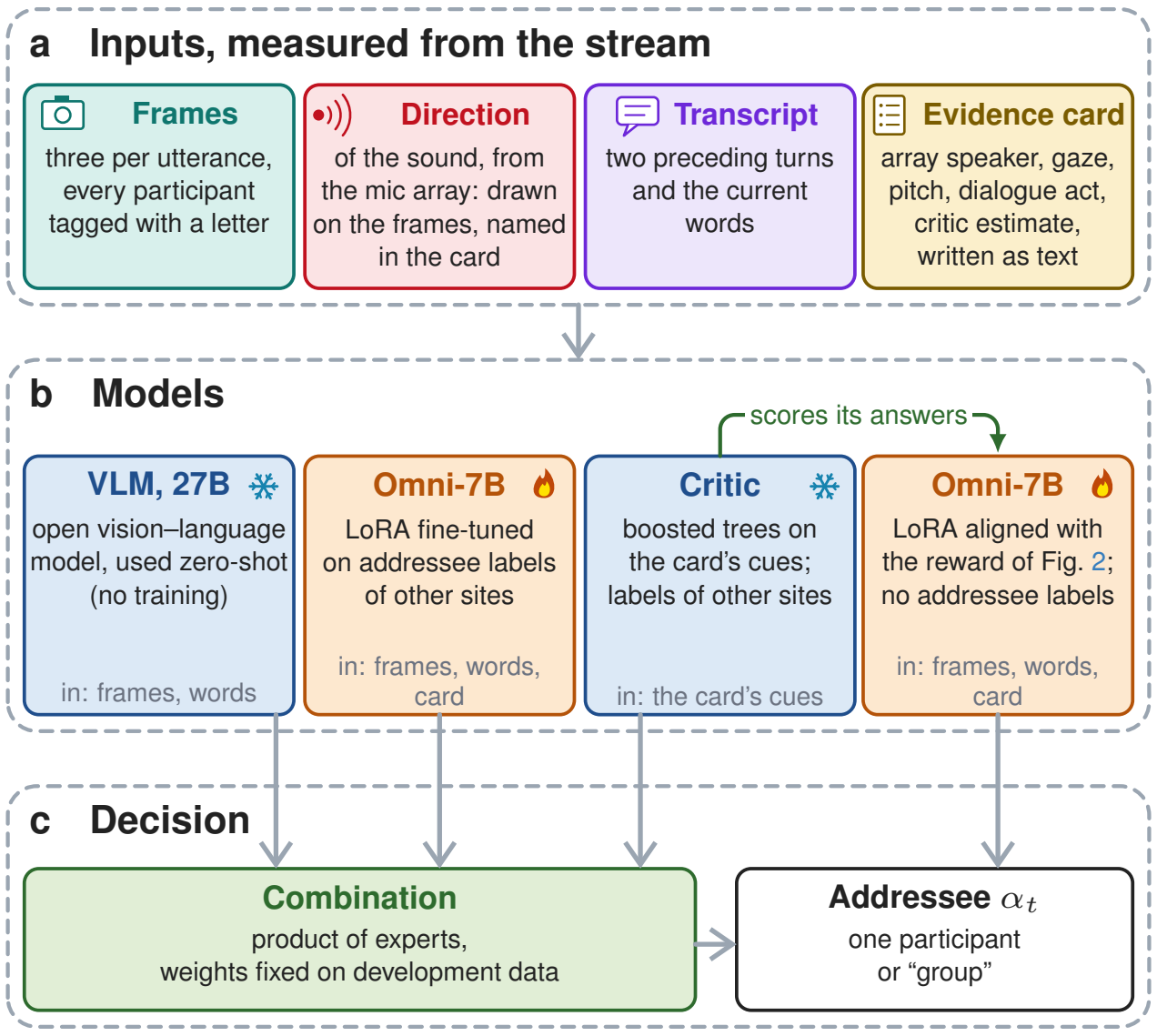
\begin{figure}[t]
\centering
\input{fig_addressee}
\caption{\textbf{Addressee detection: inputs, models and decision.} \textbf{(a)}~All inputs are measured from
the stream: tagged frames, the sound direction (drawn on the frames and named in the card), the transcript and
an evidence card. \textbf{(b)}~Four models estimate the addressee: a frozen zero-shot vision--language model, a
small omni model fine-tuned on other sites' labels, a critic over the card's cues, and the same omni model
aligned with the reward of Fig.~\ref{fig:pipeline}. \textbf{(c)}~The first three are combined as a product of
experts; the reward-aligned policy answers alone. No model sees labels of the site~it~is~tested~on.}
\label{fig:addressee}
\end{figure}

\section{Experimental Setup}

\subsection{Datasets}
We use four public corpora (Table~\ref{tab:datasets}). AMI~\cite{ami} (table-top array, close-up cameras,
manual addressee annotations) and EasyCom~\cite{easycom} (AR glasses with a head-worn array, target-of-speech
labels) serve speaker and addressee detection; AVDIAR~\cite{avdiar} (a fixed camera-and-array device) serves
the co-location study; and AMI with the development set of single-channel VoxConverse~\cite{voxconverse}
($216$ recordings) serves online inference. Concurrent speakers are $\geq 41^\circ$ apart in AMI and a median
$46^\circ$ in EasyCom (never within $15^\circ$), but a median $24^\circ$ in AVDIAR with $17\%$ within
$15^\circ$ (per-recording overlap and faces in view: supplementary material, Fig.~\SFigData). AMI and EasyCom
pair an array and a camera with speaker \emph{and} addressee labels (Table~\ref{tab:capability}), which fixes
the corpora for the prompting and alignment experiments. We use only these public releases and
collect no new data from human participants, so the work required no~institutional~\mbox{review}.

For prompting and alignment we score every addressee-labelled utterance of a multi-speaker
exchange, $9{,}498$ utterances in $29$ meetings or sessions from AMI's Idiap site, its Edinburgh and TNO sites, and
EasyCom (per-site counts in Table~\ref{tab:rq1}). No model or critic is applied to a site whose labels it saw (leave-one-site-out on AMI, held-out
sessions on EasyCom), and all thresholds and weights were fixed in advance on \mbox{development} data.

\begin{table}[t]
\caption{\textbf{The corpora used in this paper and the co-location of their concurrent speakers.} Used for: S,
speaker detection; A, addressee detection; C, co-location; O, online inference. Overlap: share of speech time
with two or more speakers. Both vis.: share of overlap frames with both speakers on camera. $\Delta\theta$:
their angular separation (undefined for single-channel VoxConverse).}
\label{tab:datasets}
\centering
\footnotesize
\setlength{\tabcolsep}{2.6pt}
\resizebox{\linewidth}{!}{%
\begin{tabular}{@{}llccccc@{}}
\toprule
Corpus & Used for & Overlap & Both vis. & Med.\ $\Delta\theta$ & $<15^\circ$ & $<30^\circ$ \\
\midrule
AMI~\cite{ami}                 & S, A, O & 14.6\% & room view & $\geq41^\circ$ & 0\%  & 0\%  \\
EasyCom~\cite{easycom}         & S, A    & 17.7\% & 32\%      & $45.9^\circ$   & 0\%  & 10\% \\
AVDIAR~\cite{avdiar}           & C       & 21.1\% & 99\%      & $24.0^\circ$   & 17\% & 72\% \\
VoxConverse~\cite{voxconverse} & O       & 3.7\%  & --        & --             & --   & --   \\
\bottomrule
\end{tabular}}
\end{table}

\subsection{Protocol and Implementation}
We re-stream each recording and expose only observations up to $t+\Delta$, sweeping $\Delta$ from $0.5$
to $5$\,s and to the offline reference ($\Delta=\infty$).
We report diarization error rate (DER); \emph{speaker accuracy}, the share of utterances whose speaker the
model names correctly; \emph{speaker-to-face recall}, the share of speech segments a rule assigns to the
correct face; addressee accuracy and macro-F1, with balanced accuracy for the group-versus-individual split;
and, for the reward, the area under the ROC curve (AUROC) of reward versus correctness. Confidence intervals
use a paired cluster bootstrap over meetings or sessions ($29$ clusters on the prompting and alignment set, $2{,}000$
resamples, seeds pooled).
The direction prompt is tested zero-shot on an open $27$B vision--language model (Ollama release
\texttt{qwen3.8:27b}, 4-bit quantized), and the aligned policy is
Qwen2.5-Omni-7B~\cite{qwen} with low-rank (LoRA) adapters, trained with three seeds per configuration. All hyper-parameters (supplementary material) were fixed before any test~run.

\section{Results}
\label{sec:results}

\subsection{Direction Improves Speaker Detection}
\input{tab_rq1}

Direction raises speaker accuracy on every dataset. Table~\ref{tab:rq1} reports every
input condition on the full evaluation set, and so doubles as the input ablation. Marking the registered direction on the frames lifts the
zero-shot model from $52.4\%$ to $79.7\%$ on AMI Idiap, from $49.2\%$ to $78.1\%$ on AMI Edinburgh/TNO and
from $59.2\%$ to $66.0\%$ on EasyCom: $+20.6$ points pooled (cluster-bootstrap $95\%$ CI $[15.8,24.5]$),
with the interval above zero on each dataset. The
evidence card adds a further $2.0$ ($+22.6$ over the camera alone, $[17.7,26.4]$).
The gain comes from the direction and not from the marker. Given the direction of another utterance of the
same recording, the model follows it: AMI accuracy falls to $35$--$36\%$, $13$--$17$ points below the
camera alone and $30.2$ points below the true direction pooled ($[22.7,36.7]$). On EasyCom a shuffled
marker leaves accuracy at the camera level ($60.3\%$ against $59.2\%$).

Direction matters most where the frames do not show who is talking. On AMI's close-up panels, where three still frames per utterance show little lip motion,
the camera alone is barely above the transcript ($+8.2$ points on Idiap, $+5.9$ on Edinburgh/TNO) and the
direction adds $27$--$29$; on EasyCom the egocentric view already gives more ($+14.9$) and the direction
adds $6.9$. For
EasyCom speakers outside the field of view, accuracy rises from $32.5\%$ to $35.2\%$ with the edge arrow
and to $48.1\%$ with the card, whose text names the array's speaker. A baseline with no model, which names the
participant seated nearest the registered direction, is correct on $80\%$ of AMI's interactive dialogue acts
(median error $5.7^\circ$), so on AMI the model with the marker matches the geometry and does not exceed it.

The addressee, in contrast, is not a sensing problem. Direction moves addressee accuracy by $1.9$ points
($[0.9,2.9]$) and macro-F1 by $1.8$ ($[0.5,3.1]$); the card leaves accuracy unchanged ($+0.5$,
$[-1.5,3.0]$) and adds $3.5$ points of macro-F1 ($[1.1,6.0]$), and on EasyCom it raises recall on
utterances addressed to the camera wearer from $47.1\%$ to $64.7\%$. The transcript alone already matches
the camera ($+0.7$ points for the transcript, $[-0.1,1.7]$). AMI's manual annotations explain why. Once a single person is addressed,
a rule over turn-taking and attention cues picks the right one $79\%$ of the time. Whether one person or the group is
addressed is the bottleneck: wording, turn structure and even human-annotated gaze reach only
$56$--$63\%$ balanced accuracy, and speakers look at their addressee in $45\%$ of individually
addressed utterances.
Supervision helps a small model but does not pass the zero-shot one: fine-tuning
Qwen2.5-Omni-7B on addressee labels from the other sites brings it level with the zero-shot $27$B model, and
a product-of-experts combination of the two models and the critic adds only $1.7$ points of accuracy
($[0.1,3.5]$; Fig.~\ref{fig:addressee}, details in the supplementary material). On these corpora the
addressee stays near $50$--$60\%$ accuracy for every method we tried.

\subsection{Vision Rescues Co-located Speakers}
Among corpora with clean per-person face labels, the co-located regime exists only in AVDIAR. Two
concurrent speakers at the same bearing never occur among visible speakers in AMI or EasyCom (Table~\ref{tab:datasets}); AVDIAR has them in $17\%$ of
concurrent-speaker frames within $15^\circ$, so it is where the hypothesis can be tested (supplementary
material, Fig.~\SFigColoc).
Direction fails exactly where predicted, and vision closes the gap. With direction only, speaker-to-face recall falls to
$59\%$ (sequence-cluster bootstrap $95\%$ CI $[40,72]$) within $15^\circ$, from $75\%$ at $15$--$30^\circ$;
the visual-only rule recovers the same bin to $95\%$ ($[89,100]$, $n{=}78$ segments), while
the two tie at $74\%$ on single-speaker speech. The
reliability-weighted fusion matches the visual-only rule in every overlap bin, where vision is near ceiling
($94$--$99\%$) with every speaker in view; its value is that it need not know in advance which cue a scene will defeat, and on EasyCom it
is direction that carries the off-camera speakers vision cannot see (Sec.~\ref{sec:where}). Keeping one acoustic peak per concurrent speaker is what makes direction usable under overlap at
all: the strongest peak alone attributes $35\%$ of co-located speech, against $59\%$ with one peak per
concurrent speaker (supplementary material, Table~\STabDoa).

\subsection{Verifiable Reward versus Self-Evaluation}
\begin{figure}[t]
\centering
\includegraphics[width=\linewidth]{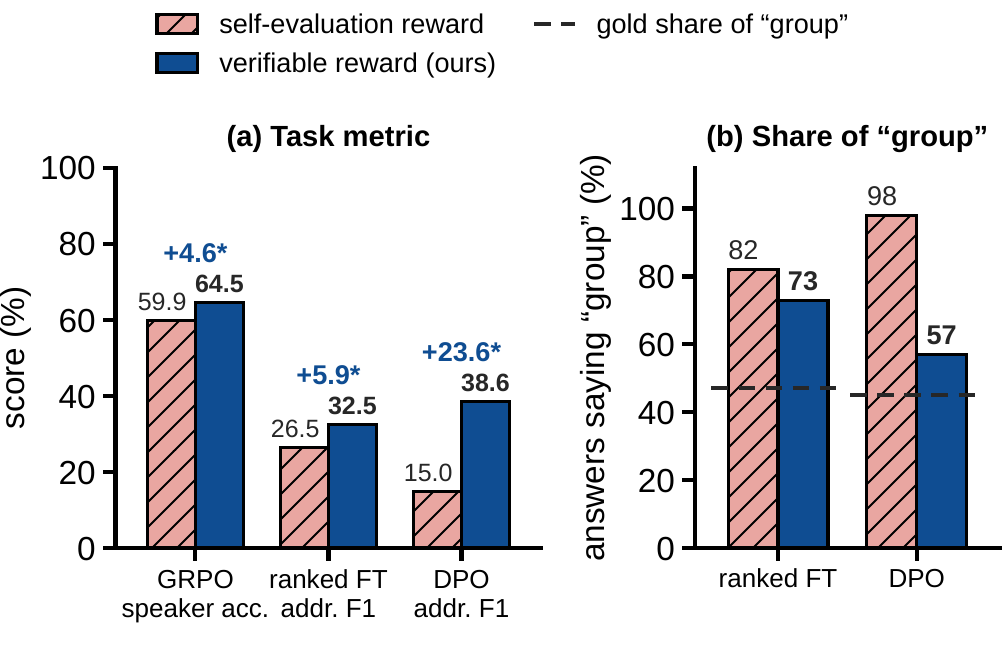}
\caption{\textbf{Verifiable reward versus self-evaluation} (same policy, data and optimizer; three seeds
pooled). \textbf{(a)}~The metric each optimizer is run for: speaker accuracy under GRPO, addressee macro-F1 under
reward-ranked fine-tuning (ranked FT) and DPO, with paired differences in points ($^{*}$: the $95\%$ interval
excludes zero). \textbf{(b)}~Share of answers that name the group, against its share among the gold labels
(dashed). DPO excludes Edinburgh/TNO, whose labels its~reward~critic~saw.}
\label{fig:reward}
\end{figure}

As a proxy for correctness, the direction reward separates correct from
incorrect answers and penalizes confident-wrong predictions, whereas a self-evaluation reward does not:
as a reward-quality probe on held-out AMI, the direction reward's AUROC for flagging a correct answer is
$0.79$ versus $0.57$ for the self-evaluation reward. The self-evaluation reward
also saturates near $1.0$ within a few training steps --- the model rates almost all of its own answers
correct --- so it supplies little learning signal.
For the speaker, we fine-tune the policy with GRPO under each reward on identical AMI data and
test on the full evaluation set, none of whose meetings or participants appear in training
(Fig.~\ref{fig:reward}; all numbers are in the supplementary material, Table~\STabReward). The direction reward lifts pooled speaker accuracy from $51.5\%$ (untrained) to $64.5\%$ ($+13.0$ points,
$[8.5,17.2]$) and beats the self-evaluation reward ($59.9\%$) by $4.6$ points ($[3.8,5.4]$), with
the interval above zero on each dataset. It is also the steadier signal: across three seeds, pooled
accuracy spans $62$--$66\%$ under the direction reward and $54$--$65\%$ under self-evaluation,
which wins one seed of the three. The two rewards also differ in whether the addressee tokens are trained, and a matched control rules this out
as the cause: trained without them, as the direction reward is, self-evaluation falls to $50.7\%$, no better than the untrained policy, and the direction
reward leads it by $13.8$ points ($[12.8,14.9]$; Table~\STabReward).

For the addressee, a speaker-only reward is not enough: it leaves the addressee
untouched, and the GRPO policies above answer ``group'' for $93$--$100\%$ of utterances. With the addressee
term of Eq.~\eqref{eq:radr}, the two offline optimizers train stably and the verifiable reward again beats
self-evaluation. Under reward-ranked fine-tuning it raises addressee macro-F1 by $5.9$ points
($[4.4,7.2]$; the interval excludes zero on each dataset) and accuracy by $1.8$ ($[0.3,3.0]$; no gain
on EasyCom), while answering ``group'' less often than the self-evaluation policy ($66$--$81\%$ against
$80$--$86\%$). Under DPO the self-evaluation policy answers ``group'' for $98\%$ of utterances and ours
keeps a balanced mix ($55$--$58\%$), which is worth $23.6$ points of macro-F1 ($[20.4,26.6]$) and $3.3$
of accuracy ($[-1.8,8.2]$). Speaker accuracy is unchanged on AMI; transferred to EasyCom, the AMI-trained
policy loses $5.4$ points of speaker accuracy under reward-ranked fine-tuning ($[4.0,6.6]$) and none under~DPO.

\subsection{Online Inference: Latency versus Accuracy}
\begin{figure}[t]
\centering
\includegraphics[width=\linewidth]{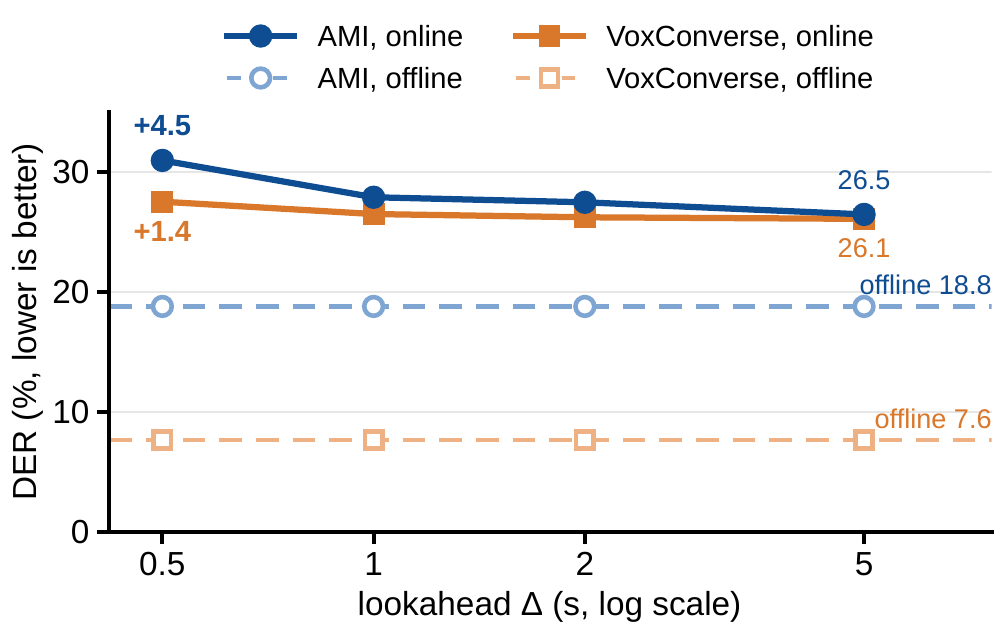}
\caption{\textbf{Online latency--accuracy on AMI and VoxConverse} ($0$\,s collar, overlap scored, so not
comparable to published leaderboards; lower DER is better). The penalty from $5$\,s to $0.5$\,s lookahead
(bold) is larger on overlap-heavy AMI ($+4.5$ points) than on VoxConverse ($+1.4$). Solid: online;
dashed: offline references ($\Delta=\infty$), whose gap reflects the pipeline change, not~lookahead.}
\label{fig:latency}
\end{figure}

Short lookahead costs a few DER points, mostly on overlap. The offline diarizer (pyannote~\cite{pyannote}), a different pipeline, reproduces the
published AMI figure ($18.8\%$ DER); the online diarizer (diart~\cite{diart}) runs at $26.5\%$ with $5$\,s of
lookahead, and its DER rises by $4.5$ points as the lookahead shrinks to $0.5$\,s (Fig.~\ref{fig:latency}). On
VoxConverse~\cite{voxconverse} the rise is gentler, $26.1\%$ to $27.5\%$, consistent with its lower overlap.
Revising committed labels would be safe but would buy little: later evidence breaks only about $5\%$ of
correct commitments, and a $5$\,s horizon removes $15\%$ of the committed errors on AMI against $41\%$
for offline re-clustering (supplementary material, Table~\STabRevision).

\begin{figure}[t]
\centering
\includegraphics[width=\linewidth]{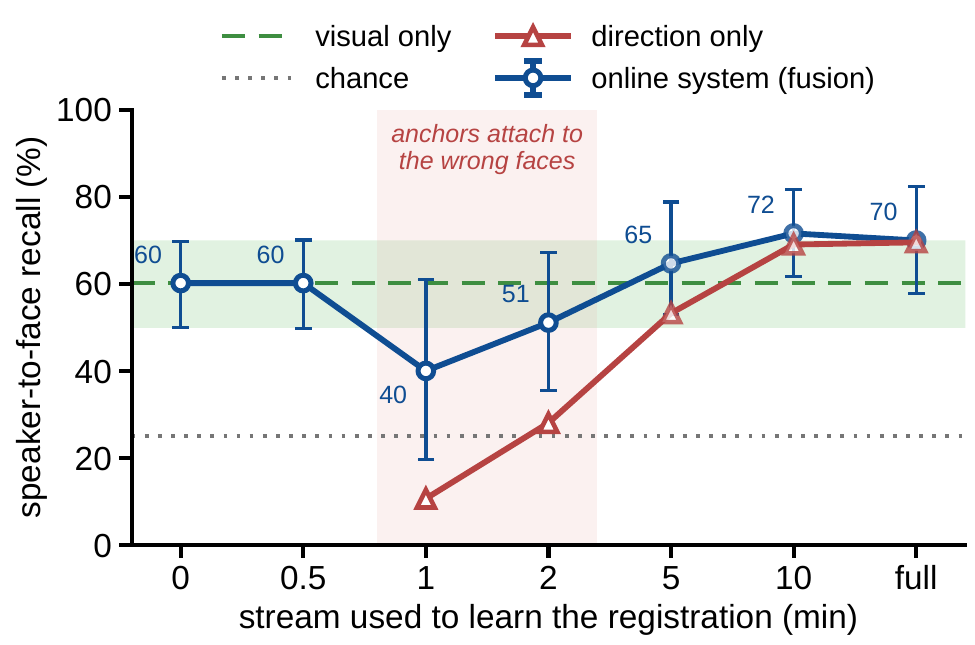}
\caption{\textbf{The cost of learning the registration online.} Speaker-to-face recall on the held-out
second half of eight AMI meetings versus the stream used to learn the array--camera registration
($n{=}13{,}765$ single-speaker blocks, $95\%$ meeting-level bootstrap intervals). The online system uses the
visual-only rule until two anchors exist and the fusion after (green band: vision's interval); it dips below vision
at $1$--$2$\,min (shaded), when the first anchors attach seat directions to the wrong faces, and overtakes it
from $5$\,min (``full'': the offline registration).}
\label{fig:regimes}
\end{figure}

Starting cold has a further cost: the registration must be learned, and its first guess is wrong. When
the array and camera are separate, uncalibrated devices, $\Pi$ must be learned from the stream. We fit the
label-free anchors (Sec.~\ref{sec:registration}) from the first $T$ seconds of each of eight AMI meetings
and score speaker-to-face recall on the held-out second half (Fig.~\ref{fig:regimes}). With no context the system
falls back to the visual-only rule ($60\%$). At $1$\,min, $78\%$ of the anchors attach a correct seat direction to
the wrong face, because only two or three participants have spoken; the fusion then trusts a direction
that points consistently at the wrong person, and the system drops to $40\%$ before recovering to $65\%$ at
$5$\,min and $72\%$ at $10$\,min ($70\%$ offline). A robot can calibrate $\Pi$ once but must still learn which voice, face
and seat belong together, which we do not measure (details and a causal refitting run are in the supplementary~material).

\section{Discussion}
\label{sec:where}
A verifiable geometry signal helps exactly where a single modality fails, and the corpora isolate these
complementary failure modes. In EasyCom, $68\%$ of overlap frames have a concurrent speaker off-camera; the
array flags $78\%$ of them as off-screen, where vision recovers none, so direction rescues vision. In
AVDIAR, concurrent speakers are co-located and visible, so vision rescues direction. In AMI the array
already separates the speakers, and the direction marked in the image helps a vision--language model
wherever the frames do not show who is talking ($+27$--$29$ points on AMI's close-up panels, $+7$ on
EasyCom; Table~\ref{tab:rq1}).

Several limitations remain. Among corpora with clean per-person faces the co-located regime exists only in
AVDIAR, and confirming the effect on a larger far-field corpus such as MISP~\cite{misp} is future work. The
direction reward is only as informative as the array: on the AMI training utterances the measured direction
names the true speaker about $70\%$ of the time. The addressee term of the verifiable reward relies on a critic
trained with labels from other sites, so it is label-free only for the site it is applied to. On the addressee, no method we tried
exceeds a zero-shot $27$B model. AMI Idiap served as our development site; Edinburgh/TNO and EasyCom are
the clean replications. GRPO outcomes vary across seeds, and the verifiable reward costs speaker accuracy when an
AMI-trained policy is transferred to EasyCom with reward-ranked fine-tuning. We have not tested on a
robot (the EasyCom camera wearer stands in for one; Vernissage~\cite{vernissage} is the natural next test),
and the registration learned from the stream
is unreliable in the first two minutes of a meeting (Fig.~\ref{fig:regimes}).

\section{Conclusion}
Acoustic direction, registered into the image, is a cheap and verifiable geometric signal that improves
multi-party speaker detection on fixed and head-worn arrays and grounds model alignment without
self-evaluation; the addressee stays bounded by the decision between one person and the group, which
sensing alone does not resolve. We also show when each cue is needed.

{\small
\bibliographystyle{ieeenat_fullname}
\bibliography{ref}
}

\clearpage
\maketitlesupplementary
\appendix
\renewcommand{\thetable}{S\arabic{table}}\renewcommand{\thefigure}{S\arabic{figure}}
\renewcommand{\theHtable}{S\arabic{table}}\renewcommand{\theHfigure}{S\arabic{figure}}
\setcounter{totalnumber}{2}\setcounter{topnumber}{1}\setcounter{bottomnumber}{1}\setcounter{dbltopnumber}{1}
\renewcommand{\textfraction}{0.2}\renewcommand{\topfraction}{0.8}\renewcommand{\bottomfraction}{0.6}
\renewcommand{\floatpagefraction}{0.8}\renewcommand{\dbltopfraction}{0.8}\renewcommand{\dblfloatpagefraction}{0.8}
\input{supp}

\end{document}

%% file: preamble.tex
\usepackage{booktabs}
\usepackage{pifont}
\usepackage{colortbl}
\definecolor{capyes}{HTML}{2E8B3D}\definecolor{capno}{HTML}{C0392B}\definecolor{caprow}{HTML}{E6EEF8}
\newcommand{\cyes}{\textcolor{capyes}{\ding{51}}}
\newcommand{\cno}{\textcolor{capno}{\ding{55}}}
\definecolor{cappart}{HTML}{D9822B}\newcommand{\cpart}{\textcolor{cappart}{$\circ$}}
\newcommand{\ours}{BEARING}

\usepackage{flafter}
\makeatletter
\AtBeginDocument{\clubpenalty=10000 \@clubpenalty=10000 \widowpenalty=10000 \displaywidowpenalty=10000}
\makeatother

\newcommand{\SFigData}{S1}
\newcommand{\SFigColoc}{S2}
\newcommand{\STabDoa}{S1}
\newcommand{\STabReward}{S2}
\newcommand{\STabRevision}{S3}

\usepackage{tikz}
\usetikzlibrary{positioning,arrows.meta,fit,backgrounds,calc}
\usepackage{pgfplots}
\pgfplotsset{compat=1.18}
\definecolor{figdotours}{HTML}{2E6B2E}\definecolor{figdotbase}{HTML}{A6ACB5}
\DeclareRobustCommand{\dotreinf}{\tikz[baseline=-0.6ex]\draw[draw=black!60, fill=figdotours, line width=0.4pt] (0,0) circle (0.085cm);}
\DeclareRobustCommand{\dotunused}{\tikz[baseline=-0.6ex]\draw[draw=black!60, fill=white, line width=0.4pt] (0,0) circle (0.085cm);}
\DeclareRobustCommand{\dotdiscour}{\tikz[baseline=-0.6ex]\draw[draw=black!60, fill=figdotbase, line width=0.4pt] (0,0) circle (0.085cm);}

\newcommand{\todo}[1]{}
\newcommand{\TODO}[1]{}

%% file: fig1_data.tex
\def\FigTime{424}
\def\FigAz{252}
\def\FigAzOff{1}
\def\FigS{1}
\def\FigText{As soon as you thought to move the remote control you have the light.}
\def\FigTextShort{\ldots\,move the remote control you have the light.}
\def\FigGoldSpk{B}
\def\FigGoldAdr{C}
\def\FigVSpk{C}
\def\FigVAdr{B}
\def\FigDSpk{B}
\def\FigDAdr{C}
\def\FigUx{0.3049}
\def\FigUy{0.5373}
\def\FigXD{0.1199}
\def\FigYD{0.4040}
\def\FigHD{0.0974}
\def\FigXB{0.2475}
\def\FigYB{0.5335}
\def\FigHB{0.0898}
\def\FigXA{0.8723}
\def\FigYA{0.4998}
\def\FigHA{0.0793}
\def\FigXC{0.8134}
\def\FigYC{0.4014}
\def\FigHC{0.1086}
\def\FigPending{0}
\def\FigAccV{54}
\def\FigAccDir{74}
\def\FigDirPts{20.6}
\def\FigGrpoSelf{60}
\def\FigGrpoOurs{65}
\def\FigGrpoPts{4.6}
\def\FigRaftF{5.9}
\def\FigDpoF{23.6}
\def\FigRaftSelfF{0.27}
\def\FigRaftOursF{0.32}
\def\FigDpoSelfF{0.15}
\def\FigDpoOursF{0.39}
\def\FigDpoSelfGrp{98}
\def\FigDpoOursGrp{57}
\def\FigGoldGrp{45}
\def\FigN{9{,}498}

%% file: fig2_data.tex
\def\FtwoN{8}
\def\FtwoRows{5}
\def\FtwoAz{248}
\def\FtwoOff{4}
\def\FtwoSpk{B}
\def\FtwoText{\ldots\,did you send us an email about this?}
\def\FtwoGaze{A}
\def\FtwoGazePct{83}
\def\FtwoMutual{A}
\def\FtwoAct{question}
\def\FtwoBeta{0.6}
\def\FtwoQg{0.49}
\def\FtwoXD{0.191}
\def\FtwoYD{0.422}
\def\FtwoXB{0.281}
\def\FtwoYB{0.551}
\def\FtwoXA{0.752}
\def\FtwoYA{0.555}
\def\FtwoXC{0.836}
\def\FtwoYC{0.390}
\def\FtwoAzD{299}
\def\FtwoAzB{243}
\def\FtwoAzC{217}
\def\FtwoAzA{123}
\def\FtwoSpka{B}
\def\FtwoAdra{A}
\def\FtwoConfa{0.30}
\def\FtwoSa{1}
\def\FtwoCala{$-$0.49}
\def\FtwoRadra{$+$0.43}
\def\FtwoOursa{$+$0.94}
\def\FtwoOursVa{0.937}
\def\FtwoSelfa{0.44}
\def\FtwoSelfVa{0.438}
\def\FtwoRankOursa{1}
\def\FtwoRankSelfa{8}
\def\FtwoSpkb{A}
\def\FtwoAdrb{B}
\def\FtwoConfb{0.70}
\def\FtwoSb{0}
\def\FtwoCalb{$-$0.49}
\def\FtwoRadrb{$+$0.53}
\def\FtwoOursb{$+$0.04}
\def\FtwoOursVb{0.040}
\def\FtwoSelfb{0.73}
\def\FtwoSelfVb{0.725}
\def\FtwoRankOursb{2}
\def\FtwoRankSelfb{4}
\def\FtwoSpkc{D}
\def\FtwoAdrc{B}
\def\FtwoConfc{0.49}
\def\FtwoSc{0}
\def\FtwoCalc{$-$0.24}
\def\FtwoRadrc{$-$0.69}
\def\FtwoOursc{$-$0.93}
\def\FtwoOursVc{-0.929}
\def\FtwoSelfc{0.53}
\def\FtwoSelfVc{0.534}
\def\FtwoRankOursc{4}
\def\FtwoRankSelfc{7}
\def\FtwoSpkd{B}
\def\FtwoAdrd{group}
\def\FtwoConfd{0.95}
\def\FtwoSd{1}
\def\FtwoCald{0.00}
\def\FtwoRadrd{$-$2.00}
\def\FtwoOursd{$-$1.00}
\def\FtwoOursVd{-1.002}
\def\FtwoSelfd{0.92}
\def\FtwoSelfVd{0.922}
\def\FtwoRankOursd{5}
\def\FtwoRankSelfd{1}
\def\FtwoSpke{B}
\def\FtwoAdre{D}
\def\FtwoConfe{0.90}
\def\FtwoSe{1}
\def\FtwoCale{$-$0.01}
\def\FtwoRadre{$-$2.00}
\def\FtwoOurse{$-$1.01}
\def\FtwoOursVe{-1.010}
\def\FtwoSelfe{0.88}
\def\FtwoSelfVe{0.875}
\def\FtwoRankOurse{7}
\def\FtwoRankSelfe{3}

%% file: fig_teaser.tex
\begin{tikzpicture}[x=1cm, y=1cm, font=\sffamily\scriptsize, >={Latex[length=1.7mm, width=1.4mm]},
  line cap=round, line join=round,
  stage/.style={draw=flowc, dash pattern=on 2.5pt off 2pt, line width=0.8pt, rounded corners=6pt},
  flowarrow/.style={draw=flowc, line width=1.1pt, -{Straight Barb[length=1.8mm, width=2.4mm]}},
  title/.style={font=\sffamily\footnotesize\bfseries, anchor=base, inner sep=0pt, text=ink},
  box/.style 2 args={draw=#1, fill=#2, line width=0.7pt, rounded corners=3pt},
  halo/.style={preaction={draw, white, line width=#1}}, halo/.default=2.6pt,
  seat/.style={circle, draw=ink, fill=white, line width=0.5pt, inner sep=0pt, minimum size=3.3mm,
               font=\sffamily\scriptsize\bfseries, text=ink},
  seatc/.style={seat, draw=white, fill=#1, text=white, line width=0.7pt},
  leg/.style={font=\sffamily\scriptsize\bfseries, anchor=west, inner sep=0pt, text=#1},
  ttl/.style={font=\sffamily\small\bfseries, anchor=north west, inner sep=0pt, text=#1},
  big/.style={font=\sffamily\LARGE\bfseries, anchor=north west, inner sep=0pt, text=#1},
  sub/.style={font=\sffamily\scriptsize, text=mute, anchor=north west, inner sep=0pt, align=left},
  row/.style={font=\sffamily\scriptsize, text=ink, anchor=east, inner sep=1.5pt},
]
\definecolor{ink}{HTML}{222222}\definecolor{mute}{HTML}{6B7280}\definecolor{flowc}{HTML}{9AA5B1}
\definecolor{snd}{HTML}{C1121F}                                             %
\definecolor{spk}{HTML}{0F4D92}\definecolor{spkfill}{HTML}{DCE9F6}          %
\definecolor{adr}{HTML}{B45309}\definecolor{adrfill}{HTML}{FDE8CF}          %
\definecolor{adrmark}{HTML}{F59E0B}                                         %
\definecolor{onl}{HTML}{3F4B5B}\definecolor{onlfill}{HTML}{F1F3F5}          %
\definecolor{good}{HTML}{2F7D3B}\definecolor{goodfill}{HTML}{E6F3E4}        %
\definecolor{basfill}{HTML}{F1F3F5}                                         %
\def\Hf{5.6}                                  %
\def\Wi{5.3}\def\Hi{4.336}\def\Yi{1.12}       %
\path (0,\Hf+0.55) -- (17.4,\Hf+0.55);        %
\newcommand{\dumbbell}[8]{%
  \pgfmathsetmacro{\xa}{#7+(#2-#5)/(#6-#5)*#8}\pgfmathsetmacro{\xb}{#7+(#3-#5)/(#6-#5)*#8}%
  \draw[#4!55, line width=2.2pt, line cap=butt] (\xa,#1) -- (\xb,#1);
  \draw[mute, fill=white, line width=0.7pt] (\xa,#1) circle (0.065);
  \fill[#4] (\xb+0.1,#1) -- (\xb-0.07,#1+0.115) -- (\xb-0.07,#1-0.115) -- cycle;   %
  \node[anchor=east, text=mute, inner sep=1.5pt] at (\xa-0.07,#1) {#2};
  \node[anchor=west, font=\sffamily\scriptsize\bfseries, text=#4, inner sep=1.5pt] at (\xb+0.11,#1) {#3};}

\node[title] at (2.8,\Hf+0.25) {a\quad The microphone array hears where};
\draw[stage] (0,0) rectangle (5.6,\Hf);
\begin{scope}[shift={(0.15,\Yi)}]
  \node[anchor=south west, inner sep=0] at (0,0) {\includegraphics[width=\Wi cm]{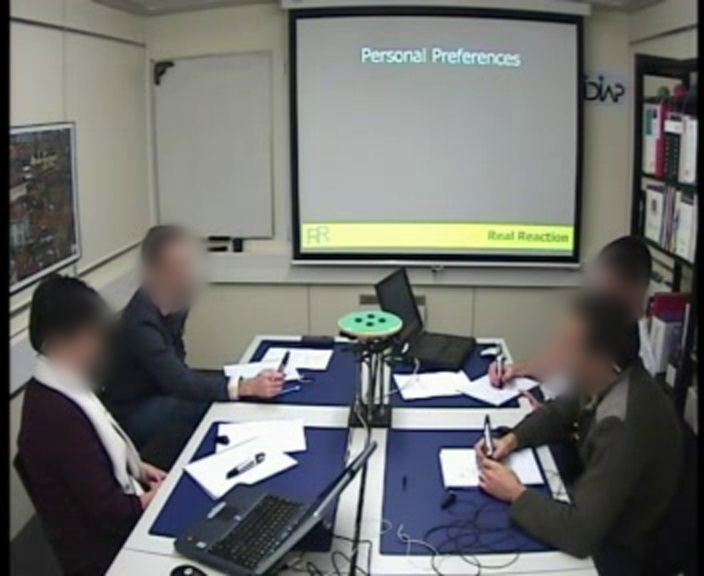}};
  \coordinate (arr) at (0.524*\Wi, 0.441*\Hi);               %
  \coordinate (u) at (\FigUx*\Wi, \FigUy*\Hi);               %
  \coordinate (B) at (\FigXB*\Wi, \FigYB*\Hi); \coordinate (C) at (\FigXC*\Wi, \FigYC*\Hi);
  \coordinate (A) at (\FigXA*\Wi, \FigYA*\Hi); \coordinate (D) at (\FigXD*\Wi, \FigYD*\Hi);
  \draw[white, line width=1pt, preaction={draw, ink, line width=1.8pt}] (arr) circle (0.29);
  \draw[snd, line width=1.3pt, ->, halo, shorten >=0.05cm, shorten <=0.29cm] (arr) -- (u);
  \draw[adrmark, line width=1.3pt, densely dashed, ->, halo] ($(B)+(0.3,-0.58)$) to[bend right=14] ($(C)+(-0.43,-0.2)$);
  \node[seat] at ($(A)+(0.12,0.39)$) {A}; \node[seatc=spk] at ($(B)+(-0.1,0.45)$) {B};
  \node[seatc=adrmark] at ($(C)+(0.05,0.52)$) {C}; \node[seat] at ($(D)+(0,0.46)$) {D};
  \draw[draw=ink, line width=0.5pt] (0,0) rectangle (\Wi,\Hi);
\end{scope}
\node[seatc=spk] at (0.37,0.86) {B}; \node[leg=spk] at (0.65,0.86) {Who is speaking?};
\draw[adrmark, line width=1.3pt, densely dashed, ->] (3.0,0.86) -- (3.47,0.86);
\node[seatc=adrmark] at (3.71,0.86) {C}; \node[leg=adr] at (3.99,0.86) {To whom?};
\draw[snd, line width=1.3pt, ->] (0.17,0.52) -- (0.57,0.52); \node[leg=snd] at (0.65,0.52) {Sound direction};
\draw[ink, line width=0.9pt] (3.71,0.52) circle (0.13); \node[leg=ink] at (3.99,0.52) {Mic array};
\node[anchor=west, inner sep=0pt, text=ink, font=\sffamily\fontsize{6}{7.5}\selectfont] at (0.17,0.19)
  {\textbf{\FigGoldSpk{} says:} \textit{``\FigTextShort''}};

\def\xc{6.3}
\node[title] at (8.25,\Hf+0.25) {b\quad It tells the model who};
\draw[stage] (6.15,0) rectangle (10.35,\Hf);
\foreach \top/\dc/\fc/\tc/\name/\img/\spkr/\adrs/\mark/\word/\mc in {%
    5.45/mute/basfill/ink/Camera only/fig1_in_v/\FigVSpk/\FigVAdr/$\times$/wrong/snd,
    2.75/good/goodfill/good/Camera + sound direction/fig1_in_dir/\FigDSpk/\FigDAdr/\checkmark/correct/good}{
  \draw[box={\dc}{\fc}] (\xc,\top-2.6) rectangle (\xc+3.9,\top);
  \node[font=\sffamily\scriptsize\bfseries, text=\tc, anchor=north west, inner sep=0] at (\xc+0.13,\top-0.14) {\name};
  \node[anchor=north west, inner sep=0] (th\img) at (\xc+0.13,\top-0.52) {\includegraphics[width=2.25cm]{figs/\img.jpg}};
  \draw[draw=ink, line width=0.5pt] (th\img.south west) rectangle (th\img.north east);
  \node[sub, anchor=north, align=center] at (\xc+3.19,\top-0.72) {model says};
  \node[font=\sffamily\Large\bfseries, text=ink, anchor=center, inner sep=0] at (\xc+3.19,\top-1.4)
    {\spkr\,{\normalsize$\rightarrow$}\,\adrs};
  \node[font=\sffamily\scriptsize\bfseries, text=\mc, anchor=center, inner sep=0] at (\xc+3.19,\top-1.92) {\mark\ \word};
}
\draw[snd, line width=1.2pt, ->, halo=2.2pt, shorten >=0.1cm, shorten <=0.12cm]
  ($(thfig1_in_dir.south west)+(0.524*2.25, 0.441*2.25*576/704)$) -- ($(thfig1_in_dir.south west)+(\FigUx*2.25, \FigUy*2.25*576/704)$);

\def\xt{11.05}
\node[title] at (14.15,\Hf+0.25) {c\quad Our four contributions};
\draw[stage] (10.9,0) rectangle (17.4,\Hf);
\pgfmathtruncatemacro\FigDpoSelfPct{round(100*\FigDpoSelfF)}\pgfmathtruncatemacro\FigDpoOursPct{round(100*\FigDpoOursF)}
\tikzset{ctxt/.style={font=\sffamily\scriptsize, inner sep=0pt}, cnum/.style={font=\sffamily\small, inner sep=0pt}}
\newcommand{\card}[5]{%
  \draw[draw=flowc, fill=white, line width=0.6pt, rounded corners=3pt] (\xt,#1-#2) rectangle (\xt+6.2,#1);
  \fill[#3, rounded corners=1pt] (\xt+0.02,#1-#2+0.1) rectangle (\xt+0.09,#1-0.1);
  \node[circle, fill=#3, text=white, inner sep=0pt, minimum size=3.1mm, font=\sffamily\scriptsize\bfseries]
    at (\xt+0.36,#1-0.25) {#4};
  \node[font=\sffamily\footnotesize\bfseries, text=ink, anchor=base west, inner sep=0pt] at (\xt+0.62,#1-0.34) {#5};}
\newcommand{\res}[6]{%
  \node[ctxt, text=ink!80, anchor=base west] at (\xt+0.2,#1) {#2};
  \node[cnum, text=mute, anchor=base] at (\xt+3.15,#1) {#3};
  \draw[#6, line width=1.9pt, line cap=butt, -{Triangle[length=2.1mm, width=2.7mm]}] (\xt+3.62,#1+0.1) -- (\xt+4.3,#1+0.1);
  \node[cnum, font=\sffamily\small\bfseries, text=#6, anchor=base] at (\xt+4.78,#1) {#4};
  \node[fill=#6, text=white, rounded corners=2.2pt, font=\sffamily\scriptsize\bfseries, inner xsep=2.6pt,
    inner ysep=1.7pt, anchor=east, minimum width=0.8cm] at (\xt+6.08,#1+0.1) {#5};}
\newcommand{\rlab}[4]{%
  \node[ctxt, text=mute, anchor=base] at (\xt+3.15,#1) {#2};
  \node[ctxt, font=\sffamily\scriptsize\bfseries, text=#4, anchor=base] at (\xt+4.78,#1) {#3};}
\def\tp{5.45}
\card{\tp}{1.18}{good}{1}{Direction as a visual prompt}
\ifnum\FigPending=1 \node[fill=snd, text=white, font=\sffamily\tiny\bfseries, rounded corners=3pt, inner xsep=3pt,
  inner ysep=1.5pt, anchor=east] at (\xt+6.05,\tp) {baseline rerun pending};\fi
\res{\tp-0.74}{speaker accuracy}{\FigAccV}{\FigAccDir}{+\FigDirPts}{good}
\rlab{\tp-1.04}{camera only}{+ direction}{good}
\def\tp{4.18}
\card{\tp}{1.18}{good}{2}{Vision when voices share a bearing}
\res{\tp-0.74}{speaker-to-face recall}{59}{95}{+36}{good}
\rlab{\tp-1.04}{direction only}{+ vision}{good}
\def\tp{2.91}
\card{\tp}{1.46}{good}{3}{A reward that can be checked}
\res{\tp-0.72}{speaker accuracy}{\FigGrpoSelf}{\FigGrpoOurs}{+\FigGrpoPts}{good}
\res{\tp-1.08}{addressee F1}{\FigDpoSelfPct}{\FigDpoOursPct}{+\FigDpoF}{good}
\rlab{\tp-1.37}{self-evaluation}{measured}{good}
\def\tp{1.36}
\card{\tp}{1.21}{onl}{4}{Online decisions}
\res{\tp-0.74}{error (DER)\,$\downarrow$}{26.5}{31.0}{+4.5}{onl}
\rlab{\tp-1.04}{5\,s lookahead}{0.5\,s}{onl}

\draw[flowarrow] (5.6,\Hf/2) -- (6.15,\Hf/2);
\draw[flowarrow] (10.35,\Hf/2) -- (10.9,\Hf/2);
\end{tikzpicture}

%% file: tab_capability.tex
\begin{table*}[t]
\caption{\textbf{Corpora, benchmarks and the methods built on them}: the modalities and labels a corpus provides, the
inputs and questions of a benchmark, and the modalities and tasks of a method. Online: deciding causally under a
latency budget. Measured reward: aligning the model with a reward computed from a measurement, not from the
model's own judgement or from text labels. For a corpus or benchmark, Online means it defines an online,
latency-scored protocol, and no corpus or benchmark supplies a reward; for a method, Unique spk.\ counts its
evaluation data. \cpart: partial. n.s.: not stated by the
authors. n/a: no speaker task. $^{\dagger}$: as reported by AMUSE~\cite{amuse}. $^{\star}$: as counted by
AVA-AVD~\cite{avaavd}. $^{\Sigma}$: summed over sessions or splits, which the authors do not say are disjoint.
$^{*}$: per clip. $^{\ddagger}$: addressee labels for a subset of the scenario meetings. $^{\S}$:
conversation-group membership, not per-utterance addressee. $^{\P}$: whether the camera~wearer~is~\mbox{addressed}.}
\label{tab:capability}
\centering
\footnotesize
\setlength{\tabcolsep}{4.2pt}
\renewcommand{\arraystretch}{0.94}
\begin{tabular}{@{}l c cccc ccc c@{}}
\toprule
 & & \multicolumn{4}{c}{\textbf{Modalities}} & \multicolumn{3}{c}{\textbf{Tasks}} & \\
\cmidrule(lr){3-6}\cmidrule(lr){7-9}
 & Unique spk. & Mic array & Audio & Visual & Transcript & Speaker & Addressee & Online & Measured reward \\
\midrule
\multicolumn{10}{c}{\textbf{Corpora}} \\
\midrule
\multicolumn{10}{c}{\textcolor{black!65}{Exocentric / in-the-wild video}} \\
AVA-ActiveSpeaker~\cite{avaasd}        & 430$^{\dagger}$  & \cno  & \cyes & \cyes & \cno  & \cyes & \cno  & \cno & \cno \\
AVA-AVD~\cite{avaavd}                  & 1.5k             & \cno  & \cyes & \cyes & \cno  & \cyes & \cno  & \cno & \cno \\
VoxConverse~\cite{voxconverse}         & 800+$^{\dagger}$ & \cno  & \cyes & \cno  & \cno  & \cyes & \cno  & \cno & \cno \\
Friends-MMC~\cite{friendsmmc}          & 18               & \cno  & \cyes & \cyes & \cyes & \cyes & \cno  & \cno & \cno \\
TV-MMPC~\cite{convstructure}           & 156              & \cno  & \cyes & \cyes & \cyes & \cyes & \cyes & \cno & \cno \\
\cmidrule(lr){1-10}
\multicolumn{10}{c}{\textcolor{black!65}{Meetings and table-top conversation}} \\
AMI~\cite{ami}                         & 190              & \cyes & \cyes & \cyes & \cyes & \cyes & \cpart$^{\ddagger}$ & \cno & \cno \\
AVDIAR~\cite{avdiar}                   & 11$^{\star}$     & \cyes & \cyes & \cyes & \cno  & \cyes & \cno  & \cno & \cno \\
MISP 2021/22~\cite{misp}               & 347              & \cyes & \cyes & \cyes & \cyes & \cyes & \cno  & \cno & \cno \\
MISP-Meeting~\cite{misp2025}           & 263              & \cyes & \cyes & \cyes & \cyes & \cyes & \cno  & \cno & \cno \\
CHiME-9 MCoRec~\cite{mcorec}           & 56$^{\Sigma}$     & \cno  & \cyes & \cyes & \cyes & \cyes & \cpart$^{\S}$ & \cno & \cno \\
\cmidrule(lr){1-10}
\multicolumn{10}{c}{\textcolor{black!65}{Egocentric and robot-mounted}} \\
Vernissage~\cite{vernissage}           & 26               & \cyes & \cyes & \cyes & \cno  & \cyes & \cyes & \cno & \cno \\
EasyCom~\cite{easycom}                 & 52$^{\Sigma}$     & \cyes & \cyes & \cyes & \cyes & \cyes & \cyes & \cno & \cno \\
Ego4D AVD + Social~\cite{ego4d}        & n.s.             & \cno  & \cyes & \cyes & \cyes & \cyes & \cpart$^{\P}$ & \cno & \cno \\
CHiME-8 MMCSG~\cite{mmcsg}             & 138$^{\Sigma}$    & \cyes & \cyes & \cyes & \cyes & \cyes & \cno  & \cyes & \cno \\
EgoLife~\cite{egolife}                 & 6                & \cyes & \cyes & \cyes & \cyes & \cyes & \cno  & \cno & \cno \\
\midrule
\multicolumn{10}{c}{\textbf{Benchmarks}} \\
\midrule
AV-SpeakerBench~\cite{avspeakerbench}  & 2--5+$^{*}$      & \cno  & \cyes & \cyes & \cno  & \cyes & \cno  & \cno & \cno \\
M3-SLU~\cite{m3slu}                    & 2--8$^{*}$       & \cno  & \cyes & \cno  & \cyes & \cyes & \cno  & \cno & \cno \\
AMUSE benchmark~\cite{amuse}           & 350+             & \cno  & \cyes & \cyes & \cyes & \cyes & \cno  & \cno & \cno \\
\midrule
\multicolumn{10}{c}{\textbf{Methods}} \\
\midrule
AV-CONV~\cite{avconv}                  & 50               & \cyes & \cyes & \cyes & \cno  & \cyes & \cyes & \cno  & \cno \\
QuAVF (Ego4D TTM)~\cite{quavf}         & n.s.             & \cno  & \cyes & \cyes & \cno  & \cno  & \cpart$^{\P}$ & \cno  & \cno \\
Online TTM~\cite{onlinettm}            & n.s.             & \cno  & \cyes & \cyes & \cyes & \cpart & \cpart$^{\P}$ & \cyes & \cno \\
Vernissage AE~\cite{mazzola}           & 20               & \cno  & \cno  & \cyes & \cno  & \cno  & \cyes & \cpart & \cno \\
AMUSE / RAFT~\cite{amuse}              & 350+             & \cno  & \cyes & \cyes & \cyes & \cyes & \cno  & \cno  & \cno \\
RL on Qwen2.5-Omni~\cite{echoink,omnir1} & n/a            & \cno  & \cyes & \cpart & \cno  & \cno  & \cno  & \cno  & \cno \\
\rowcolor{caprow}
\textbf{\ours{} (Ours)}                & 84$^{\Sigma}$     & \cyes & \cyes & \cyes & \cyes & \cyes & \cyes & \cyes & \cyes \\
\bottomrule
\end{tabular}
\end{table*}

%% file: fig_method.tex
\begin{tikzpicture}[x=1cm, y=1cm, font=\sffamily\scriptsize, line cap=round, line join=round,
  stage/.style={draw=flowc, dash pattern=on 2.5pt off 2pt, line width=0.8pt, rounded corners=6pt},
  flowarrow/.style={draw=flowc, line width=1.1pt, rounded corners=5pt, -{Straight Barb[length=1.8mm, width=2.4mm]}},
  flowlbl/.style={font=\sffamily\fontsize{6.5}{8}\selectfont, text=ink, fill=white, inner sep=1.5pt},
  box/.style 2 args={draw=#1, fill=#2, line width=0.7pt, rounded corners=3pt},
  title/.style={font=\sffamily\footnotesize\bfseries, anchor=base, inner sep=0pt, text=ink},
  ttl/.style={font=\sffamily\scriptsize\bfseries, anchor=north west, inner sep=0pt, text=#1},
  tny/.style={font=\sffamily\fontsize{6}{7}\selectfont},
  col/.style={tny, text=mute, anchor=north, align=center, inner sep=0pt},
  cell/.style={font=\sffamily\scriptsize, text=ink, anchor=center, inner sep=0pt},
  note/.style={font=\sffamily\fontsize{6.1}{8}\selectfont, text=ink, anchor=west, inner sep=0pt},
]
\definecolor{ink}{HTML}{222222}\definecolor{mute}{HTML}{6B7280}\definecolor{flowc}{HTML}{9AA5B1}
\definecolor{snd}{HTML}{C1121F}                                             %
\definecolor{mea}{HTML}{7A5C00}\definecolor{meafill}{HTML}{FBEFCB}          %
\definecolor{frz}{HTML}{1F4E8C}\definecolor{frzfill}{HTML}{DCE9F6}          %
\definecolor{trn}{HTML}{B45309}\definecolor{trnfill}{HTML}{FDE8CF}          %
\definecolor{our}{HTML}{2E6B2E}\definecolor{ourfill}{HTML}{E1EED6}          %
\definecolor{bas}{HTML}{6B7280}\definecolor{basfill}{HTML}{ECEEF1}          %
\def\Hf{5.0}
\path (0,\Hf+0.55) -- (17.4,\Hf+0.55);     %
\newcommand{\fire}[2]{\begin{scope}[shift={(#1)}, scale=#2]
  \fill[trn!85!red] (0,-0.5) .. controls (-0.6,-0.5) and (-0.55,0.05) .. (-0.16,0.5) .. controls (-0.14,0.25) and (0.08,0.25) .. (0.04,0.78)
    .. controls (0.55,0.35) and (0.6,-0.5) .. (0,-0.5) -- cycle;
  \fill[yellow!85!orange] (0,-0.42) .. controls (-0.3,-0.42) and (-0.28,-0.1) .. (-0.05,0.12) .. controls (-0.02,-0.02) and (0.1,0.0) .. (0.1,0.22)
    .. controls (0.32,0.0) and (0.3,-0.42) .. (0,-0.42) -- cycle;\end{scope}}
\newcommand{\ice}[2]{\begin{scope}[shift={(#1)}, scale=#2]
  \foreach \a in {0,60,...,300}{\draw[frz!70!cyan, line width=0.55pt, line cap=round] (0,0) -- (\a:0.5);
    \draw[frz!70!cyan, line width=0.45pt, line cap=round] (\a:0.3) -- ++(\a+50:0.15) (\a:0.3) -- ++(\a-50:0.15);}\end{scope}}
\newcommand{\yes}{{\color{our}\checkmark}}\newcommand{\no}{{\color{snd}$\times$}}

\node[title] at (1.875,\Hf+0.25) {a\quad Measure, without labels};
\draw[stage] (0,0) rectangle (3.75,\Hf);
\node[anchor=north west, inner sep=0] (im) at (0.15,\Hf-0.15) {\includegraphics[width=2.5cm]{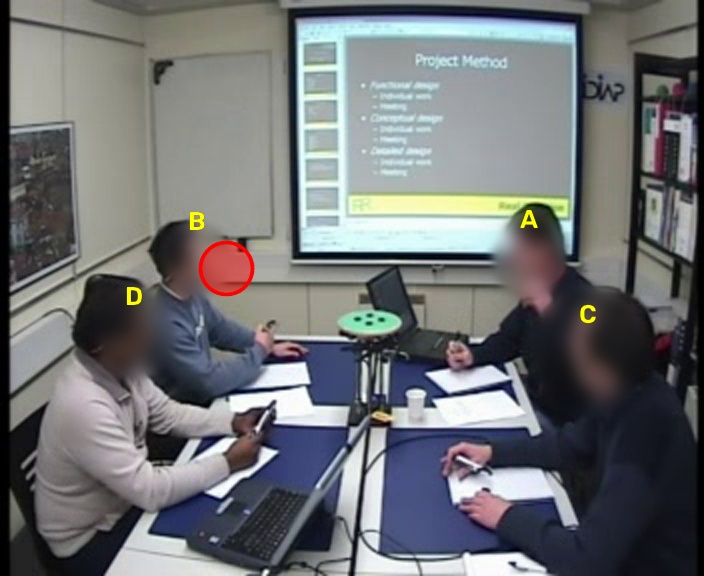}};
\draw[draw=ink, line width=0.5pt] (im.south west) rectangle (im.north east);
\begin{scope}[shift={(3.2,\Hf-0.72)}]
  \draw[draw=mea, fill=meafill, line width=0.5pt, rounded corners=1.5pt] (-0.2,-0.17) rectangle (0.2,0.17);
  \foreach \s/\x/\y in {A/\FtwoXA/\FtwoYA, B/\FtwoXB/\FtwoYB, C/\FtwoXC/\FtwoYC, D/\FtwoXD/\FtwoYD}{
    \coordinate (seat\s) at ({(\x-0.5)*1.15},{(\y-0.47)*4.2});
    \node[circle, draw=ink, fill=white, line width=0.4pt, inner sep=0pt, minimum size=2.4mm,
          font=\sffamily\fontsize{5}{6}\selectfont\bfseries, text=ink] at (seat\s) {\s};}
  \draw[snd, line width=0.9pt, -{Latex[length=1.3mm, width=1.2mm]}, shorten >=1.3mm] (0,0) -- (seat\FtwoSpk);
  \fill[ink] (0,0) circle (0.04);
  \node[tny, text=snd, anchor=north, align=center, inner sep=0] at (0,-0.62) {\textbf{mic array}\\$\hat\theta=\FtwoAz^\circ$};
\end{scope}
\node[note, anchor=north west, text width=3.45cm, align=left] at (0.15,2.62) {\textbf{\FtwoSpk:} \textit{``\FtwoText''}};
\draw[box={mea}{meafill}] (0.15,0.15) rectangle (3.6,1.8);
\node[ttl=mea] at (0.27,1.68) {Evidence $e_t$};
\foreach \y/\k/\v in {1.17/voice/{from \FtwoSpk's seat}, 0.79/gaze/{\FtwoSpk{} $\to$ \FtwoGaze{} and \FtwoMutual{} $\to$ \FtwoSpk},
                      0.41/words/{a \FtwoAct}}{
  \fill[mea] (0.36,\y) circle (0.04);
  \node[font=\sffamily\scriptsize, anchor=west, inner sep=0, text=ink] at (0.5,\y) {\textbf{\k:} \v};}

\node[title] at (5.3,\Hf+0.25) {b\quad Sample};
\draw[stage] (4.35,1.3) rectangle (6.25,3.7);
\draw[box={frz}{frzfill}] (4.5,2.15) rectangle (6.1,3.55);
\node[font=\sffamily\small\bfseries, text=frz, anchor=center] at (5.3,3.2) {Policy $\pi$};
\node[font=\sffamily\fontsize{6.5}{8}\selectfont, text=ink, anchor=center] at (5.3,2.82) {Qwen2.5-Omni};
\ice{4.99,2.42}{0.22}\node[tny, text=frz, anchor=west, inner sep=0] at (5.17,2.42) {frozen};
\node[draw=trn, fill=trnfill, line width=0.7pt, rounded corners=3pt, font=\sffamily\scriptsize\bfseries, text=trn,
      minimum width=1.6cm, minimum height=0.5cm, inner sep=0] at (5.3,1.72) {\hspace{0.32cm}LoRA};
\fire{4.95,1.725}{0.26}

\def\xr{6.85}
\node[title] at (\xr+3.825,\Hf+0.25) {c\quad Score each answer against the measurements};
\draw[stage] (\xr,0) rectangle (\xr+7.65,\Hf);
\fill[ourfill, rounded corners=3pt] (\xr+2.03,1.3) rectangle (\xr+6.27,\Hf-0.12);            %
\fill[basfill, rounded corners=3pt] (\xr+6.35,1.3) rectangle (\xr+7.55,\Hf-0.12);             %
\node[ttl=our, anchor=north] at (\xr+4.16,\Hf-0.22) {Verifiable reward (ours)};
\node[ttl=bas, anchor=north, font=\sffamily\fontsize{6.5}{8}\selectfont\bfseries] at (\xr+6.95,\Hf-0.23) {Self-eval.};
\node[col] at (\xr+0.8,\Hf-0.6) {answers\\(\FtwoRows{} of \FtwoN{} shown)};
\node[col] at (\xr+1.76,\Hf-0.6) {conf.\\$c$};
\node[col] at (\xr+2.47,\Hf-0.6) {direction\\$s$};
\node[col] at (\xr+3.4,\Hf-0.6) {calib.\\$-(c-s)^2$};
\node[col] at (\xr+4.38,\Hf-0.6) {addressee\\$R_{\mathrm{adr}}$};
\node[col] at (\xr+5.6,\Hf-0.6) {total\\$r$};
\node[col] at (\xr+6.95,\Hf-0.6) {the model's\\own score};
\draw[flowc, line width=0.5pt] (\xr+0.12,\Hf-1.17) -- (\xr+7.53,\Hf-1.17);
\foreach \i/\spkr/\adrs/\cf/\sv/\cal/\radr/\ours/\ov/\slf/\slv/\ro/\rs in {%
   0/\FtwoSpka/\FtwoAdra/\FtwoConfa/\FtwoSa/\FtwoCala/\FtwoRadra/\FtwoOursa/\FtwoOursVa/\FtwoSelfa/\FtwoSelfVa/\FtwoRankOursa/\FtwoRankSelfa,
   1/\FtwoSpkb/\FtwoAdrb/\FtwoConfb/\FtwoSb/\FtwoCalb/\FtwoRadrb/\FtwoOursb/\FtwoOursVb/\FtwoSelfb/\FtwoSelfVb/\FtwoRankOursb/\FtwoRankSelfb,
   2/\FtwoSpkc/\FtwoAdrc/\FtwoConfc/\FtwoSc/\FtwoCalc/\FtwoRadrc/\FtwoOursc/\FtwoOursVc/\FtwoSelfc/\FtwoSelfVc/\FtwoRankOursc/\FtwoRankSelfc,
   3/\FtwoSpkd/\FtwoAdrd/\FtwoConfd/\FtwoSd/\FtwoCald/\FtwoRadrd/\FtwoOursd/\FtwoOursVd/\FtwoSelfd/\FtwoSelfVd/\FtwoRankOursd/\FtwoRankSelfd,
   4/\FtwoSpke/\FtwoAdre/\FtwoConfe/\FtwoSe/\FtwoCale/\FtwoRadre/\FtwoOurse/\FtwoOursVe/\FtwoSelfe/\FtwoSelfVe/\FtwoRankOurse/\FtwoRankSelfe}{
  \pgfmathsetmacro{\y}{\Hf-1.6-0.47*\i}
  \node[cell, font=\sffamily\scriptsize\bfseries, anchor=west] at (\xr+0.18,\y) {\spkr\,{\normalfont$\to$}\,\adrs};
  \node[cell, text=mute] at (\xr+1.76,\y) {\cf};
  \node[cell] at (\xr+2.47,\y) {\ifnum\sv=1 \yes\else \no\fi};
  \node[cell] at (\xr+3.4,\y) {\cal};
  \node[cell] at (\xr+4.35,\y) {\radr};
  \ifnum\ro=1 \node[cell, font=\sffamily\scriptsize\bfseries, text=our, anchor=east] at (\xr+5.45,\y) {\ours};
  \else \node[cell, anchor=east] at (\xr+5.45,\y) {\ours};\fi
  \pgfmathsetmacro{\w}{0.26*\ov}
  \pgfmathparse{\ov>0 ? "our" : "bas!60"}\edef\bc{\pgfmathresult}
  \fill[\bc] (\xr+5.87,\y-0.11) rectangle (\xr+5.87+\w,\y+0.11);
  \ifnum\rs=1 \node[cell, font=\sffamily\scriptsize\bfseries, anchor=east] at (\xr+6.88,\y) {\slf};
  \else \node[cell, text=mute, anchor=east] at (\xr+6.88,\y) {\slf};\fi
  \fill[bas!75] (\xr+6.93,\y-0.11) rectangle ({\xr+6.93+0.5*\slv},\y+0.11);
  \ifnum\ro=1 \node[tny, text=our, anchor=west, inner sep=0] at (\xr+5.5,\y+0.23) {\textbf{1st of \FtwoN}};
              \node[tny, text=snd, anchor=east, inner sep=0] at (\xr+7.47,\y+0.23) {\textbf{last}};\fi
  \ifnum\rs=1 \node[tny, text=ink, anchor=east, inner sep=0] at (\xr+7.47,\y+0.23) {\textbf{1st}};\fi
}
\draw[ink!55, line width=0.5pt] (\xr+5.87,\Hf-1.46) -- (\xr+5.87,1.38);
\node[note] at (\xr+0.18,1.0) {$s=1$ if the named speaker is within $25^\circ$ of the measured $\hat\theta$; $c$ must match.};
\node[note] at (\xr+0.18,0.63) {$R_{\mathrm{adr}}=\mathrm{clip}\big(\log p_\psi(\hat\alpha\,|\,e_t)-\beta\log q(\hat\alpha)\big)$:
  a frozen critic $p_\psi$ reads $e_t$;};
\node[note] at (\xr+0.18,0.27) {subtracting the base rate $q$ means no constant answer (``group'') pays.};

\def\xo{15.1}
\node[title] at (\xo+1.15,\Hf+0.25) {d\quad Align};
\draw[stage] (\xo,0) rectangle (\xo+2.3,\Hf);
\foreach \top/\name/\desc/\kind in {4.85/GRPO/{up-weight above-average answers}/1,
    3.25/{Reward-ranked\\fine-tuning}/{imitate the top-ranked answer}/2,
    1.65/DPO/{prefer the best to the worst}/3}{
  \draw[box={trn}{trnfill}] (\xo+0.15,\top-1.5) rectangle (\xo+2.15,\top);
  \node[ttl=trn, text width=1.8cm, align=left] (T) at (\xo+0.25,\top-0.11) {\name};
  \node[font=\sffamily\fontsize{6.5}{7.6}\selectfont, text=ink, anchor=north west, inner sep=0, text width=1.8cm, align=left]
    at ($(T.south west)+(0,-0.06)$) {\desc};
  \foreach \j in {0,...,4}{                    %
    \ifnum\kind=1 \pgfmathparse{\j<2 ? "our" : (\j>2 ? "bas!60" : "white")}\fi
    \ifnum\kind=2 \pgfmathparse{\j<1 ? "our" : "white"}\fi
    \ifnum\kind=3 \pgfmathparse{\j<1 ? "our" : (\j>3 ? "bas!60" : "white")}\fi
    \edef\fc{\pgfmathresult}
    \draw[draw=ink!70, fill=\fc, line width=0.4pt] (\xo+0.38+0.28*\j,\top-1.3) circle (0.085);}
}

\draw[flowarrow] (3.75,2.5) -- (4.35,2.5);
\draw[flowarrow] (6.25,2.5) -- (\xr,2.5);
\draw[flowarrow] (\xr+7.65,2.5) -- (\xo,2.5);
\draw[flowarrow] (3.75,4.35) -- node[flowlbl, pos=0.5] {checked against} (\xr,4.35);
\draw[flowarrow] (\xo+1.15,0) -- (\xo+1.15,-0.42) -- node[flowlbl, pos=0.5] {update the LoRA weights \ (baseline: the same
  loop with the self-evaluation reward)} (5.3,-0.42) -- (5.3,1.3);
\foreach \x/\dc/\fc/\lab in {0.0/mea/meafill/measured, 1.45/frz/frzfill/frozen, 2.75/trn/trnfill/trained}{
  \draw[draw=\dc, fill=\fc, line width=0.6pt, rounded corners=1.5pt] (\x,-0.36) rectangle (\x+0.26,-0.14);
  \node[tny, text=ink, anchor=west, inner sep=0] at (\x+0.33,-0.25) {\lab};}
\foreach \x/\dc/\fc/\lab in {0.0/our/ourfill/{our reward}, 1.45/bas/basfill/baseline}{
  \draw[draw=\dc, fill=\fc, line width=0.6pt, rounded corners=1.5pt] (\x,-0.72) rectangle (\x+0.26,-0.5);
  \node[tny, text=ink, anchor=west, inner sep=0] at (\x+0.33,-0.61) {\lab};}
\ice{1.58,-0.25}{0.15}\fire{2.88,-0.245}{0.14}
\draw[stage, line width=0.6pt, rounded corners=1.5pt] (2.65,-0.72) rectangle (2.91,-0.5);
\node[tny, text=ink, anchor=west, inner sep=0] at (2.98,-0.61) {stage};
\end{tikzpicture}

%% file: fig_addressee.tex
\begin{tikzpicture}[x=1cm, y=1cm, font=\sffamily\scriptsize, line cap=round, line join=round,
  stage/.style={draw=flowc, dash pattern=on 2.5pt off 2pt, line width=0.8pt, rounded corners=6pt},
  flowarrow/.style={draw=flowc, line width=1.0pt, rounded corners=4pt, -{Straight Barb[length=1.6mm, width=2.2mm]}},
  box/.style 2 args={draw=#1, fill=#2, line width=0.7pt, rounded corners=3pt},
  title/.style={font=\sffamily\footnotesize\bfseries, anchor=north west, inner sep=0pt, text=ink},
  hd/.style={font=\sffamily\fontsize{6.6}{7.6}\selectfont\bfseries, anchor=north, inner sep=0pt, text=#1, align=center},
  tx/.style={font=\sffamily\fontsize{5.4}{6.8}\selectfont, anchor=north, inner sep=0pt, text=ink, align=center},
  io/.style={font=\sffamily\fontsize{5.3}{6.2}\selectfont, anchor=south, inner sep=0pt, text=mute, align=center},
]
\definecolor{ink}{HTML}{222222}\definecolor{mute}{HTML}{6B7280}\definecolor{flowc}{HTML}{9AA5B1}
\definecolor{snd}{HTML}{C1121F}
\definecolor{mea}{HTML}{7A5C00}\definecolor{meafill}{HTML}{FBEFCB}
\definecolor{frz}{HTML}{1F4E8C}\definecolor{frzfill}{HTML}{DCE9F6}
\definecolor{trn}{HTML}{B45309}\definecolor{trnfill}{HTML}{FDE8CF}
\definecolor{our}{HTML}{2E6B2E}\definecolor{ourfill}{HTML}{E1EED6}
\newcommand{\fire}[2]{\begin{scope}[shift={(#1)}, scale=#2]
  \fill[trn!85!red] (0,-0.5) .. controls (-0.6,-0.5) and (-0.55,0.05) .. (-0.16,0.5) .. controls (-0.14,0.25) and (0.08,0.25) .. (0.04,0.78)
    .. controls (0.55,0.35) and (0.6,-0.5) .. (0,-0.5) -- cycle;
  \fill[yellow!85!orange] (0,-0.42) .. controls (-0.3,-0.42) and (-0.28,-0.1) .. (-0.05,0.12) .. controls (-0.02,-0.02) and (0.1,0.0) .. (0.1,0.22)
    .. controls (0.32,0.0) and (0.3,-0.42) .. (0,-0.42) -- cycle;\end{scope}}
\newcommand{\ice}[2]{\begin{scope}[shift={(#1)}, scale=#2]
  \foreach \a in {0,60,...,300}{\draw[frz!70!cyan, line width=0.55pt, line cap=round] (0,0) -- (\a:0.5);
    \draw[frz!70!cyan, line width=0.45pt, line cap=round] (\a:0.3) -- ++(\a+50:0.15) (\a:0.3) -- ++(\a-50:0.15);}\end{scope}}

\definecolor{vid}{HTML}{0F766E}\definecolor{vidfill}{HTML}{D8F0EC}          %
\definecolor{dirfill}{HTML}{FBE3E4}                                         %
\definecolor{txt}{HTML}{6D28D9}\definecolor{txtfill}{HTML}{EDE6FB}          %
\draw[stage] (0,5.25) rectangle (8.3,7.4);
\node[title] at (0.15,7.27) {a\quad Inputs, measured from the stream};
\foreach \xa/\xb/\c/\f/\h/\t in {
  0.12/2.08/vid/vidfill/{Frames}/{three per utterance,\\every participant\\tagged with a letter},
  2.16/4.12/snd/dirfill/{Direction}/{of the sound, from\\the mic array: drawn\\on the frames, named\\in the card},
  4.20/6.16/txt/txtfill/{Transcript}/{two preceding turns\\and the current\\words},
  6.22/8.18/mea/meafill/{Evidence card}/{array speaker, gaze,\\pitch, dialogue act,\\critic estimate,\\written as text}}{
  \draw[box={\c}{\f}] (\xa,5.37) rectangle (\xb,6.85);
  \node[hd=\c, font=\sffamily\fontsize{6.2}{7.2}\selectfont\bfseries] at ({(\xa+\xb)/2+0.19},6.73) {\h};
  \node[tx] at ({(\xa+\xb)/2},6.4) {\t};}
\begin{scope}[shift={(0.4,6.63)}]   %
  \draw[vid, fill=white, line width=0.5pt, rounded corners=0.6pt] (-0.15,-0.1) rectangle (0.15,0.1);
  \fill[vid] (-0.06,0.1) rectangle (0.04,0.14);
  \draw[vid, line width=0.5pt] (0,0) circle (0.055);
\end{scope}
\begin{scope}[shift={(2.36,6.63)}]  %
  \fill[snd] (-0.1,0) circle (0.04);
  \foreach \r in {0.09,0.16,0.23}{\draw[snd, line width=0.5pt] (-0.1,0) ++(-40:\r) arc (-40:40:\r);}
\end{scope}
\begin{scope}[shift={(4.58,6.63)}]  %
  \draw[txt, fill=white, line width=0.5pt, rounded corners=1pt] (-0.15,-0.07) rectangle (0.15,0.13);
  \fill[white] (-0.07,-0.07) -- (-0.1,-0.15) -- (0,-0.07) -- cycle;
  \draw[txt, line width=0.5pt, line join=round] (-0.07,-0.07) -- (-0.1,-0.15) -- (0,-0.07);
  \draw[txt, line width=0.4pt] (-0.09,0.065) -- (0.09,0.065) (-0.09,0.0) -- (0.05,0.0);
\end{scope}
\begin{scope}[shift={(6.41,6.63)}]  %
  \draw[mea, fill=white, line width=0.5pt, rounded corners=0.6pt] (-0.11,-0.14) rectangle (0.11,0.14);
  \foreach \y in {0.07,0,-0.07}{\fill[mea] (-0.06,\y) circle (0.013); \draw[mea, line width=0.4pt] (-0.02,\y) -- (0.07,\y);}
\end{scope}
\draw[flowarrow] (4.15,5.25) -- (4.15,4.85);

\draw[stage] (0,2.15) rectangle (8.3,4.85);
\node[title] at (0.15,4.72) {b\quad Models};
\foreach \xa/\xb/\c/\h/\t/\i in {
  0.12/2.08/frz/{VLM, 27B}/{open vision--language\\model, used zero-shot\\(no training)}/{in: frames, words},
  2.16/4.12/trn/{Omni-7B}/{LoRA fine-tuned\\on addressee labels\\of other sites}/{in: frames, words,\\card},
  4.20/6.16/frz/{Critic}/{boosted trees on\\the card's cues;\\labels of other sites}/{in: the card's cues},
  6.22/8.18/trn/{Omni-7B}/{LoRA aligned with\\the reward of Fig.~\ref{fig:pipeline};\\no addressee labels}/{in: frames, words,\\card}}{
  \draw[box={\c}{\c fill}] (\xa,2.27) rectangle (\xb,4.15);
  \node[hd=\c] at ({(\xa+\xb)/2},4.04) {\h};
  \node[tx] at ({(\xa+\xb)/2},3.7) {\t};
  \node[io] at ({(\xa+\xb)/2},2.33) {\i};}
\ice{1.86,3.93}{0.2}\fire{3.9,3.93}{0.17}\ice{5.94,3.93}{0.2}\fire{7.96,3.93}{0.17}
\draw[our, line width=0.8pt, rounded corners=3pt, -{Latex[length=1.4mm, width=1.3mm]}] (5.18,4.15) -- (5.18,4.45) -- (7.2,4.45) -- (7.2,4.17);
\node[font=\sffamily\fontsize{5.5}{6.4}\selectfont, text=our, fill=white, inner sep=1pt] at (6.19,4.45) {scores its answers};

\draw[stage] (0,0) rectangle (8.3,1.75);
\node[title] at (0.15,1.62) {c\quad Decision};
\draw[box={our}{ourfill}] (0.12,0.12) rectangle (5.0,1.15);
\node[hd=our] at (2.56,1.04) {Combination};
\node[tx] at (2.56,0.72) {product of experts,\\weights fixed on development data};
\draw[box={ink}{white}] (5.3,0.12) rectangle (8.18,1.15);
\node[hd=ink] at (6.74,1.04) {Addressee $\alpha_t$};
\node[tx] at (6.74,0.72) {one participant\\or ``group''};
\foreach \x in {1.95,3.14,4.6}{\draw[flowarrow] (\x,2.27) -- (\x,1.15);}
\draw[flowarrow] (7.2,2.27) -- (7.2,1.15);
\draw[flowarrow] (5.0,0.6) -- (5.3,0.6);
\end{tikzpicture}

%% file: tab_rq1.tex
\begin{table*}[t]
\caption{\textbf{Speaker and addressee detection for every input condition}, on the full evaluation set: every addressee-labelled utterance of AMI Idiap ($3{,}483$), AMI Edinburgh/TNO ($2{,}675$) and EasyCom ($3{,}340$). ``off-cam.'' is EasyCom speaker accuracy when the speaker is outside the glasses' field of view. ``Vision'' frames carry no marker; ``marker not explained'' shows the direction's marker without telling the model what it means. The shuffled direction is the marker of another utterance of the same recording. The registered direction raises speaker accuracy on every dataset, whereas the addressee columns barely move and the transcript alone is as good as any visual input. ``Card'' is the evidence card (without the critic's estimate for the 7B model, which is fine-tuned over three seeds); the combination takes its speaker from the 27B model. All values are percentages; best per column in bold.}
\label{tab:rq1}
\centering
\small
\setlength{\tabcolsep}{3.5pt}
\begin{tabular}{@{}lcccccccccc@{}}
\toprule
 & \multicolumn{4}{c}{speaker accuracy ($\uparrow$)} & \multicolumn{3}{c}{addressee accuracy ($\uparrow$)} & \multicolumn{3}{c}{addressee macro-F1 ($\uparrow$)} \\
\cmidrule(lr){2-5}\cmidrule(lr){6-8}\cmidrule(lr){9-11}
input to the model & Idiap & Ed./TNO & EasyCom & off-cam. & Idiap & Ed./TNO & EasyCom & Idiap & Ed./TNO & EasyCom \\
\midrule
\multicolumn{11}{c}{\textbf{Zero-shot 27B vision--language model}} \\
\midrule
Transcript only & 44.2 & 43.3 & 44.3 & 41.9 & 53.7 & \textbf{61.9} & 51.5 & 43.8 & \textbf{51.7} & 41.6 \\
Vision & 52.4 & 49.2 & 59.2 & 32.5 & 53.9 & 61.6 & 49.4 & 42.6 & 48.5 & 42.8 \\
Vision, marker not explained & 75.3 & 72.9 & 60.6 & 31.2 & 56.1 & 61.2 & 50.4 & 46.6 & 48.3 & 43.8 \\
Vision + shuffled dir. & 35.1 & 36.3 & 60.3 & 32.3 & 53.3 & 59.8 & 50.5 & 42.1 & 47.1 & 42.3 \\
Vision + direction & 79.7 & 78.1 & 66.0 & 35.2 & 57.0 & 61.4 & 51.6 & 47.7 & 48.3 & 43.3 \\
Vision + direction + card & \textbf{80.3} & 80.5 & 69.1 & \textbf{48.1} & 56.5 & 56.7 & 52.2 & \textbf{50.6} & 47.6 & \textbf{46.3} \\
\midrule
\multicolumn{11}{c}{\textbf{Qwen2.5-Omni-7B, vision + direction + evidence card}} \\
\midrule
Untrained & 78.9 & \textbf{81.7} & 55.1 & 33.8 & 52.5 & 51.2 & 44.0 & 28.9 & 19.9 & 25.4 \\
Cross-site supervised & 79.4 & 81.1 & \textbf{74.0} & 46.3 & 56.3 & 61.5 & \textbf{52.7} & 40.8 & 46.3 & 42.6 \\
\midrule
\multicolumn{11}{c}{\textbf{Combination}} \\
\midrule
27B + 7B + critic & \textbf{80.3} & 80.5 & 69.1 & \textbf{48.1} & \textbf{58.6} & 60.8 & 49.9 & 50.2 & 43.2 & 36.4 \\
\bottomrule
\end{tabular}
\end{table*}

%% file: supp.tex
\section{Overview and Clarifications}
\label{sec:clarify}
This document supplements the main paper in three ways. It states exactly what each experiment used (this
section). It gives the details needed to reproduce the work (Secs.~\ref{sec:impl} and~\ref{sec:data}). And it
reports further analyses (Secs.~\ref{sec:speaker}--\ref{sec:examples}). Figures~S1--S2 and Tables~S1--S3 are the ones the main paper cites
and keep their numbers; new items are numbered from Figure~S3 and Table~S4 and sit with the text that discusses
them, so the floats do not appear in numerical order.

A video (\texttt{demo\_video.mp4}, $3.5$\,min, with the meeting's audio) accompanies this paper as an
ancillary file. It shows the
cold-start system of Sec.~\ref{sec:online} running on one AMI meeting (IS1008a, room view): the first $150$\,s, then
one minute from 10:00. This is the speaker-to-face system with the label-free registration learned from the stream
(Fig.~\ref{fig:supp_running}), not the vision--language model of the main results table. Each participant carries a
yellow letter (A--D) above the head in the image. The compass at the top left (``array (top view)'') is drawn in the
array's own coordinates, not in the camera's: the red arrow is the measured direction of the current sound with its
azimuth in degrees, and the yellow letters on the rim are the participants whose seat direction the registry has
learned so far, placed at the direction the registry currently assigns to them. Once the registry is right the arrow
points at the letter of the participant the array hears; while it forms, a letter can sit at the wrong direction
(at 1:41, for example, D is still placed at $17^\circ$, next to C, although D speaks from about $292^\circ$; the
refit at 1:45 moves it to $302^\circ$), and only the white corner marks give the truth. Before any seat is learned the compass
says so, and in silence it shows no arrow. In the image, the red disc is that direction registered into the camera
view; it appears once the registry has two anchors and moves as the registry is refitted every $15$\,s from the
stream so far. The green box is the participant the system names as the speaker, and white corner marks show who is
speaking according to the headset transcripts. The status bar gives the elapsed time, the number of anchors, the
rule in use (visual-only until two anchors exist, the fusion after) and the system's speaker against the headset
label, marked correct or wrong; below it a key shows a small copy of every mark with its meaning. In the first minute
the registry is empty or wrong and the system falls back to lip motion; the wrong calls are left in. Faces are
blurred in the video only.

We first clarify seven points of the main paper.
\begin{enumerate}
\item \emph{How the direction is placed on AMI's close-up panels.} On the close-up mosaic the marker needs only
the seat the measured direction points to, not a pixel position. The results table takes these seat directions from
each room's documentation: the published schematic at Idiap, and at Edinburgh and TNO the seat directions surveyed
on four other meetings of the same room (each seat's median speech direction there, from those meetings' speaker
labels; no labels of the meeting at hand are used). The label-free registration of
Sec.~3.1 of the main paper is what places the direction in a room-view image, in the pilot and the online study
(Sec.~5.4 of the main paper). The direction prompt does not depend on the documented geometry: with the
registration's label-free seats in place of the documented ones at Idiap, the zero-shot model reaches $71.7\%$,
$19.3$ points above vision alone ($[12.9,24.7]$). At Edinburgh and TNO the room view ($352{\times}288$ pixels, faces
$10$--$15$ pixels apart) is too coarse for the registration, so the surveyed seats are used~there.
\item \emph{The direction on EasyCom.} The EasyCom rows of the results table use the glasses array alone, steered
over the whole utterance, with no close-talk microphone. The direction referenced to a speaker's close-talk microphone, which the Method
describes, is used only in the off-camera analysis of the main paper's Discussion, detailed in Sec.~\ref{sec:coloc}.
\item \emph{The fusion rule.} Each visible face receives the score $z + \lambda\,r\,(1-d/90^\circ)$ with $\lambda=1$
(Table~\ref{tab:supp_fusion} also tries $0.5$ and $2$), where $z$ is its
active-speaker score averaged over the segment and standardized across the visible faces, $d$ is the angle between
the face and the nearest of the $k$ strongest direction peaks ($k$ concurrent speakers), and $r$ is the ratio of the
weakest to the strongest of those peaks' power ($r=1$ for a single speaker). The ``sharpness'' of the main paper is
this ratio.
\item \emph{What the models are given.} In the speaker and addressee experiments the models receive the reference
transcript of the current utterance and of the two preceding turns, the speaker labels of those preceding turns, and
three frames taken at $20$, $50$ and $80\%$ of the annotated utterance. No speech recognizer and no system
diarization is in that loop. These inputs are the same in every condition, so the comparisons between conditions
are like for~like.
\item \emph{What ``online'' covers.} Lookahead is varied for diarization only, with an audio-only diarizer
(Table~\ref{tab:supp_der}). Speaker and addressee decisions are causal in that they use evidence up to the end of
the utterance; their delay is the model's inference time (Table~\ref{tab:supp_hparams}). The speaker and addressee
experiments use the corpora's participant identities, and the cold-start study uses face tracks consolidated over the
whole recording (Sec.~\ref{sec:impl}). The direction itself can be computed causally. Its noise mask keeps time--frequency bins more than
$10$\,dB above each bin's noise floor, the $20$th percentile of the bin's power; estimated from the past audio only
(updated every $8$\,s), the floor leaves the marker unchanged on $99.6\%$ of AMI utterances and the zero-shot model's
speaker accuracy unchanged ($79.9\%$ against $79.7\%$ at Idiap, $78.1\%$ at Edinburgh/TNO). EasyCom's direction uses
only the audio of the utterance itself.
\item \emph{The self-evaluation baseline} is our own implementation of the self-evaluation part of AMUSE's reward
and not AMUSE's full objective, which also uses perceptual agents and task correctness.
\item \emph{The base-rate weight.} On the training meetings $\beta=0.6$ gives the correct answer $-0.17$ on average
against $-0.48$ for always answering ``group'' (Table~\ref{tab:supp_baserate}), the same ordering as the $-0.25$ and
$-0.42$ the Method quotes, which were measured on the $400$ pilot utterances of the development site.
\end{enumerate}

\section{Method and Implementation Details}
\label{sec:impl}
\input{tab_supp_hparams}
The array--camera registration is label-free and has four steps. (1)~\emph{People}: face tracks are
clustered by identity (mean ArcFace embedding per track), and a person's seat is their most frequent face
position. (2)~\emph{Source directions} come from sound alone, as the peaks of the direction histogram.
(3)~\emph{Face--direction matching}: faces are matched one-to-one to directions by how much each face's
active-speaker score rises when sound comes from that direction; a matched pair is an \emph{anchor}.
(4)~\emph{Mapping}: a direction maps to a pixel by piecewise-linear interpolation between neighbouring
anchors. In the alignment experiments the policy sees the same three frames with the direction marker as the
zero-shot model; GRPO runs without the evidence card, and reward-ranked fine-tuning and DPO with it.

On AMI the direction of an utterance is the strongest peak of a histogram of SRP-PHAT block estimates ($300$--$3500$\,Hz,
low-energy blocks masked) over the utterance, and the participant named by the direction is the one whose seat is
nearest that peak. On EasyCom the direction is the peak of a PHAT-weighted response steered with the measured array
transfer functions of the glasses ($1{,}020$ directions, $200$--$6000$\,Hz) and accumulated over the utterance; it is
drawn as a circle at its projection into the camera image, or as an arrow at the image edge when it falls outside
the field of view. For the reward, the direction is consistent with a named speaker if that speaker's seat lies
within $25^\circ$ of one of up to two peaks of the utterance.

Face tracks are consolidated before use: a single four-person meeting can yield hundreds of track fragments that must be merged to about four identities. Faces use a detector with ArcFace embeddings~\cite{arcface}; active-speaker uses TalkNet~\cite{talknet};
diarization uses pyannote~\cite{pyannote}; DOA uses SRP-PHAT. The direction prompt is tested on an open $27$B
vision--language model zero-shot (\texttt{qwen3.8:27b}, Ollama digest \texttt{22130167c4c2}, listed as \texttt{qwen3.8:27b-mtp-q4\_K\_M}; greedy
decoding) with three frames per
utterance, the two preceding turns and the current words. The aligned policy is Qwen2.5-Omni-7B~\cite{qwen} with rank-$16$ LoRA
adapters on the language model's attention projections: GRPO runs $300$ steps with four samples per
utterance; reward-ranked fine-tuning and DPO run two epochs (DPO $\beta_{\mathrm{DPO}}=0.1$). Each
configuration is trained with three seeds, and all hyper-parameters were fixed before any test run.

The system prompt on AMI reads: \emph{``You analyse recorded meetings. Four participants, A, B, C and D, sit around a
table. Each image shows the four participants' close-up cameras as four panels, in seat order around the table,
taken during the CURRENT utterance. The yellow letter on each panel tells you who that participant is. A red frame
marks the panel of the participant whose seat the CURRENT utterance's sound comes from, measured by a microphone
array on the table. Task: decide who speaks the CURRENT utterance and to whom it is addressed. The addressee is one
participant's letter if the speaker singles out one person (for example answers them, asks them something, or reacts
to what they just said), or `group' if it is addressed to more than one person or to everyone. Also give your
confidence, from 0 to 1, that both fields are right. Answer only with JSON.''} The sentence about the red frame is
omitted in the conditions without the direction, and the EasyCom prompt describes the glasses camera, the red
circle and the edge arrow instead. The user message lists the preceding turns as ``[letter]: words'' and then the
current utterance with its speaker marked unknown. An evidence card, when used, is appended as text; a real one
reads: \emph{``Voice direction (microphone array): the current voice comes from D's seat. Gaze during the utterance:
A $\to$ unclear; B $\to$ C; C $\to$ B; D $\to$ A. Spoke just before (array): C, $0.7$\,s before. The speaker looks
mainly at A ($77\%$ of the time). Looking at the speaker: nobody. Likely act type: backchannel. Addressee estimate from
turn-taking, gaze, wording and intonation: one person $90\%$ (most likely C), the whole group $10\%$.''} The card
therefore names the array's speaker in words, which is why even the untrained 7B model reaches the accuracy of
the direction when the card is present.

The critic behind the last line of the card and behind the addressee term of the reward is a gradient-boosted tree
classifier (depth $3$, $150$ rounds, learning rate $0.05$) over turn-taking, prosody, head-pose and wording features
that are available at the end of the utterance. For policies tested at Idiap it is trained on the six Edinburgh and
TNO meetings ($2{,}675$ utterances), and for policies tested at Edinburgh, TNO and EasyCom on the eleven Idiap
meetings ($3{,}483$); the card on EasyCom uses a critic trained on eight other EasyCom sessions. The DPO policies were
trained with the first critic only, which is why they are not reported at Edinburgh and TNO.
Table~\ref{tab:supp_hparams} lists the training settings and the cost of every run.

\section{Data and Protocol}
\label{sec:data}
\setcounter{figure}{0}
\begin{figure*}[b]
\centering
\includegraphics[width=0.72\textwidth]{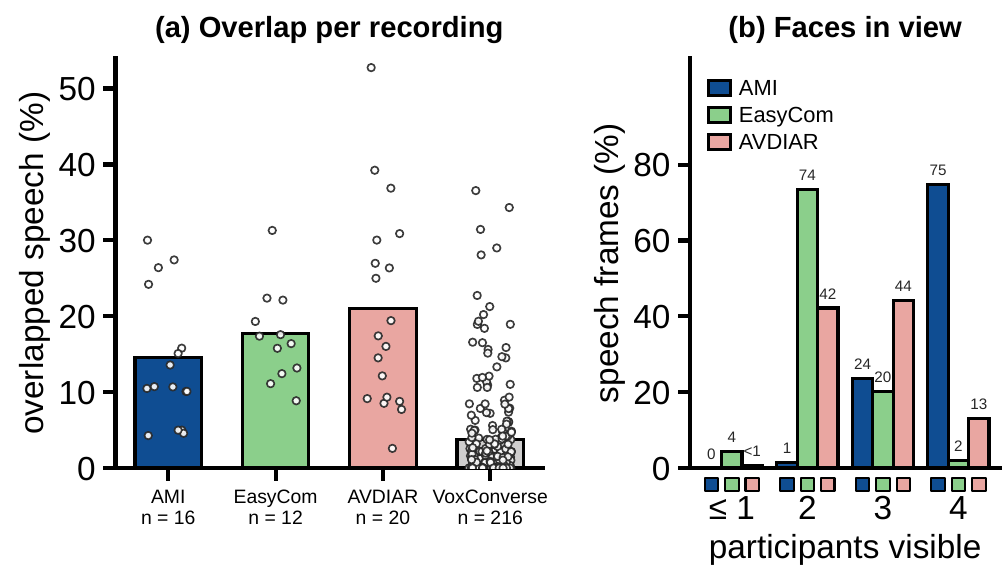}
\caption{\textbf{Dataset statistics.} \textbf{(a)}~Share of speech time with two or more concurrent speakers.
Each dot is one recording ($n$ recordings per corpus: the $16$ meetings of AMI's standard test split used for diarization, the $12$ EasyCom sessions,
the $20$ AVDIAR sequences and the VoxConverse development set); the bar is the pooled ratio reported in
the datasets table of the main paper. One AVDIAR sequence (two speakers talking over each other, $93.5\%$) lies above
the axis and is not drawn; the pooled bar includes it. \textbf{(b)}~Number of participants whose face is visible, over speech frames (zero and one are merged). EasyCom and AVDIAR: annotated
face boxes in the single camera; AMI: faces found in the participants' close-up cameras (head-pose estimates
every $0.5$\,s, $15$ meetings). AMI keeps all four participants in view in $75\%$ of speech frames, the egocentric
EasyCom camera usually sees two, and AVDIAR two to four. VoxConverse is released without~video.}
\label{fig:data}
\end{figure*}
Fig.~\ref{fig:data} complements the datasets table of the main paper with the spread across recordings and
the number of participants a camera sees while someone is speaking.

The corpora in detail. AMI~\cite{ami} provides an 8-microphone table array, multi-camera video, headset ground
truth, and dialogue-act, addressee and focus-of-attention annotations. EasyCom~\cite{easycom} is an AR-glasses
dataset (6-mic array, egocentric wide-FOV video, per-participant close mics, face boxes with identities).
AVDIAR~\cite{avdiar} records social gatherings with a fixed device (two cameras, six microphones, up to five
people), with speaker RTTMs, face boxes with identity, and overlapping speech. VoxConverse~\cite{voxconverse}
is single-channel audio-visual diarization data from online video, with far less overlapped speech.
The co-located regime is \emph{rare}: among corpora with clean on-camera per-person face labels only AVDIAR
provides it, though far-field meeting corpora (MISP~\cite{misp}, participants co-seated several metres from a
6-mic array) and the small audio-visual tracking set CAV3D~\cite{cav3d} contain even tighter bearings with
lower-resolution or sparser faces. MISP~2025~\cite{misp2025} and MMCSG~\cite{mmcsg} also pair an array with a
co-located camera and speaker-attributed labels, but neither has addressee labels; AMI (manual dialogue-act
addressees) and EasyCom (its target-of-speech field) are the public corpora with array, camera, speaker
\emph{and} addressee labels that we could use at this scale.

For prompting and alignment we score every addressee-labelled utterance of a multi-speaker
exchange: $3{,}483$ from $11$ AMI meetings recorded at Idiap, $2{,}675$ from $6$ meetings recorded at
Edinburgh and TNO, and $3{,}340$ from the $12$ EasyCom sessions ($9{,}498$ utterances in $29$ meetings
or sessions). The addressee is one participant or the group. In EasyCom we score the utterances of the
camera wearer's conversation partners, and the wearer, who can be addressed, stands in for an embodied
listener. AMI frames are $2{\times}2$ mosaics of the four close-up cameras; an earlier pilot used $400$
Idiap utterances on the low-resolution room view, which we report alongside. No model or critic is
applied to a site whose labels it saw (leave-one-site-out on AMI, held-out sessions on EasyCom), and
every threshold and weight was fixed on development data. The eleven Idiap meetings are IS1000a, IS1001a--c, IS1003d, IS1006b, IS1006d and IS1008a--d; the six others are
ES2008a--b, ES2009c--d and TS3005a--b. Idiap was our development site: prompts, the layout of the marker, the
contents of the evidence card and the fusion weights of the combination were chosen there, so the Edinburgh, TNO and
EasyCom columns are the ones untouched by development; Table~\ref{tab:supp_seeds} gives the reward comparison per
site. Several AMI meetings of one series share their four participants; our bootstrap resamples
meetings.

\section{Additional Speaker Detection Results}
\label{sec:speaker}
\input{tab_supp_speaker}
The main results table of the main paper doubles as the input ablation. For the speaker each cue adds to the last: the
transcript alone reaches $44\%$, the camera adds $9.9$ points ($[8.2,11.9]$), the registered direction
$20.6$ more and the evidence card a further $2.0$, while a shuffled direction puts AMI $13$--$17$ points
below the camera alone. For the addressee the ordering does not hold: the transcript alone matches every
visual condition. The per-dataset intervals of the direction's gain are $[23.7,30.6]$ (Idiap), $[25.3,31.6]$
(Edinburgh/TNO) and $[5.4,8.1]$ (EasyCom). The model reads the marker even unprompted: shown the true marker
without being told what it means, it reaches $69.5\%$ pooled (row ``marker not explained''), three quarters of
the gain.

Table~\ref{tab:supp_speaker} puts the model next to a rule with no model, on the same $9{,}498$ utterances. Naming the
participant whose seat is nearest the measured direction is right on $73.9\%$ of them ($79.8\%$ at Idiap, $79.2\%$ at
Edinburgh/TNO, $63.6\%$ on EasyCom), against $74.5\%$ for the zero-shot model shown the direction and $76.4\%$ with the
evidence card. A paired cluster bootstrap over meetings puts the model shown the direction level with the rule
($+0.6$ points pooled, $[-1.7,2.9]$) and the full system, with the evidence card, above it: $+2.5$ points pooled
($[1.6,3.7]$), $+5.5$ on EasyCom ($[4.1,6.9]$), $+1.4$ at Edinburgh/TNO ($[1.0,1.7]$) and $+0.6$ at Idiap
($[-0.5,2.0]$). The split in the last columns says where the model's own contribution lies. On AMI the model names the marked participant on $97\%$ of utterances: it is right on
$98.7\%$ of the utterances where the direction is right at Idiap and on $5.0\%$ of those where it is wrong, so with
close-up panels it reads the red frame and takes almost nothing from the pixels. On EasyCom it names the marked
participant on only $54\%$ of utterances and is still right on $55.3\%$ of those where the direction is wrong, close
to the $57.7\%$ it reaches on the same utterances without any marker, so in the egocentric view the image and the direction are both
used and the model ($66.0\%$, and $69.1\%$ with the card) is above the rule ($63.6\%$). The prompt thus hands a
general model the measurement without any training, as the row with the unexplained marker shows, and the model adds
to it where the image carries what the direction lacks, most of all in the egocentric view. (The $80\%$ of the main paper, with a median error of $5.7^\circ$, is
this rule on the $6{,}158$ AMI utterances of the set: $79.5\%$.)

\setcounter{figure}{2}
\begin{figure}[b]
\centering
\includegraphics[width=\linewidth]{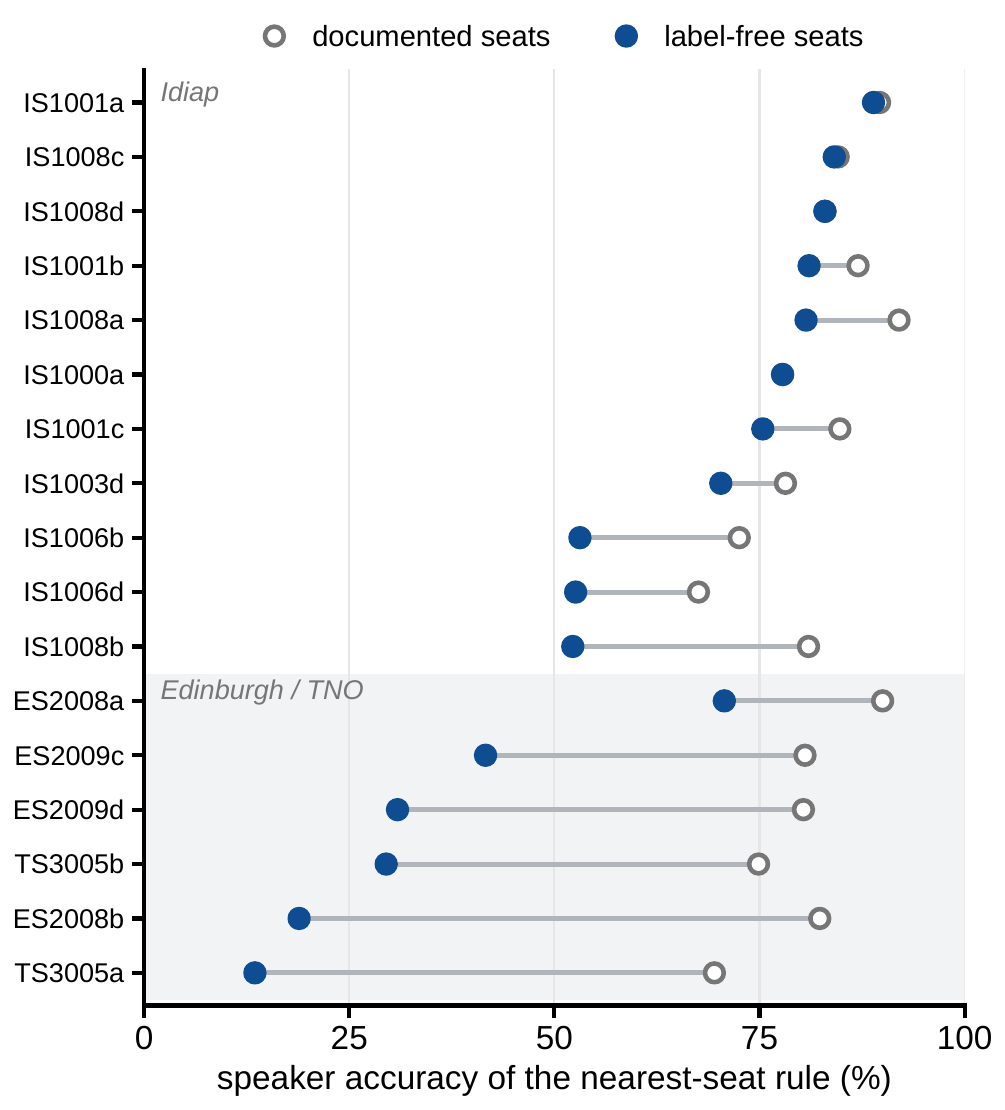}
\caption{\textbf{Documented against label-free seats, meeting by meeting.} Speaker accuracy of the no-model
nearest-seat rule on each AMI evaluation meeting, with the documented seat directions used for the results table
(open) and with the anchors of the label-free registration (filled). The two coincide in most Idiap meetings; at
Edinburgh and TNO the registration does not find the seats in the low-resolution room view.}
\label{fig:supp_registration}
\end{figure}

We also replaced the documented seat directions with those of the label-free registration
(Sec.~\ref{sec:clarify}, item~1). At Idiap the two kinds of seat lie a median $5.8^\circ$ apart and name the same
participant on $86.2\%$ of utterances; in eight of the eleven meetings they agree on at least $85\%$. Shown the
label-free marker, the zero-shot model reaches $71.7\%$, $19.3$ points above vision alone ($[12.9,24.7]$). At
Edinburgh and TNO the room view ($352{\times}288$ pixels, seated faces $10$--$15$ pixels apart) is too coarse for the
anchors, and the surveyed seats are used there. Fig.~\ref{fig:supp_registration} gives the comparison meeting by
meeting, and Fig.~\ref{fig:supp_regexamples} shows the registration at work.

An earlier pilot used $400$ Idiap utterances on the low-resolution AMI room view, scored with McNemar tests
for paired conditions. It agrees with the main result: the camera alone
reaches $46.8\%$ and the registered DOA lifts it to $62.0\%$ ($+15.2$ points, meeting-level bootstrap CI
$[12,19]$, McNemar $p<0.01$; $+17$ under overlap; shuffled control $41.7\%$). Fig.~\ref{fig:supp_pilot} shows utterances of
that pilot with the frame the model received.

\section{Additional Addressee Detection Results}
\label{sec:addressee}
\input{tab_supp_addressee}
Table~\ref{tab:supp_addressee} adds two simple baselines and a label-trained critic. Always answering ``group'' is
right on $47.2\%$ of utterances and naming the previous other speaker on $42.4\%$; the critic alone, a small
classifier over turn-taking, gaze and wording cues, reaches $54.5\%$. A paired cluster bootstrap over meetings
compares the full system, the $27$B model with vision, direction and evidence card, with both rules. Pooled over the
three sites it is $7.8$ points more accurate than always answering ``group'' ($[3.0,12.8]$) and $12.6$ points more
accurate than the previous-speaker rule ($[8.3,17.1]$), and its macro-F1 is $4.4$ points above that rule's
($[0.2,8.1]$), the harder baseline on this measure because it never collapses to one class. The gain is clearest on
EasyCom ($+12.2$ and $+15.0$ points of accuracy, $+10.1$ macro-F1); at Edinburgh/TNO the previous-speaker rule
matches its macro-F1 ($-1.0$, $[-7.4,5.7]$). The input matters less for the addressee than for the speaker: the
transcript alone is within about a point of every richer input, which is the sense in which the main paper calls the
addressee a question of language more than of sensing; the registered direction answers who is speaking, while the
remaining decision, one person or everyone, cannot be read off where people are seated.

Supervision helps a small model but does not pass the zero-shot one: fine-tuning
Qwen2.5-Omni-7B on addressee labels from the other sites ($4{,}000$ utterances per fold) raises its
addressee accuracy by $7.4$ points ($[3.5,11.3]$) and its macro-F1 by $18.5$ ($[15.8,20.7]$), which
brings a $7$B model level with the zero-shot $27$B one. A product-of-experts combination of the two
models and the critic, weighted on development data, adds only $1.7$ points of accuracy over the
zero-shot camera-only model ($[0.1,3.5]$) and no macro-F1 ($-1.4$, $[-4.7,1.0]$).

\section{Verifiable Reward versus Self-Evaluation: Full Results}
\label{sec:supp_reward}
\setcounter{table}{1}
\input{tab_rq3}
Table~\ref{tab:rq3} gives the numbers behind the reward comparison figure of the main paper, per site and
pooled, with the paired differences and their intervals. It also holds the matched control for the GRPO
comparison (row ``Self-eval., masked''): the self-evaluation reward trained with the same loss mask as the
direction reward, which keeps the addressee tokens out of the loss.
The $400$-utterance pilot on the room view agrees: $66\pm1\%$
against $57\pm5\%$ over three seeds, from $48\%$ untrained, with a gold-label oracle at $66\%$.
Adding visual terms (lip motion, listener gaze) to the direction reward does not improve speaker accuracy on the
pilot ($64\%$), so the reward keeps the measured direction as its only speaker term.

\input{tab_supp_seeds}
Table~\ref{tab:supp_seeds} gives the GRPO comparison seed by seed. The direction reward is the steadier of the two
(pooled accuracy $64.5 \pm 2.2\%$ over three seeds against $59.9 \pm 5.3\%$ for self-evaluation and
$50.7 \pm 13.4\%$ for its masked control); self-evaluation wins the second seed ($64.8\%$ against $62.1\%$), as the
main paper notes.

\setcounter{figure}{3}
\begin{figure*}[b]
\centering
\includegraphics[width=\textwidth]{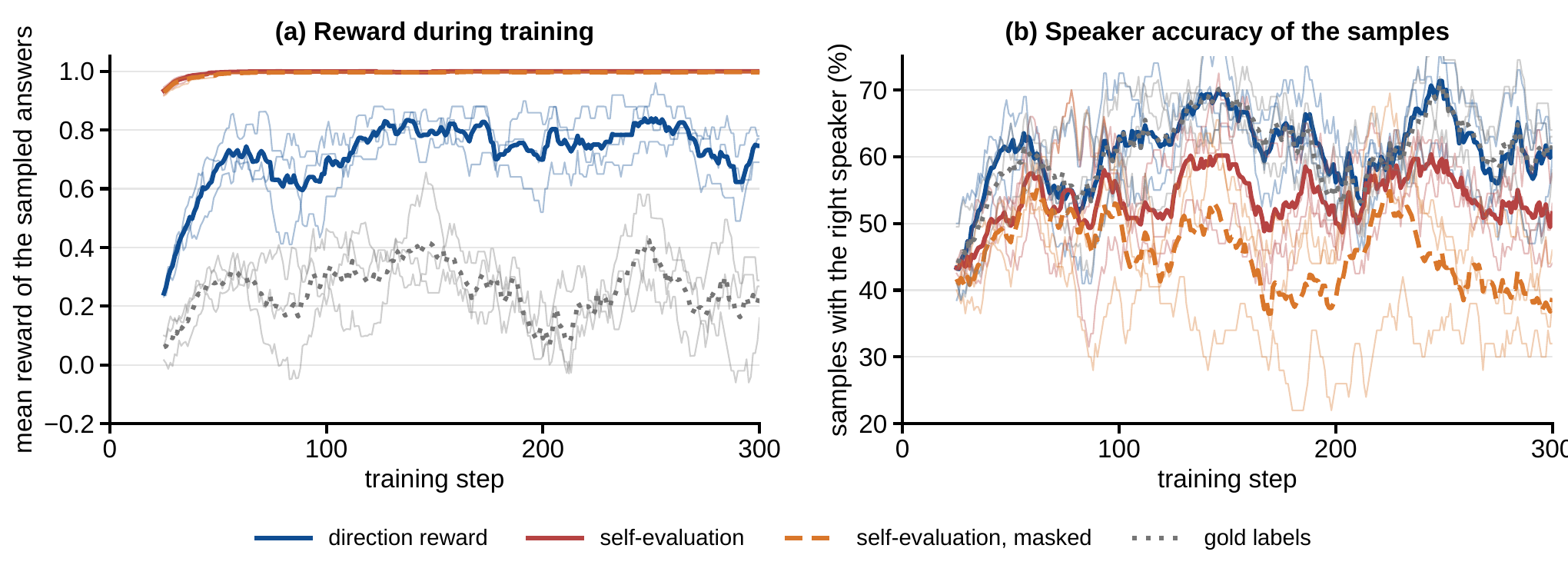}
\caption{\textbf{The GRPO runs themselves.} For each reward, the answers sampled during training (eight per step;
thin lines: the three seeds; thick lines: their mean; $25$-step moving average). \textbf{(a)}~Mean reward the
samples receive; each reward is on its own scale. \textbf{(b)}~Share of the samples that name the true speaker,
which no reward except the gold-label one sees. The self-evaluation reward is near its maximum from the start and
gives every sample the same value, whereas the direction reward grows as the samples improve.}
\label{fig:supp_training}
\end{figure*}

Fig.~\ref{fig:supp_training} shows why, on the training runs themselves. The self-evaluation reward averages $0.93$
over the first $25$ steps and $1.00$ over the last $25$, with or without the mask, so after a few dozen steps every
sampled answer receives the same reward and the update carries no information about the speaker; the accuracy of its
samples then drifts differently in each seed, which is the seed variance of Table~\ref{tab:supp_seeds}. The
direction reward rises from $0.24$ to $0.75$, and the share of samples that name the true speaker rises with it,
from $43\%$ to $61\%$, the same course as under a reward computed from the gold speaker labels ($45\%$ to $62\%$).

Two further runs ask what the direction reward buys (Table~\ref{tab:supp_seeds}, last rows). First, replacing the
measured direction in the reward by the gold speaker label, with the calibration penalty, the masked addressee
tokens and the training data unchanged, gives $64.9 \pm 1.9\%$ pooled over three seeds, against $64.5 \pm 2.2\%$ for
the direction reward (a difference of $-0.4$ points). On these data
the measured direction is as good a training signal for the speaker as the labels it stands in for. Second, we evaluate the policies on frames without
the red marker. The direction-reward policy then reaches $51.8\%$, against $43.8\%$ for the untrained policy
($+8.0$ points, $[5.1,11.5]$), compared with $+13.6$ points for seed~0 when both see the marker. Part of what the reward teaches
is therefore to read the marker, and part carries over to frames without it; the gain without the marker comes
mostly from utterances whose speaker the direction names correctly ($56.5\%$ against $46.5\%$ there, and $38.6\%$
against $36.1\%$ elsewhere), so the policy appears to have learned visual cues of the active speaker that the
direction rewarded during training. The evaluation without the marker is for seed~0 only.

\input{tab_supp_probe}
Table~\ref{tab:supp_probe} gives the reward-quality probe of the main paper for each component. The direction reward
ranks correct answers well when one person speaks ($0.91$) and less well under overlap ($0.65$), where a second
voice also produces a peak; each component of the self-evaluation reward is near chance.
Table~\ref{tab:supp_baserate} shows how the base-rate weight of the addressee term was chosen. Without the weight the
correct answer and the constant answer ``group'' earn almost the same ($-0.85$ and $-0.87$), and with the raw critic
probability on the pilot utterances the constant answer earns more than the correct one ($0.71$ against $0.66$),
which is the collapse the main paper describes; at the chosen weight the correct answer leads both fixed
rules by at least $0.3$.

\setcounter{figure}{1}
\begin{figure}[b]
\centering
\includegraphics[width=\linewidth]{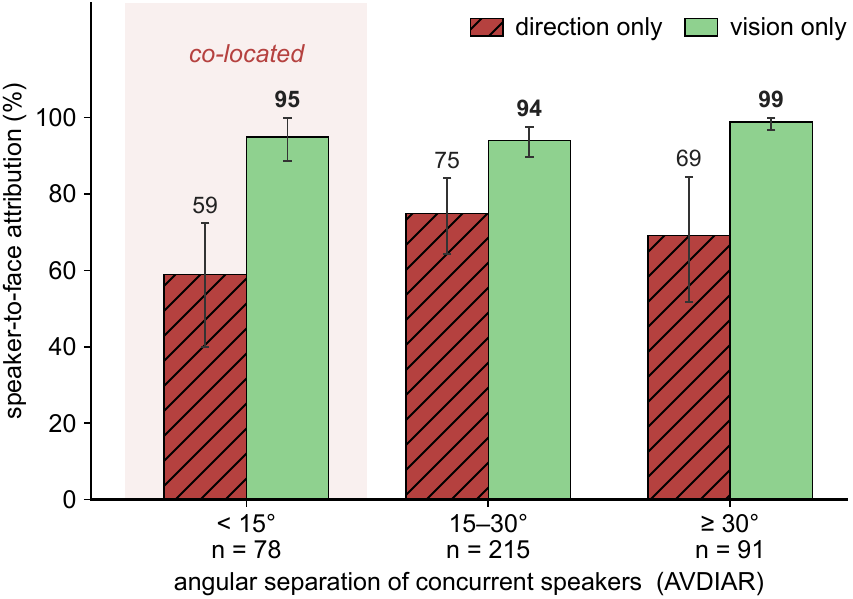}
\caption{\textbf{Vision carries attribution when speakers share a bearing.} Speaker-to-face recall on AVDIAR by
the angular separation of the concurrent speakers ($95\%$ sequence-cluster bootstrap intervals; $n$ counts
segments). Direction-only attribution collapses to $59\%$ when two concurrent speakers are within $15^\circ$
(shaded); visual active-speaker detection recovers it to $95\%$ and holds $94$--$99\%$ in the wider bins.}
\label{fig:coloc}
\end{figure}
\input{tab_supp_baserate}

\section{Co-located Speakers and Direction-of-Arrival Estimation}
\label{sec:coloc}
\setcounter{table}{0}
\input{tab_doa}
Fig.~\ref{fig:coloc} gives the per-bin attribution behind the co-location result of the main paper: direction
alone falls to $59\%$ ($[40,72]$) within $15^\circ$ from $75\%$ at $15$--$30^\circ$, visual active-speaker
detection recovers the co-located bin to $95\%$ ($[89,100]$, $n{=}78$ segments over $12$ sequences), and the
fusion matches vision in every bin.
Multi-source peak-picking is what makes direction usable under overlap. Direction-only
attribution keeps the top-$k$ SRP-PHAT peaks ($k$ = number of concurrent speakers, $15^\circ$ minimum
separation) and assigns each to its nearest visible face. Table~\ref{tab:doa_ablation} compares it with a
single-source variant that keeps only the strongest peak. The two coincide on single-speaker speech
($74\%$), but the single-source rule can attribute at most one of two concurrent speakers and falls to
$35\%$ in the co-located bin, where multi-source recovers $59\%$ ($+24$ points) and $+11$ points overall
($60\%\to71\%$; sequence-cluster CIs $[49,72]$ and $[60,81]$). Even so, the co-located bin stays at
$59\%$: that residual is the gap vision closes (Fig.~\ref{fig:coloc}).

\input{tab_supp_fusion}
Table~\ref{tab:supp_fusion} adds the fusion, which the figure omits, and intervals for every bin. AVDIAR has $20$
sequences and $469$ scored speaker turns; a turn counts as overlapped if another visible speaker's turn intersects
it, and its separation is the smallest angle between the concurrent faces at its middle frame. The visual-only rule
is TalkNet~\cite{talknet} with its released weights. The fusion equals the visual-only rule within the intervals in
every bin, as the main paper reports, and so does every direction weight we tried and a learned fusion; on this corpus
$99\%$ of overlapped speakers are in view. An oracle that takes whichever cue is right reaches $97.4\%$ against
$91.9\%$, so the two cues fail on different segments and leave room for a better fusion. All three rules score about $74\%$ on
single-speaker turns, and the two cues miss different ones: either is right on $94.1\%$ of them.

Two direction estimates are used on EasyCom. The array alone, steered over the whole utterance with the measured
array transfer functions and scored against the motion-capture direction of the speaker on eight sessions, has a
median error of $9.8^\circ$ and is within $15^\circ$ on $62\%$ of $2{,}647$ utterances ($7.4^\circ$ and $75\%$ for a
single speaker, $13.0^\circ$ and $54\%$ under overlap); this is the direction of the prompt experiments. The direction
referenced to the speaker's close-talk microphone has a median error of $7.1^\circ$ for $120$ on-camera speakers
($88\%$ within $15^\circ$). A single $50$\,ms frame of the same array is useless in the restaurant
noise, which is why the off-camera analysis uses the close-talk reference: of $421$ utterances whose speaker has no
face box, $332$ are truly outside the field of view according to the motion-capture poses, and the referenced
direction flags $78.0\%$ of them as off-screen with a median error of $17.4^\circ$, while $94.4\%$ of on-camera
speakers stay inside the field of view. That analysis is an upper bound for a device without close-talk
microphones; with the array alone, speaker accuracy of the nearest-seat rule on EasyCom is $67.4\%$ for on-camera
and $40.8\%$ for off-camera speakers.

\section{Online Inference: Revision and Cold Start}
\label{sec:online}
\setcounter{table}{2}
\input{tab_revision}
Our online read-out never rewrites a committed label, so we measure what revision \emph{would} buy.
Taking the $\Delta{=}0.5$\,s stream as the committed transcript, Table~\ref{tab:revision} reports, for each
longer horizon, how many $10$\,ms frames it would relabel, how many committed errors that fixes and how
many correct commitments it breaks (hypotheses are mapped to reference speakers by the optimal assignment
DER uses; the frame-level error tracks DER to within one point).
Revision is safe but saturates within seconds. Later evidence breaks only about $5\%$ of
correct commitments on AMI and $3\%$ on VoxConverse, whatever the horizon, so committing early costs
little. From $1$\,s to $5$\,s the net share of committed errors removed rises only from $9.7\%$ to
$15.0\%$ on AMI ($3.7\%$ to $5.2\%$ on VoxConverse), which is why the latency--accuracy curve of the main paper
flattens.
The remaining gap to offline is clustering, not lookahead. The offline system removes $41\%$
(AMI) and $72\%$ (VoxConverse) of the committed errors, far beyond any online horizon, because it
re-clusters the whole recording. What online revision can recover tracks overlap ($15\%$ on
overlap-heavy AMI against $5\%$ on VoxConverse); the rest is a global-clustering gap.

The registration must be learned, and its first guess is wrong. The latency curve of the main paper assumes
the array--camera registration $\Pi$ exists. When the array and camera are separate, uncalibrated
devices it must be learned from the stream; AMI publishes its room geometry, which we do not use, so
the meetings stand in for an unknown scene. We fit the label-free anchors (main paper, Method)
from the first $T$ seconds of each of the eight registered meetings and score attribution on the
held-out second half, with the face tracks consolidated over the whole recording and voice embeddings unused, so the curve isolates
the alignment (the cold-start figure of the main paper). With no context, direction cannot be placed in the image and
the system falls back to lip motion ($60\%$). Anchors appear within two minutes ($58\%$ of blocks
covered at $1$\,min, $90\%$ at $2$\,min), but at $1$\,min $78\%$ of them attach a correct seat direction
to the wrong face, because only two or three participants have spoken and the matching between
direction peaks and faces is unconstrained. The fused rule then trusts a direction that points
consistently at the wrong person, and the system drops below vision, to $40\%$, before recovering as
the others take the floor: $65\%$ at $5$\,min and $72\%$ at $10$\,min, against $70\%$ for the offline
registration. A causal run that refits every $15$\,s shows the same course by elapsed time
(Fig.~\ref{fig:supp_running}): $28\%$ in the first $30$\,s, $64\%$ by $2$--$5$\,min and $67\%$ after
$20$\,min, close to the visual-only rule while the registry forms and above it, within overlapping intervals, once most blocks have a
registered direction. Requiring three anchors does not
remove the dip; weighting anchors by their support, or using the fixed circular order of seats around
the table, would, and we leave both to future work. A robot whose array and camera share one body can
calibrate $\Pi$ once; what it must still learn online is which voice, face and seat belong together,
which this experiment does not measure.

\begin{figure}[t]
\setcounter{figure}{4}
\centering
\includegraphics[width=\linewidth]{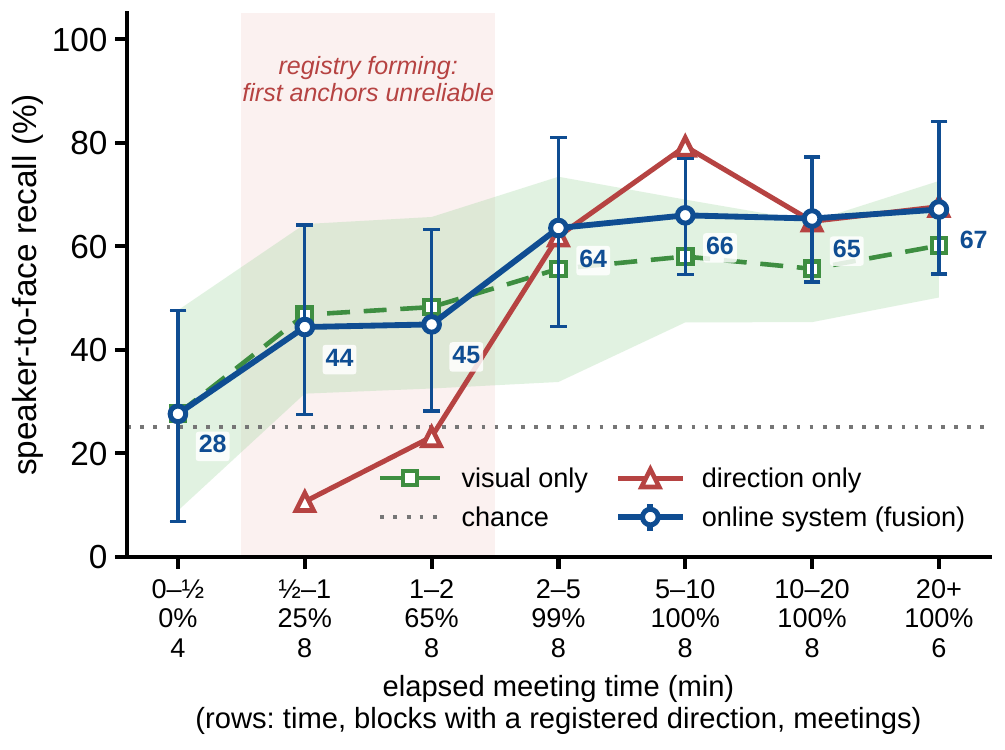}
\caption{\textbf{The causal run of the cold-start study.} Speaker-to-face recall on single-speaker blocks of the
eight registered AMI meetings, binned by elapsed time, with the registration refitted every $15$\,s from the
stream so far ($95\%$ meeting-level bootstrap intervals). Below each bin: the share of blocks with a registered
direction and the number of meetings with speech in that bin (the first $30$\,s hold $87$ blocks from four
meetings). The online system uses the visual-only rule until two anchors exist and the fusion after; direction
alone is drawn where it is defined. The supplementary video shows this run on one meeting.}
\label{fig:supp_running}
\end{figure}

\input{tab_supp_der}
Table~\ref{tab:supp_der} splits the online error into its components. The online diarizer is
diart~\cite{diart} with a $0.5$\,s step on single-channel audio (the headset mix on AMI); it uses neither the
camera nor the array, so the latency study isolates the cost of committing early on the diarization read-out. On AMI a longer lookahead removes speaker confusion ($13.8$ to $10.9$) and false
alarms ($4.9$ to $3.1$) and leaves missed speech unchanged. On VoxConverse the online error is almost entirely
speaker confusion ($22.1$ at $0.5$\,s and $19.7$ at $5$\,s, against $3.1$ offline), which a few seconds of lookahead
cannot repair; this is the clustering gap of the revision analysis above. The total gap between online and offline
operation ($31.0$ against $18.8$ on AMI, $27.5$ against $7.6$ on VoxConverse) is much larger than the $4.5$ and $1.4$
points that lookahead accounts for. The diarizer needs about $0.01$\,s of computation per $0.5$\,s step on AMI and
$0.04$\,s on VoxConverse on one GPU.

\input{tab_supp_coldstart}
Table~\ref{tab:supp_coldstart} gives every run of the cold-start study. The figure of the main paper is the first
block: the eight of the fifteen meetings whose registration passes a quality check on the held-out half (median
placement error below half the spacing between neighbouring faces, scored against the headset labels), scored on the
second half after fitting on a prefix of the first. Without the check, on all fifteen meetings, the system still rises
from $47\%$ with vision only to $53\%$ after ten minutes (column ``all 15''). The run in the caption that refits every $15$\,s from the stream so far is lower in the first
minute ($28\%$ in the first $30$\,s) and reaches the same level afterwards. Waiting for three anchors does not remove
the dip. The intervals are about $\pm10$ points.

\input{tab_supp_runtime}
Table~\ref{tab:supp_runtime} gives the compute cost of each online component, measured on the same meeting as the
cold-start study. Everything up to the language model runs far faster than real time: the direction estimate
costs about $2$\,ms per $250$\,ms block on one CPU thread, a registry refit about $10$\,ms, and the diarizer
$13$\,ms per $0.5$\,s step on the GPU, the same at $0.5$\,s and $5$\,s of lookahead, since lookahead changes only
when a label is committed and not what is computed. The direction, registry and attribution together cost
under $1\%$ of real time on one CPU core, so the online cost reported in the main paper is one of waiting for
evidence, not of computation. The language model is the slow component: its
answer arrives $2$--$3$\,s after the end of an utterance; this is the delay of the speaker and addressee decisions
noted in Sec.~\ref{sec:clarify}. The cached vision components (face detection at $5$\,fps, active-speaker
scoring) took about a fifth and an eighth of the recording's duration when their caches were~built.

\section{Extended Related Work}
This section gives the full discussion that the main paper~summarizes.

\subsection{Audio-Visual Multi-Speaker Understanding}
Benchmarks that jointly reason over audio and video in multi-speaker conversation define tasks such as
audio-visual speaker association, temporal grounding, re-identification, next-speaker prediction, and
summarization; AMUSE~\cite{amuse} is representative and aligns a multimodal model with a
\emph{reflective} reward (RRO) built from the model's own self-evaluation, teacher-guided consistency and task
correctness --- none of its terms is a physical measurement of the scene. Reinforcement learning with verifiable rewards has since been applied to the
same base model we use, Qwen2.5-Omni, with answer-accuracy and format rewards~\cite{echoink,avatarrl,omnir1}
or counts~\cite{avreasoner}; in every case the reward is agreement with a text label. Conversational-role
attribution (speaker, addressee, side-participant) has also been posed for VLMs on television
video~\cite{convstructure}, without spatial sensing, and AMUSE's own data draw on AMI, AVA-ActiveSpeaker
and Friends-MMC~\cite{friendsmmc}. Such systems process pre-recorded clips and treat ``agency'' as
selecting software tools rather than using spatial audio. We instead register a physical array's
direction into the image and use it both for recognition and to ground a verifiable reward (the capability table of the main paper).

\subsection{Diarization, Active Speakers and Localization}
Neural diarization answers \emph{who spoke when}; we use pyannote~\cite{pyannote} as the backbone and
ECAPA-TDNN embeddings~\cite{ecapa} as a voice-identity cue. Online diarization must maintain identities
incrementally, which introduces the identity-fragmentation problem we also observe when consolidating
visual tracks. End-to-end online diarizers now reach competitive error rates at sub-second
latency~\cite{sortformer}; we use diart~\cite{diart} because its lookahead is an explicit parameter,
which our online study sweeps. CineSRD~\cite{cinesrd} recovers off-screen speakers in edited media with an audio
language model; our off-screen flag comes from array geometry.
Associating a visible face with active speech from lip and face motion is the active-speaker task
(TalkNet~\cite{talknet}, LoCoNet~\cite{loconet}; the AVA-ActiveSpeaker protocol~\cite{avaasd}). These
cues weaken under overlap, occlusion, and off-camera speech: in-the-wild benchmarks show the drop with
crowding~\cite{unitalk}, and video-centric localization benchmarks now include off-screen
sources~\cite{avatarbench}, which an image heatmap cannot represent and an array direction can. Berghi
and Jackson~\cite{berghi} learn array-only active-speaker detection and localization from visual
pseudo-labels; we go the other way and use the array's direction to disambiguate faces. Learned voice-to-face association on egocentric or edited video has
been studied with optimal transport (e.g.\ AVA-AVD~\cite{avaavd}); our approach is geometry-grounded
(array direction plus an off-screen flag), which we contrast with rather than replace.

Direction-of-arrival from a compact array is classically estimated with GCC-PHAT~\cite{gccphat} and
SRP-PHAT~\cite{srpphat}. We use SRP-PHAT with a measured array manifold where available, and a
close-mic-referenced variant where ambient noise defeats plain SRP. Placing a DOA in the image requires
an array--camera calibration, for which we use a label-free anchor method. Spatial audio has recently
entered language models: BAT~\cite{bat} reasons over binaural scenes, SING~\cite{sing} and
PhaseCoder~\cite{phasecoder} feed direction-of-arrival or array-agnostic spatial tokens to an LLM,
directional speech LLMs on glasses arrays~\cite{directivity} resolve talkers at $30^\circ$, and
SAVVY~\cite{savvy} answers spatial questions about sound sources in egocentric audio-visual recordings.
None registers a measured direction into the camera image, uses it jointly with faces for
conversational attribution, or uses it as a training reward. On EasyCom, world-locking the multichannel
stream to head orientation~\cite{worldlocking} is a strong published active-speaker localizer.

\subsection{Addressee Estimation and Online Perception}
Who an utterance is addressed to is weakly cued by the previous speaker and strongly cued by gaze and
head orientation~\cite{gazeaddressee}; gaze and head pose remain the dominant turn-taking cue in dyads
and triads~\cite{mmvap,gazeturn}. LLM addressee recognition has been benchmarked in multi-modal
multi-party dialogue~\cite{addresseellm,meetingllm}, and online, frame-level detection of speech
addressed to the camera wearer has been posed on Ego4D~\cite{onlinettm}; we predict the addressee among
all co-present parties. Our prompting result quantifies the split: direction helps the
\emph{speaker} question far more than the \emph{addressee} question, and even AMI's manual
focus-of-attention labels separate group from individual address only weakly (Sec.~5.1 of the main~paper).

Streaming perception makes decisions from incomplete evidence, trading latency for accuracy. Streaming
video LLMs~\cite{videollmonline,dispider,vispeak} decide when to respond from a growing stream, and
online benchmarks~\cite{ovobench} score answers against arrival time, the protocol our latency axis
follows; none consumes a microphone array. CHiME-8 MMCSG~\cite{mmcsg} scores speaker-attributed
transcription on smart glasses jointly on error rate and algorithmic latency. In the
egocentric, AR setting, EasyCom~\cite{easycom} targets the cocktail-party problem with a head-mounted
array and wide-field video; we use it to study off-camera speakers.

Drawing on the image to steer a vision--language model is an established form of prompting: a red circle around a
region directs CLIP-style models to it~\cite{redcircle}, and numbered marks on segmented regions let a model refer to
them~\cite{som}. Our marker is of this family; what differs is its source, a measured acoustic direction, and the
control with a direction borrowed from another utterance. On AVDIAR, Gebru \etal~\cite{avdiar} fuse the microphone
pair of the recording head with face tracks in a Bayesian tracker to diarize the same sequences; we use the corpus
only to stratify attribution by the angular separation of concurrent speakers and do not compare with their
tracker.

\setcounter{figure}{5}
\begin{figure*}[b]
\centering
\includegraphics[width=\textwidth]{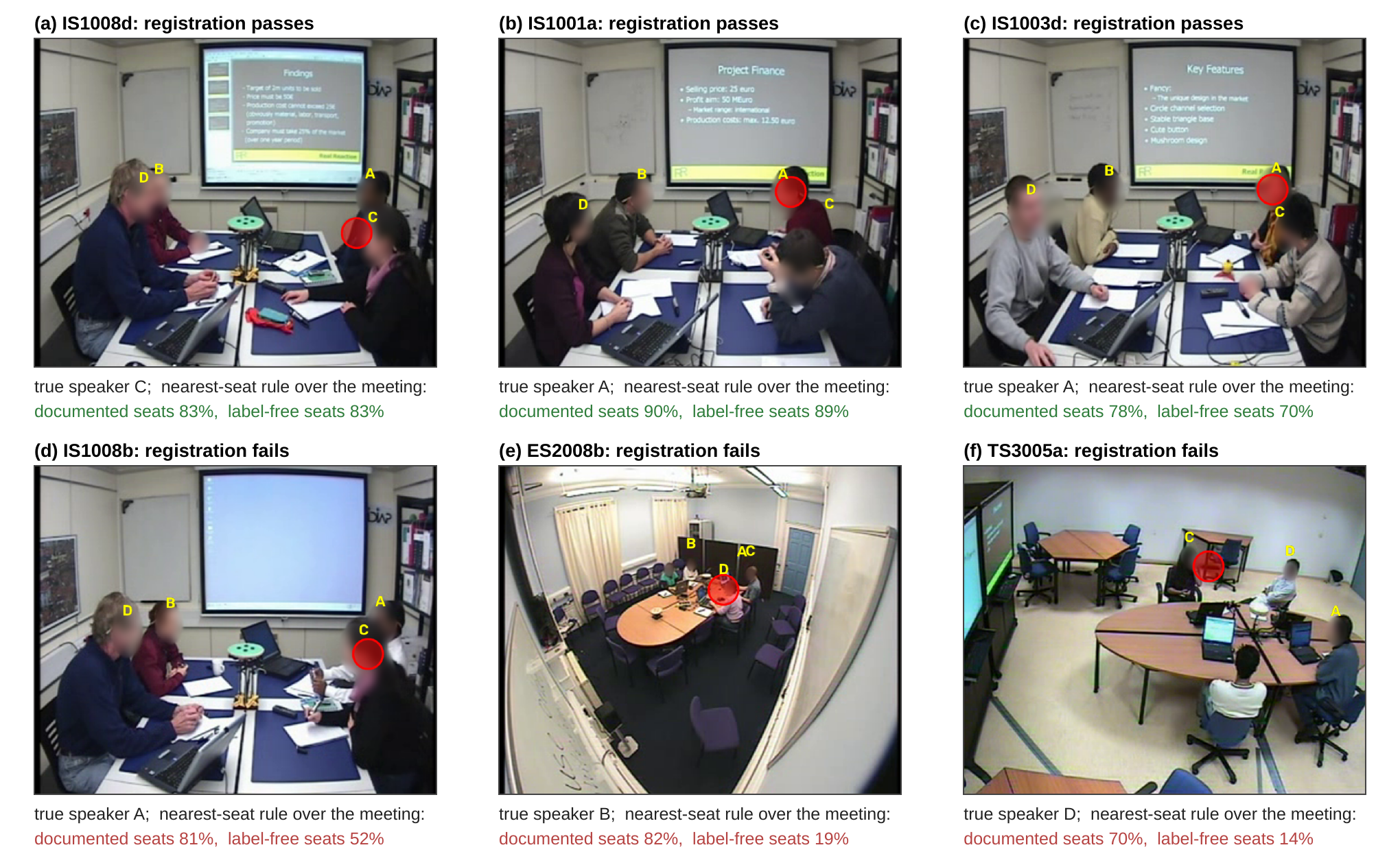}
\caption{\textbf{The label-free registration on six meetings.} One held-out single-speaker moment per meeting, from
the second half of the meeting after fitting on the first; the red circle is the measured direction of the sound
placed in the room view by the registration, and the letters are where the registration puts each participant.
Below each frame: speaker accuracy of the no-model nearest-seat rule over the whole meeting with the documented
seats and with the label-free ones. \textbf{(a)--(c)}~Idiap meetings where the registration passes its quality check (Sec.~\ref{sec:online}).
\textbf{(d)--(f)}~Meetings where it fails: an Idiap meeting where one seat is anchored to a wrong direction, and the
Edinburgh and TNO views, in which the faces are tiny or turned away. Faces are blurred here only.}
\label{fig:supp_regexamples}
\vspace*{100pt}
\end{figure*}
\section{Limitations and Broader Impact}
\label{sec:limits}
The speaker and addressee experiments use reference transcripts, the speaker labels of the preceding turns and
annotated utterance boundaries, so speech recognition and segmentation are outside the evaluation. The label-free
registration needs a camera on which the seated faces are resolved; on the low-resolution room views of Edinburgh
and TNO the surveyed seat directions are used instead. The addressee term of the reward relies on a critic trained
with labels from other sites. The co-location study uses one corpus of $20$ sequences in which nearly all faces are
visible; a larger far-field corpus such as MISP~\cite{misp} is future work. We have not tested on a robot.

A system that keeps a registry of faces, voices and seat positions, and that decides who addresses whom, processes
biometric and behavioural data of everyone in the room, including people who did not choose to interact with it.
All experiments here use public research corpora recorded with consent, and no new data were collected. A deployed
system would need consent of the people recorded, storage limited to the session, and a way for a participant to be
excluded; the anonymous identities of our registry are not linked to names, but a voice embedding and a face
embedding together identify a person as well as a name does.

\section{Qualitative Examples}
\label{sec:examples}
This section collects real utterances and frames from the experiments; each figure gives the frame the model
received, and every face is blurred here only. Fig.~\ref{fig:supp_regexamples} shows the label-free registration of
the pilot on six meetings, Fig.~\ref{fig:supp_pilot} utterances of the room-view pilot, and Figs.~\ref{fig:supp_examples}
and~\ref{fig:supp_examples2} utterances of the main evaluation on AMI's close-up cameras and EasyCom's glasses camera.

On Idiap's room view the registration places the measured direction on the speaker, so the red circle lands on the
person who is talking; in the Edinburgh and TNO rooms the faces are a few pixels apart or turned away from the camera,
the anchors attach to the wrong seats, and the circle lands on the wrong person (Sec.~\ref{sec:speaker}).

The pilot shows the same pattern as the main evaluation on a single, low-resolution camera: the circle corrects the
speaker even under overlap, a second voice can put it on the wrong person, whom the model then names, and the
addressee is often given as the group.

Fig.~\ref{fig:supp_examples} shows six utterances chosen to cover the cases of Table~\ref{tab:supp_speaker}, three
from each kind of camera: the direction corrects a wrong visual answer (a, d); the direction is wrong and the model
follows it to a wrong speaker although its visual answer was right (b, f); the direction is wrong and the model
keeps its correct answer, which happens mainly in the egocentric view (e); and the speaker is right but the
addressee is given as the group (c).

\newpage
Fig.~\ref{fig:supp_examples2} covers the remaining input conditions of the main results table. When the speaker is
outside the field of view the direction is an arrow at the image edge, which is enough in (a) and not in (b),
where the model answers correctly only once the evidence card names the array's speaker; this is the rise from
$35.2\%$ to $48.1\%$ for off-camera speakers in the main paper. In (c) the card also recovers an utterance
addressed to the camera wearer, who is never visible. Panel (d) is the shuffled-direction control: given the
direction of another utterance, the model abandons a visual answer for the marked participant. In (e) the marker
alone, without the sentence that explains it, already moves the answer to the marked participant.

\setcounter{figure}{6}
\begin{figure*}[p]
\centering
\includegraphics[width=0.74\textwidth]{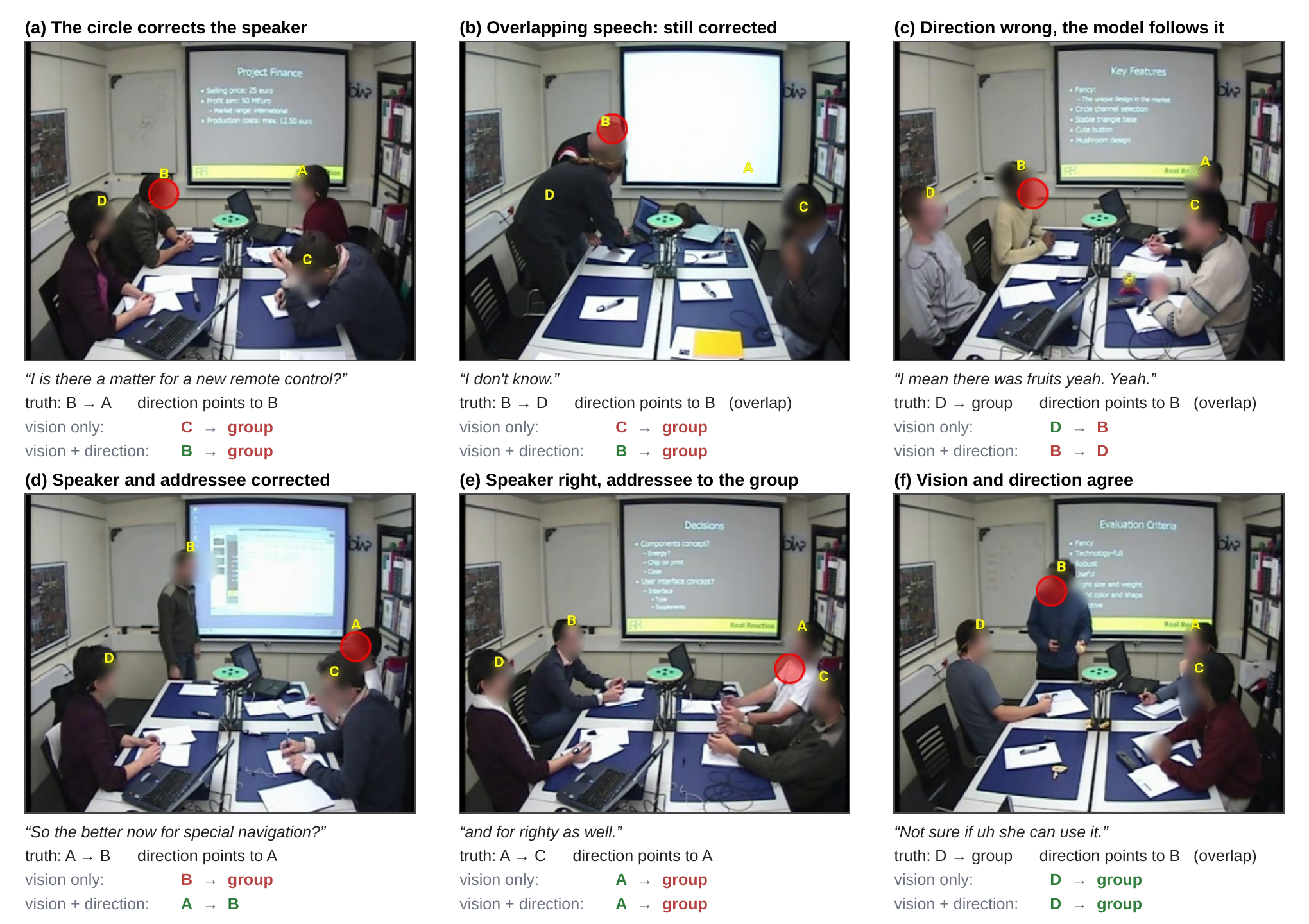}
\caption{\textbf{The pilot on the AMI room view} ($400$ Idiap utterances, zero-shot $27$B model). The middle of the
three frames the model received: one wide camera, the participants' letters, and a red circle where the label-free
registration places the measured direction. Answers without and with the circle (green: correct; red: wrong).
\textbf{(a)--(b)}~The circle corrects the speaker, also under overlap. \textbf{(c)}~A second voice puts the circle
on the wrong person and the model follows it. \textbf{(d)}~Speaker and addressee both corrected.
\textbf{(e)}~The addressee goes to the group. \textbf{(f)}~Vision and direction agree. Faces are blurred here only.}
\label{fig:supp_pilot}
\end{figure*}
\setcounter{figure}{7}
\begin{figure*}[p]
\centering
\includegraphics[width=0.79\textwidth]{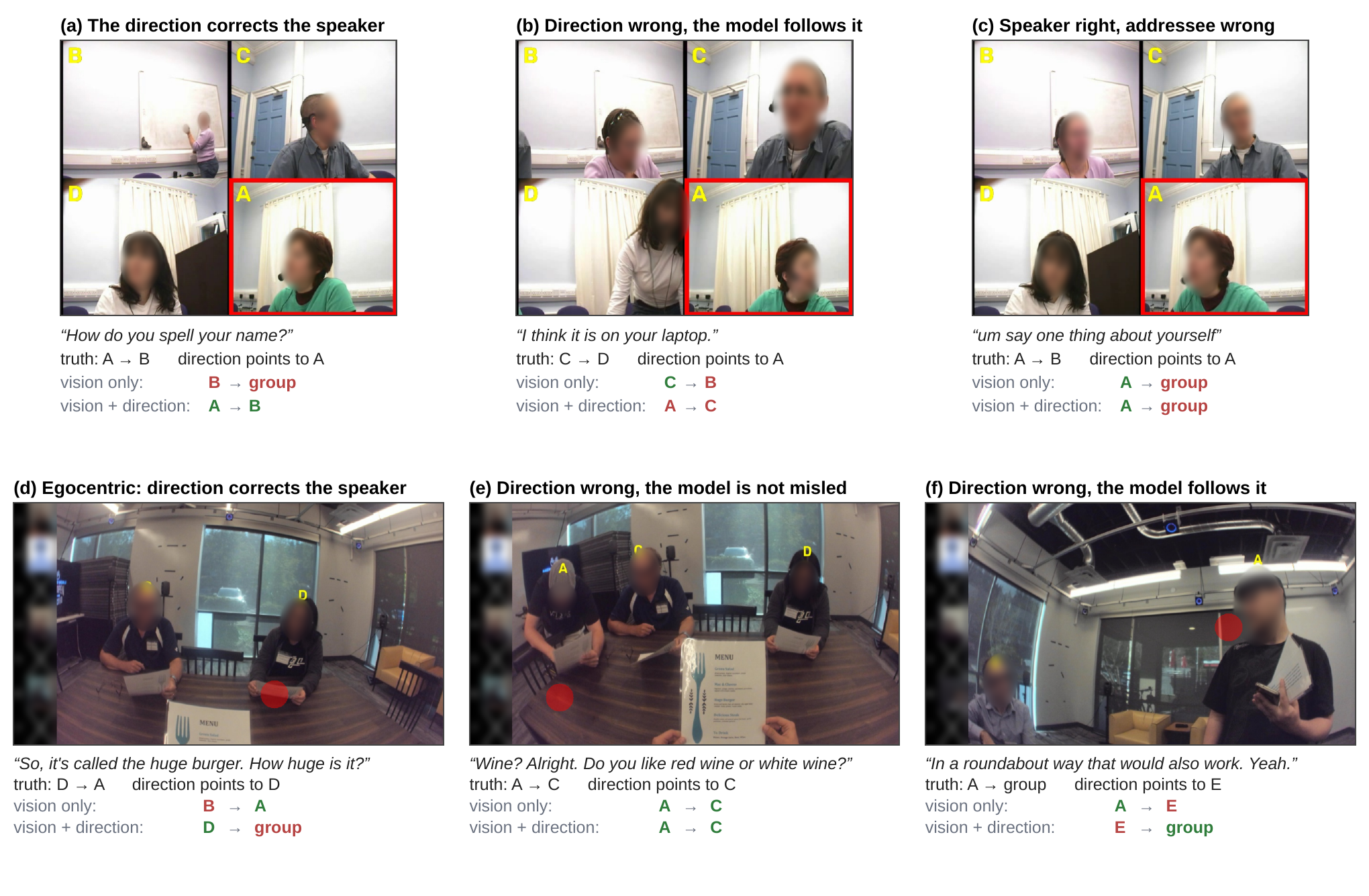}
\caption{\textbf{Evaluation utterances with the frame the model received} (the middle of three; only the faces,
and on EasyCom the corpus's strip of participant portraits at the left, are blurred here). Each panel gives
the utterance, the gold speaker and addressee, the participant the measured direction points to, and the answer of the
zero-shot $27$B model without and with the direction (green: correct; red: wrong). \textbf{(a)--(c)}~AMI: the red
frame marks the panel the direction points to. \textbf{(d)--(f)}~EasyCom: the red circle is the direction in the
glasses camera.}
\label{fig:supp_examples}
\end{figure*}
\setcounter{figure}{8}
\begin{figure*}[p]
\centering
\includegraphics[width=0.79\textwidth]{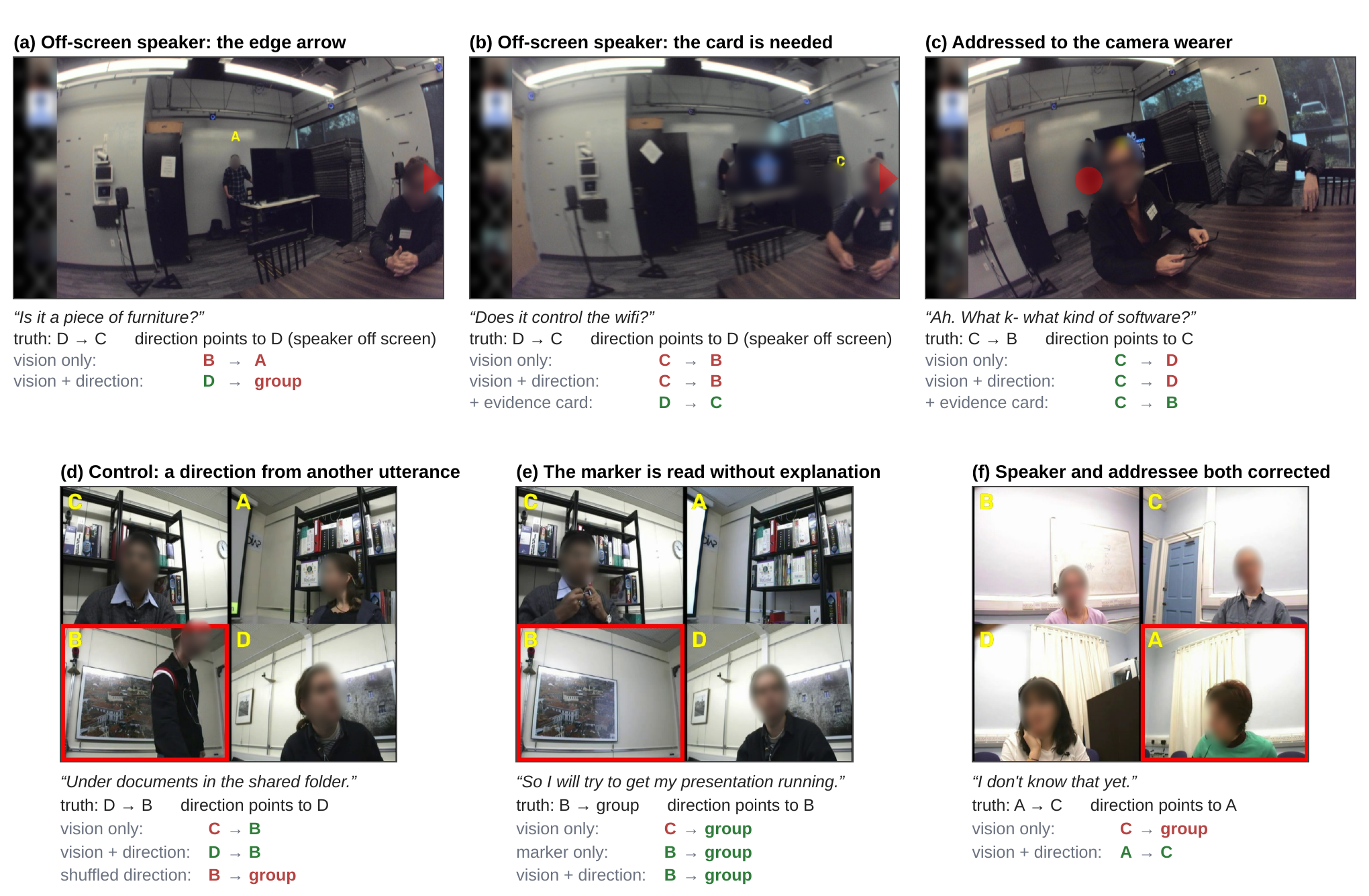}
\caption{\textbf{Off-screen speakers, the evidence card and the control conditions} (layout as in
Fig.~\ref{fig:supp_examples}; only the faces are blurred here, and on EasyCom the direction marker is redrawn
where the model saw it). \textbf{(a)--(b)}~The speaker is outside the glasses camera and the direction is a red arrow at the image
edge; in (b) the model names the speaker only when the card states the array's speaker in words.
\textbf{(c)}~An utterance addressed to the camera wearer, recovered with the card. \textbf{(d)}~The control of
the main paper: the frame carries the direction of another utterance and the model follows it. \textbf{(e)}~The
marker is drawn but the prompt does not say what it means. \textbf{(f)}~The direction corrects the speaker, and
with the right speaker the model also finds the addressee.}
\label{fig:supp_examples2}
\end{figure*}
\FloatBarrier

%% file: tab_supp_hparams.tex
\begin{table*}[tb]
\setcounter{table}{3}
\caption{\textbf{Training settings.} All runs use one $48$\,GB GPU and three seeds; answers are at most $48$ new tokens. Evaluation is greedy and takes about $2.6$\,s per utterance for the 7B policy and $2.1$\,s for the zero-shot $27$B model ($0.9$\,s with the transcript only).}
\label{tab:supp_hparams}
\centering
\footnotesize
\setlength{\tabcolsep}{4pt}
\sbox{\supptabbox}{%
\begin{tabular}{@{}lllll@{}}
\toprule
 & GRPO & reward-ranked FT & DPO & supervised FT \\
\midrule
Policy & \multicolumn{4}{l}{Qwen2.5-Omni-7B (language model part), LoRA rank $16$, $\alpha=32$, dropout $0.05$ on the attention projections} \\
\midrule
Training utterances & 594 & 594 & 594 & 4{,}000 per fold \\
Samples per utterance & 4 & 8 & 8 & -- \\
Sampling temperature & 1.0 & 1.0 & 1.0 & -- \\
Steps / epochs & 300 steps & 2 epochs & 2 epochs & 1 epoch \\
Learning rate & $5\!\times\!10^{-5}$ & $2\!\times\!10^{-5}$ & $2\!\times\!10^{-5}$ & $2\!\times\!10^{-5}$ \\
Batch & 8 answers / step & accumulate 8 & accumulate 8 & accumulate 8 \\
Other & clip $0.2$, no KL term & -- & $\beta_{\mathrm{DPO}}=0.1$ & -- \\
Evidence card in prompt & no & yes & yes & yes \\
GPU-hours per run & 2.3 & 0.8 & 1.5 & 2.8 \\
\bottomrule
\end{tabular}}%
\typeout{TABWIDTH tab:supp_hparams \the\wd\supptabbox\space of \the\linewidth}%
\resizebox{\linewidth}{!}{\usebox{\supptabbox}}
\end{table*}

%% file: tab_supp_speaker.tex
\begin{table*}[tb]
\setcounter{table}{4}
\caption{\textbf{Speaker accuracy against a baseline with no model, and split by whether the measured direction is right}, on the full evaluation set ($9{,}498$ utterances; values in percent). ``Nearest seat'' names the participant whose seat is nearest the measured direction. Brackets: $95\%$ interval of the pooled value (bootstrap over meetings and sessions). The last three columns are pooled: accuracy on utterances where the nearest seat is the true speaker (73.9\% of utterances), accuracy where it is not, and how often the model names the nearest-seat participant. The first five model rows are the zero-shot $27$B model; the 7B rows see the marker and no evidence card (three seeds pooled for the trained policies).}
\label{tab:supp_speaker}
\centering
\small
\setlength{\tabcolsep}{4pt}
\sbox{\supptabbox}{%
\begin{tabular}{@{}lccccccc@{}}
\toprule
 & \multicolumn{4}{c}{speaker accuracy} & \multicolumn{3}{c}{pooled} \\
\cmidrule(lr){2-5}\cmidrule(lr){6-8}
input / system & Idiap & Ed./TNO & EasyCom & pooled & dir.\ right & dir.\ wrong & follows dir. \\
\midrule
Nearest seat, no model & 79.8 & 79.2 & 63.6 & 73.9 {\scriptsize[70.3, 77.4]} & -- & -- & -- \\
\midrule
Vision & 52.4 & 49.2 & 59.2 & 53.9 {\scriptsize[50.7, 58.2]} & 57.4 & 43.8 & 50.0 \\
Vision, marker not explained & 75.3 & 72.9 & 60.6 & 69.5 {\scriptsize[65.4, 73.4]} & 82.3 & 33.2 & 74.0 \\
Vision + direction & 79.7 & 78.1 & 66.0 & 74.5 {\scriptsize[70.6, 78.0]} & 90.4 & 29.3 & 81.9 \\
Vision + direction + card & 80.3 & 80.5 & 69.1 & 76.4 {\scriptsize[73.4, 79.5]} & 97.6 & 16.4 & 92.7 \\
Vision + shuffled dir. & 35.1 & 36.3 & 60.3 & 44.3 {\scriptsize[39.8, 50.2]} & 45.4 & 41.2 & 39.7 \\
\midrule
7B untrained & 46.5 & 59.1 & 50.7 & 51.5 {\scriptsize[47.9, 55.7]} & 58.3 & 32.4 & 52.5 \\
7B self-evaluation & 63.6 & 64.4 & 52.5 & 59.9 {\scriptsize[56.5, 63.7]} & 71.9 & 26.1 & 66.0 \\
7B direction reward & 67.5 & 67.5 & 59.0 & 64.5 {\scriptsize[61.4, 68.1]} & 77.5 & 27.9 & 70.5 \\
\bottomrule
\end{tabular}}%
\typeout{TABWIDTH tab:supp_speaker \the\wd\supptabbox\space of \the\linewidth}%
\resizebox{\linewidth}{!}{\usebox{\supptabbox}}
\end{table*}

%% file: tab_supp_addressee.tex
\begin{table*}[tb]
\setcounter{table}{5}
\caption{\textbf{Addressee detection against trivial baselines} on the full evaluation set (percent). ``Previous other speaker'' names the last different speaker in the context. ``Critic alone'' is the argmax of the critic used in the reward and the evidence card. ``answers group'': share of answers that are ``group''; the gold share is 49.8, 52.7 and 40.0\% on the three datasets.}
\label{tab:supp_addressee}
\centering
\small
\setlength{\tabcolsep}{4pt}
\sbox{\supptabbox}{%
\begin{tabular}{@{}lcccccccccc@{}}
\toprule
 & \multicolumn{4}{c}{accuracy} & \multicolumn{3}{c}{macro-F1} & \multicolumn{3}{c}{answers group} \\
\cmidrule(lr){2-5}\cmidrule(lr){6-8}\cmidrule(lr){9-11}
system & Idiap & Ed./TNO & EasyCom & pooled & Idiap & Ed./TNO & EasyCom & Idiap & Ed./TNO & EasyCom \\
\midrule
Always ``group'' & 49.8 & 52.7 & 40.0 & 47.2 & 13.3 & 13.8 & 8.2 & 100.0 & 100.0 & 100.0 \\
Previous other speaker & 44.0 & 46.8 & 37.2 & 42.4 & 46.4 & 48.6 & 36.2 & 13.8 & 16.1 & 8.8 \\
Critic alone & 55.3 & 59.7 & 49.4 & 54.5 & 44.1 & 44.9 & 34.4 & 60.4 & 67.3 & 71.5 \\
\midrule
27B transcript only & 53.7 & 61.9 & 51.5 & 55.2 & 43.8 & 51.7 & 41.6 & 58.9 & 63.0 & 53.9 \\
27B vision & 53.9 & 61.6 & 49.4 & 54.5 & 42.6 & 48.5 & 42.8 & 62.4 & 67.0 & 38.0 \\
27B vision + direction & 57.0 & 61.4 & 51.6 & 56.3 & 47.7 & 48.3 & 43.3 & 63.9 & 70.4 & 47.0 \\
27B vision + direction + card & 56.5 & 56.7 & 52.2 & 55.0 & 50.6 & 47.6 & 46.3 & 48.8 & 52.8 & 36.3 \\
7B cross-site supervised & 56.3 & 61.5 & 52.7 & 56.5 & 40.8 & 46.3 & 42.6 & 77.4 & 76.5 & 50.2 \\
\bottomrule
\end{tabular}}%
\typeout{TABWIDTH tab:supp_addressee \the\wd\supptabbox\space of \the\linewidth}%
\resizebox{\linewidth}{!}{\usebox{\supptabbox}}
\end{table*}

%% file: tab_rq3.tex
\begin{table}[t]
\caption{\textbf{Verifiable reward versus self-evaluation} on the full evaluation set (three seeds per cell, pooled): the same policy, data and optimizer, trained with our verifiable reward or with a self-evaluation reward. Values are percentages, $\Delta$ rows are differences in points, and the best reward per column is in bold. $^{*}$: the paired $95\%$ cluster-bootstrap interval excludes zero. Pooled intervals: GRPO $[+3.8,+5.4]$; reward-ranked fine-tuning, macro-F1 $[+4.4,+7.2]$ and accuracy $[+0.3,+3.0]$; DPO, macro-F1 $[+20.4,+26.6]$ and accuracy $[-1.8,+8.2]$; GRPO against the matched control $[+12.8,+14.9]$ (seeds 0, 1, 2). DPO's reward critic was trained on Edinburgh/TNO labels, so that site~is~\mbox{excluded}~there.}
\label{tab:rq3}
\centering
\small
\setlength{\tabcolsep}{3pt}
\sbox{\supptabbox}{%
\begin{tabular}{@{}lcccc@{}}
\toprule
reward & Idiap & Ed./TNO & EasyCom & pooled \\
\midrule
\multicolumn{5}{c}{\textbf{GRPO: speaker accuracy ($\uparrow$)}} \\
\midrule
Untrained policy & 46.5 & 59.1 & 50.7 & 51.5 \\
Self-evaluation & 63.6 & 64.4 & 52.5 & 59.9 \\
Self-eval., masked & 53.0 & 52.8 & 46.6 & 50.7 \\
Direction (ours) & \textbf{67.5} & \textbf{67.5} & \textbf{59.0} & \textbf{64.5} \\
$\Delta$ ours $-$ self & $+3.9^{*}$ & $+3.1^{*}$ & $+6.5^{*}$ & $+4.6^{*}$ \\
$\Delta$ ours $-$ masked & $+14.6^{*}$ & $+14.7^{*}$ & $+12.4^{*}$ & $+13.8^{*}$ \\
\midrule
\multicolumn{5}{c}{\textbf{Reward-ranked fine-tuning: addressee macro-F1 ($\uparrow$)}} \\
\midrule
Self-evaluation & 32.3 & 20.4 & 26.9 & 26.5 \\
Ours & \textbf{39.8} & \textbf{28.2} & \textbf{29.4} & \textbf{32.5} \\
$\Delta$ macro-F1 & $+7.5^{*}$ & $+7.8^{*}$ & $+2.5^{*}$ & $+5.9^{*}$ \\
$\Delta$ accuracy & $+2.7$ & $+3.3^{*}$ & $-0.4$ & $+1.8^{*}$ \\
\midrule
\multicolumn{5}{c}{\textbf{DPO: addressee macro-F1 ($\uparrow$)}} \\
\midrule
Self-evaluation & 17.1 & -- & 12.9 & 15.0 \\
Ours & \textbf{44.0} & -- & \textbf{33.1} & \textbf{38.6} \\
$\Delta$ macro-F1 & $+26.9^{*}$ & -- & $+20.2^{*}$ & $+23.6^{*}$ \\
$\Delta$ accuracy & $+3.1$ & -- & $+3.5^{*}$ & $+3.3$ \\
\bottomrule
\end{tabular}}%
\resizebox{\linewidth}{!}{\usebox{\supptabbox}}
\end{table}

%% file: tab_supp_seeds.tex
\begin{table*}[tb]
\setcounter{table}{6}
\caption{\textbf{The GRPO reward comparison per seed}: speaker accuracy on the full evaluation set (percent; the policy sees the marker and no evidence card unless stated; ``--'': not run). ``Gold labels'': the same reward with the measured direction replaced by the gold speaker label (correctness minus the calibration penalty, addressee tokens masked). The last two rows: the untrained policy and seed~0 of the direction reward evaluated on frames without the red marker, scored from the model's own answer.}
\label{tab:supp_seeds}
\centering
\small
\setlength{\tabcolsep}{5pt}
\sbox{\supptabbox}{%
\begin{tabular}{@{}lccccccccccccc@{}}
\toprule
reward & \multicolumn{3}{c}{Idiap} & \multicolumn{3}{c}{Edinburgh/TNO} & \multicolumn{3}{c}{EasyCom} & \multicolumn{4}{c}{pooled} \\
\cmidrule(lr){2-4}\cmidrule(lr){5-7}\cmidrule(lr){8-10}\cmidrule(l){11-14}
 & seed 0 & seed 1 & seed 2 & seed 0 & seed 1 & seed 2 & seed 0 & seed 1 & seed 2 & seed 0 & seed 1 & seed 2 & mean\,$\pm$\,s.d. \\
\midrule
Untrained & \multicolumn{3}{c}{46.5} & \multicolumn{3}{c}{59.1} & \multicolumn{3}{c}{50.7} & \multicolumn{3}{c}{51.5} & -- \\
Self-evaluation & 56.2 & 70.9 & 63.8 & 58.0 & 71.2 & 64.0 & 49.3 & 53.4 & 54.9 & 54.3 & 64.8 & 60.7 & 59.9\,$\pm$\,5.3 \\
Self-evaluation, masked & 58.4 & 36.7 & 63.8 & 60.1 & 35.4 & 62.9 & 47.9 & 34.7 & 57.3 & 55.2 & 35.7 & 61.3 & 50.7\,$\pm$\,13.4 \\
Direction reward & 67.8 & 64.2 & 70.7 & 69.2 & 65.6 & 67.9 & 59.1 & 57.2 & 60.7 & 65.1 & 62.1 & 66.4 & 64.5\,$\pm$\,2.2 \\
Gold labels & 68.5 & 68.7 & 68.5 & 66.2 & 64.6 & 67.0 & 61.7 & 55.3 & 63.8 & 65.5 & 62.8 & 66.4 & 64.9\,$\pm$\,1.9 \\
\midrule
\multicolumn{14}{@{}l}{\textit{Evaluated on frames without the marker}} \\
\quad untrained & \multicolumn{3}{c}{40.1} & \multicolumn{3}{c}{42.0} & \multicolumn{3}{c}{49.2} & \multicolumn{3}{c}{43.8} & -- \\
\quad direction reward & 53.7 & -- & -- & 45.2 & -- & -- & 55.0 & -- & -- & 51.8 & -- & -- & -- \\
\bottomrule
\end{tabular}}%
\typeout{TABWIDTH tab:supp_seeds \the\wd\supptabbox\space of \the\linewidth}%
\resizebox{\linewidth}{!}{\usebox{\supptabbox}}
\end{table*}

%% file: tab_supp_probe.tex
\begin{table}[tb]
\setcounter{table}{7}
\caption{\textbf{How well each reward ranks correct answers}: AUROC for flagging a correct speaker among sampled answers of the untrained policy (1585 parsed samples of $200$ Idiap pilot utterances, from the development site, eight samples each), over all utterances, those with a single speaker and those with overlap. Self-evaluation is the mean of the three signals below it. ``Within utt.'' compares only answers to the same utterance, which is what a policy-gradient update uses.}
\label{tab:supp_probe}
\centering
\small
\setlength{\tabcolsep}{4pt}
\sbox{\supptabbox}{%
\begin{tabular}{@{}lcccc@{}}
\toprule
reward signal & all & single & overlap & within utt. \\
\midrule
Direction reward & 0.79 & 0.91 & 0.65 & 0.80 \\
Direction + lip motion & 0.80 & 0.91 & 0.67 & 0.77 \\
Self-evaluation & 0.57 & 0.58 & 0.55 & 0.57 \\
\quad answer probability & 0.60 & 0.63 & 0.57 & 0.60 \\
\quad self-check & 0.54 & 0.56 & 0.52 & 0.54 \\
\quad stated confidence & 0.53 & 0.53 & 0.52 & 0.52 \\
\bottomrule
\end{tabular}}%
\typeout{TABWIDTH tab:supp_probe \the\wd\supptabbox\space of \the\linewidth}%
\resizebox{\linewidth}{!}{\usebox{\supptabbox}}
\end{table}

%% file: tab_supp_baserate.tex
\begin{table}[tb]
\setcounter{table}{8}
\caption{\textbf{Choosing the base-rate weight $\beta$ of the addressee term} on training meetings (no test utterance is involved). Columns: AUROC of the term for the correct addressee within an utterance; mean reward of the correct answer, of always answering ``group'' and of always naming the previous speaker; the margin of the correct answer over the better constant strategy; and how often the term's best answer is ``group'' (percent). The rule fixed in advance was the highest AUROC whose margin is at least $0.3$; the chosen value is in bold.}
\label{tab:supp_baserate}
\centering
\small
\setlength{\tabcolsep}{4pt}
\sbox{\supptabbox}{%
\begin{tabular}{@{}ccccccc@{}}
\toprule
$\beta$ & AUROC & correct & ``group'' & prev.\ speaker & margin & best = group \\
\midrule
0.0 & 0.820 & -0.85 & -0.87 & -1.43 & 0.02 & 65 \\
0.1 & 0.821 & -0.75 & -0.81 & -1.30 & 0.07 & 63 \\
0.2 & 0.821 & -0.63 & -0.75 & -1.16 & 0.11 & 61 \\
0.3 & 0.819 & -0.52 & -0.68 & -1.01 & 0.16 & 59 \\
0.4 & 0.816 & -0.40 & -0.62 & -0.86 & 0.21 & 57 \\
0.5 & 0.810 & -0.29 & -0.55 & -0.70 & 0.26 & 54 \\
\textbf{0.6} & 0.807 & -0.17 & -0.48 & -0.53 & 0.31 & 51 \\
0.7 & 0.802 & -0.05 & -0.42 & -0.36 & 0.31 & 49 \\
0.8 & 0.792 & 0.07 & -0.35 & -0.19 & 0.26 & 46 \\
0.9 & 0.783 & 0.19 & -0.28 & -0.02 & 0.21 & 42 \\
1.0 & 0.771 & 0.31 & -0.21 & 0.15 & 0.16 & 40 \\
\bottomrule
\end{tabular}}%
\typeout{TABWIDTH tab:supp_baserate \the\wd\supptabbox\space of \the\linewidth}%
\resizebox{\linewidth}{!}{\usebox{\supptabbox}}
\end{table}

%% file: tab_doa.tex
\begin{table}[bt]
\caption{\textbf{Effect of DOA on AVDIAR}: speaker-to-face recall in percent for direction-only attribution, best
per row in bold. Keeping only the strongest SRP-PHAT peak cannot attribute a second concurrent speaker;
top-$k$ multi-source peak-picking is what makes direction usable under overlap.}
\label{tab:doa_ablation}
\centering
\small
\begin{tabular*}{\linewidth}{@{\extracolsep{\fill}}lccc@{}}
\toprule
 &  & single-source & multi-source \\
condition & $n$ & ($k{=}1$) & ($k{=}\#$\,speakers) \\
\midrule
single-speaker                     &  85 & \textbf{74} & \textbf{74} \\
overlap, $\Delta\theta<15^\circ$   &  78 & 35 & \textbf{59} \\
overlap, $15$--$30^\circ$          & 215 & 65 & \textbf{75} \\
overlap, $\Delta\theta\ge30^\circ$ &  91 & 59 & \textbf{69} \\
all                                & 469 & 60 & \textbf{71} \\
\bottomrule
\end{tabular*}
\end{table}

%% file: tab_supp_fusion.tex
\begin{table*}[tb]
\setcounter{table}{9}
\caption{\textbf{Co-located speakers on AVDIAR, all rules with intervals}: speaker-to-face recall in percent with $95\%$ sequence-cluster bootstrap intervals ($10{,}000$ resamples). ``Fusion'' is the rule of the main paper (direction weight $1$). ``Either'' counts a segment as correct if the direction-only or the visual-only rule is right, an upper bound for any fusion of the two. The last three columns are the fusion with direction weight $0.5$ and $2$ and a fusion learned by cross-validation over sequences (with weight $0$ the fusion is the visual-only rule).}
\label{tab:supp_fusion}
\centering
\small
\setlength{\tabcolsep}{4pt}
\sbox{\supptabbox}{%
\begin{tabular}{@{}lccccccccc@{}}
\toprule
 & & & & & & & \multicolumn{3}{c}{other fusions} \\
\cmidrule(lr){8-10}
segments & $n$ & seq. & direction-only & visual-only & fusion & either & weight $0.5$ & weight $2$ & learned \\
\midrule
single speaker & 85 & 19 & 74.1 {\scriptsize[55, 88]} & 74.1 {\scriptsize[66, 85]} & 75.3 {\scriptsize[67, 85]} & 94.1 {\scriptsize[90, 98]} & 74.1 & 75.3 & 72.9 \\
overlap, $\Delta\theta<15^\circ$ & 78 & 12 & 59.0 {\scriptsize[40, 72]} & 94.9 {\scriptsize[89, 100]} & 94.9 {\scriptsize[92, 98]} & 97.4 {\scriptsize[93, 100]} & 94.9 & 94.9 & 92.3 \\
overlap, $15$--$30^\circ$ & 215 & 19 & 74.9 {\scriptsize[64, 84]} & 94.0 {\scriptsize[90, 98]} & 94.4 {\scriptsize[91, 98]} & 98.1 {\scriptsize[96, 100]} & 94.4 & 95.8 & 96.3 \\
overlap, $\geq30^\circ$ & 91 & 16 & 69.2 {\scriptsize[52, 84]} & 98.9 {\scriptsize[97, 100]} & 98.9 {\scriptsize[97, 100]} & 98.9 {\scriptsize[97, 100]} & 98.9 & 97.8 & 93.4 \\
all segments & 469 & 20 & 71.0 {\scriptsize[60, 81]} & 91.5 {\scriptsize[88, 94]} & 91.9 {\scriptsize[90, 94]} & 97.4 {\scriptsize[96, 99]} & 91.7 & 92.3 & 90.8 \\
\bottomrule
\end{tabular}}%
\typeout{TABWIDTH tab:supp_fusion \the\wd\supptabbox\space of \the\linewidth}%
\resizebox{\linewidth}{!}{\usebox{\supptabbox}}
\end{table*}

%% file: tab_revision.tex
\begin{table}[b]
\caption{\textbf{Effect of revision}, all values in percent, best per corpus in bold. Relative to the $0.5$\,s
committed stream: the fraction of active frames a longer horizon would relabel; of the committed
\emph{errors}, the fraction it fixes; of the committed \emph{correct} frames, the fraction it breaks; and
the net fraction of committed errors removed. Streaming revision is safe but saturates within seconds;
most of the remaining gap to offline is global re-clustering, not~lookahead.}
\label{tab:revision}
\centering
\small
\setlength{\tabcolsep}{6pt}
\sbox{\supptabbox}{%
\begin{tabular}{@{}lcccc@{}}
\toprule
revise to & relabeled & fixed ($\uparrow$) & broken ($\downarrow$) & net ($\uparrow$) \\
\midrule
\multicolumn{5}{c}{\textbf{AMI}} \\
\midrule
$1$\,s   & 14.8 & 20.4 & \textbf{5.0} &  9.7 \\
$2$\,s   & 17.6 & 23.8 & 5.4 & 12.2 \\
$5$\,s   & 18.6 & 26.5 & 5.4 & 15.0 \\
offline  & 28.2 & \textbf{53.5} & 5.9 & \textbf{41.0} \\
\midrule
\multicolumn{5}{c}{\textbf{VoxConverse}} \\
\midrule
$1$\,s   &  6.5 &  9.4 & \textbf{2.2} &  3.7 \\
$2$\,s   &  8.7 & 11.9 & 2.7 &  4.8 \\
$5$\,s   & 10.6 & 14.7 & 3.6 &  5.2 \\
offline  & 26.1 & \textbf{78.7} & 2.4 & \textbf{72.3} \\
\bottomrule
\end{tabular}}%
\resizebox{\linewidth}{!}{\usebox{\supptabbox}}   %
\end{table}

%% file: tab_supp_der.tex
\begin{table}[tb]
\setcounter{table}{10}
\caption{\textbf{Diarization error by component} for each lookahead (percent of reference speech time; $0$\,s collar, overlap scored). The online diarizer is diart on the single-channel headset mix (AMI) or the released audio (VoxConverse); it uses no camera and no array. For offline AMI only the total ($18.8$) is~reported.}
\label{tab:supp_der}
\centering
\small
\setlength{\tabcolsep}{4pt}
\sbox{\supptabbox}{%
\begin{tabular}{@{}llcccc@{}}
\toprule
corpus & system & missed & false alarm & confusion & DER \\
\midrule
AMI & online, 0.5\,s & 12.3 & 4.9 & 13.8 & 31.0 \\
 & online, 1\,s & 11.0 & 3.8 & 13.1 & 27.9 \\
 & online, 2\,s & 13.0 & 3.2 & 11.3 & 27.5 \\
 & online, 5\,s & 12.5 & 3.1 & 10.9 & 26.5 \\
\midrule
VoxConverse & online, 0.5\,s & 3.1 & 2.4 & 22.1 & 27.5 \\
 & online, 1\,s & 2.7 & 2.3 & 21.5 & 26.5 \\
 & online, 2\,s & 4.2 & 2.0 & 20.0 & 26.2 \\
 & online, 5\,s & 4.4 & 2.1 & 19.7 & 26.1 \\
 & offline & 2.3 & 2.3 & 3.1 & 7.6 \\
\bottomrule
\end{tabular}}%
\typeout{TABWIDTH tab:supp_der \the\wd\supptabbox\space of \the\linewidth}%
\resizebox{\linewidth}{!}{\usebox{\supptabbox}}
\end{table}

%% file: tab_supp_coldstart.tex
\begin{table}[tb]
\setcounter{table}{11}
\caption{\textbf{Learning the registration from the stream: all runs.} Speaker-to-face recall (percent) on single-speaker blocks of the held-out second half, by the stream used to fit the anchors. ``Direction'': direction-only rule (undefined until two anchors exist). ``System'': visual-only until two anchors exist, then the fusion, with meeting-level $95\%$ intervals; the visual-only rule scores 60 throughout. ``Three'': the same, waiting for three anchors. ``All 15'': the system on all $15$ meetings (visual-only 47); ``failed'': the seven meetings on which the registration does not pass its check (visual-only 35). A causal run that refits every $15$\,s and scores the blocks that follow gives, by elapsed time: 28\% at 0-0.5\,min, 44\% at 0.5-1\,min, 45\% at 1-2\,min, 64\% at 2-5\,min, 66\% at 5-10\,min, 65\% at 10-20\,min, 67\% at 20+\,min.}
\label{tab:supp_coldstart}
\centering
\small
\setlength{\tabcolsep}{4pt}
\sbox{\supptabbox}{%
\begin{tabular}{@{}cccccc@{}}
\toprule
 & \multicolumn{3}{c}{eight registered meetings} & & \\
\cmidrule(lr){2-4}
stream (min) & direction & system & three & all 15 & failed \\
\midrule
0 & -- & 60 {\scriptsize[50, 70]} & 60 & 47 & 35 \\
0.5 & -- & 60 {\scriptsize[50, 70]} & 60 & 47 & 35 \\
1 & 11 & 40 {\scriptsize[20, 61]} & 40 & 37 & 35 \\
2 & 28 & 51 {\scriptsize[36, 67]} & 52 & 47 & 42 \\
5 & 53 & 65 {\scriptsize[53, 79]} & 66 & 51 & 37 \\
10 & 69 & 72 {\scriptsize[62, 82]} & 72 & 53 & 35 \\
first half & 70 & 70 {\scriptsize[58, 82]} & 70 & 51 & 33 \\
\bottomrule
\end{tabular}}%
\typeout{TABWIDTH tab:supp_coldstart \the\wd\supptabbox\space of \the\linewidth}%
\resizebox{\linewidth}{!}{\usebox{\supptabbox}}
\end{table}

%% file: tab_supp_runtime.tex
\begin{table*}[t]
\setcounter{table}{12}
\caption{\textbf{Compute cost of the online components.} Measured on AMI meeting IS1008a ($944$\,s of audio; the
diarizer also on EN2002a, $2{,}143$\,s) on one 16-core workstation CPU (one thread unless stated) and one $48$\,GB
GPU. RTF: seconds of computation per second of audio. The direction estimate is implemented as a whole-recording
batch, so its per-block figure is the batch cost divided by the number of blocks. The face detector and the
active-speaker scorer are run once per recording and cached; the times given are those logged when the caches
were~built.}
\label{tab:supp_runtime}
\centering
\footnotesize
\setlength{\tabcolsep}{3pt}
\begin{tabular}{@{}llll@{}}
\toprule
component & unit of work & cost & RTF \\
\midrule
SRP-PHAT direction ($8$ mics, $180$ azimuths $\times$ $4$ elevations), CPU & $0.25$\,s block & $2.2$\,ms ($2.0$ with 32 threads) & $0.009$ \\
Registration refit (label-free anchors), CPU & one refit, every $15$\,s & $9$--$12$\,ms & $<0.001$ \\
Speaker-to-face rule (direction, vision, fusion), CPU & $0.25$\,s block & $0.05$\,ms & $<0.001$ \\
Online diarizer (diart; pyannote segmentation 3.0, WeSpeaker ResNet34), GPU & $0.5$\,s step & $13$\,ms (same at $5$\,s lookahead); $114$\,MiB & $0.026$ \\
Face detection (InsightFace, $5$\,fps), cached & recording & $199$\,s for $944$\,s$^{\dagger}$ & $0.21^{\dagger}$ \\
Active-speaker scores (TalkNet, $25$\,fps), cached & recording & $118$\,s for $944$\,s$^{\dagger}$ & $0.12^{\dagger}$ \\
Speaker and addressee answer, zero-shot $27$B model, GPU & utterance (3 frames) & $2.1$\,s ($0.9$\,s transcript only) & -- \\
Speaker and addressee answer, aligned 7B policy, GPU & utterance (3 frames) & $2.6$\,s & -- \\
\bottomrule
\end{tabular}
\end{table*}